\documentclass[11pt]{article}

\usepackage{amsmath,bm,makecell,subfigure}
\usepackage{amsthm}
\usepackage{diagbox}
\usepackage{multirow,enumerate}
\usepackage{threeparttable}
\usepackage{threeparttablex}
\usepackage[ruled,linesnumbered]{algorithm2e}
\usepackage{graphicx}
\usepackage[T1]{fontenc}    
\usepackage{hyperref}       
\usepackage{amsfonts}       
\usepackage[capitalize,noabbrev]{cleveref}
\usepackage{changes,soul}
\usepackage{graphicx}
\usepackage[T1]{fontenc}
\usepackage[utf8]{inputenc}
\usepackage{authblk}
\usepackage{thmtools,thm-restate}
\usepackage{adjustbox}
\usepackage{enumerate}
\usepackage{bm}
\usepackage{amssymb,amsmath}
\usepackage{ulem}
\usepackage{cancel}
\usepackage{subfigure,color}
\usepackage{setspace}

\newtheorem{theorem}{Theorem}
\newtheorem{definition}[theorem]{Definition}
\newtheorem{lemma}[theorem]{Lemma}
\newtheorem{corollary}[theorem]{Corollary}
\newtheorem{remark}[theorem]{Remark}
\newtheorem{proposition}[theorem]{Proposition}

 \usepackage[letterpaper,margin=1in]{geometry} 
 
\title{Understanding Deep Neural Networks via Linear Separability of Hidden Layers}

\author{Chao Zhang$^{1}$, Xinyu Chen$^{1}$, Wensheng Li$^{1}$, Lixue Liu$^{1}$, Wei Wu$^{1}$, Dacheng Tao$^{2}$\\
\small{$^1$School of Mathematical Sciences, Dalian University of Technology, China} \\
	\texttt{czhang1015@foxmail.com} \\
	\small{$^{2}$School of Computer Science, The University of Sydney, Australia} 
}
\date{}
\begin{document}

\maketitle

\begin{abstract}
In this paper, we measure the linear separability of hidden layer outputs to study the characteristics of deep neural networks. In particular, we first propose Minkowski difference based linear separability measures (MD-LSMs) to evaluate the linear separability degree of two points sets. Then, we demonstrate that there is a synchronicity between the linear separability degree of hidden layer outputs and the network training performance, {\it i.e.}, if the updated weights can enhance the linear separability degree of hidden layer outputs, the updated network will achieve a better training performance, and vice versa. Moreover, we study the effect of activation function and network size (including width and depth) on the linear separability of hidden layers. Finally, we conduct the numerical experiments to validate our findings on some popular deep networks including multilayer perceptron (MLP), convolutional neural network (CNN), deep belief network (DBN), ResNet, VGGNet, AlexNet, vision transformer (ViT) and GoogLeNet.
\end{abstract}
 

\section{Introduction}
Deep neural networks have been successfully used to solve many complicated learning tasks, for example, computer vision \cite{zhang2020causal,liu2021emergence,kolesnikov2022uvim}, natural language processing \cite{sood2020improving,bragg2021flex,fries2022bigbio} and other engineering applications \cite{wang2022metateacher,bonnier2022engineering}. The empirical observation that the large network size is beneficial to the generalization performance of deep networks, is contrary to the traditional view of statistical learning theory, which deems that high model complexity will cause the overfitting. In recent years, many research works attempt to explain why deep neural networks perform well in these difficult tasks by using some theoretical analysis tools such as statistical learning theory (SLT) \cite{vapnik1999nature}, PAC-Bayes framework \cite{mcallester1999pac} and neural tangent kernel (NTK) \cite{jacot2018neural}. However, there still remains a gap between these theoretical results and the user experiences of deep neural networks. For example, following the classical STL methods, the complexity measures of deep networks were expressed as increasing functions w.r.t. the network size, and the relevant generalization bounds will become loose when the network size is large ({\it cf.} \cite{neyshabur2015norm,golowich2020size,harvey2017nearly,bartlett2017spectrally}).  Moreover, some results on NTK and over-parameterized neural networks are obtained under the assumption that the network width is infinite ({\it cf.} \cite{du2019gradient,arora2019fine,li2018learning,du2018gradient,song2019quadratic,allen2019convergence, zou2020gradient}). Some recent theoretical works on deep neural networks are summarized in the appendix (part \ref{supp:related}).



\subsection{Background and Motivation}

Two point sets are said to be linearly separable if they can be correctly split by using a hyperplane, and the concept of linear separability plays an important part in measuring the capability of neural networks \cite{2002New,2004Searching,2010Choice}. In the literature, there are two main research issues on the linear separability of neural networks: 1) whether the current network can achieve all dichotomies, {\it i.e.,} the mapping capability \cite{Kurt}; and 2) how many dichotomies can be recorded by a network with the specific structure, {\it i.e.,} the memory capability \cite{2020Memory,2006Geometrical}. However, to the best of our knowledge, there is rarely few work to consider the layer-wise changes of the linear separability degree of two point sets when they pass through a network with multiple hidden layers.  

Our study is initially motivated by the relationship between the linear separability degree of hidden layer outputs and the process of training networks. We would like to know whether the linear separability degree of hidden layer outputs increases when the training accuracy increases. If applicable, the linear separability degree can be treated as a criterion for evaluating the performance of each hidden layer during the training phase. 

Consider a one-hidden-layer MLP ${\rm net}(\cdot):\mathbb{R}^N \rightarrow \{0,1\}$ with the output-layer weight vector ${\bf w}$ and the hidden-layer weight matrix ${\bf V}$. Let ${\rm hid}(\cdot): \mathbb{R}^N \rightarrow \mathbb{R}^H$ be the hidden layer of the MLP, and ${\rm hid}({\cal X})$ be the set of hidden layer outputs w.r.t. the input set ${\cal X}:=\{{\bf x}_m\}_{m=1}^M$. The set ${\rm hid}({\cal X})$ is called hidden-layer outputs in the following discussion, if no confusion arises. Denote $\triangle {\bf w} $ and $\triangle {\bf V}$ as the weight updates provided by a training algorithm implemented on the training set ${\cal S} = \{({\bf x}_m,{\bf y}_m)\}_{m=1}^M\subset \mathbb{R}^N \times \{0,1\}$. The updated network is denoted as ${\rm net}'(\cdot)$ with updated weights ${\bf w}' = {\bf w} + \triangle {\bf w}$ and ${\bf V}' = {\bf V}+\triangle {\bf V}$. Denote ${\rm hid}'(\cdot)$ as the updated hidden layer with the weights ${\bf V}'$. Under these notations, we obtain the following theorem which motivates the research of this paper:

\begin{theorem}[Synchronicity]\label{thm:main}
Assume that the updated weights ${\bf w}'$ achieves the highest classification accuracy on ${\cal S}$ when the hidden-layer weights of ${\rm net}'(\cdot)$ is updated to be ${\bf V}'$. Then, ${\rm net}'(\cdot)$ has higher classification accuracy on ${\cal S}$ than ${\rm net} (\cdot)$ if and only if the linear separability degree of ${\rm hid}'({\cal X})$ is larger than that of ${\rm hid}({\cal X})$.
\end{theorem}
 
This theorem demonstrates the synchronicity between the linear separability degree of hidden layer outputs and the network training performance. When training a network, the change of the linear separability degree of hidden layer outputs synchronizes with the change of the training accuracy. Different from the existing works on the theoretical analysis of neural networks, the linear separability degree provides a feasible manner to directly study the relationship between the structure parameters and the network capability. Accordingly, a desired linear separability measure (LSM) should meet the following requirements: 
\begin{enumerate}[(1)]
	\item It should have a low computational cost, because we would like to layer-wisely examine the linear separability degree of the hidden-layer outputs after each training iteration;
	 
	\item It should be insensitive to the outliers, because the stochastic gradient descent methods sometimes cause abnormal hidden layer outputs;
	 
	\item It should be of a well-defined mathematical form in order to facilitate the further theoretical analysis. 
\end{enumerate}
 
Some mathematical terms mentioned in the existing works actually can be treated as LSMs of two point sets, for example, the generalized Rayleigh quotient $J_{\omega} = \max_{\bm{\omega}} \frac{\bm{\omega}^T {\bf S}_b\bm{\omega}}{\bm{\omega}^T{\bf S}_w\bm{\omega}}$ in Fisher linear discriminant analysis (LDA) and the sum of slack variables in linear support vector machine (L-SVM) with soft margin.\footnote{The discussion on Fisher LDA is arranged in Section \ref{supp:lda}.} However, the LDA-based LSM, which is based on the means of point sets, is sensitive to the outliers in the sets and has a high computational cost because of eigenvalue decomposition; and the computation of the L-SVM based LSM is still time-consuming especially when the sample size is large. Moreover, Ben-Israel \textit{et al}. \cite{Ben-Israel} introduce the linear divisible angle to measure the linear separability degree of two point sets, where the labels of the data are treated as a new attribute to convert the dimension of points from $N$ to $N+1$, and then Fisher LDA is used to compute the generalized Rayleigh quotient of the converted points. Gabidullina \textit{et al}. \cite{Gabidullina} adopted the smallest thickness of the classified hyperplane as the LSM for the linearly inseparable sets. Since this measure is computed via a minimax optimization problem, it has a high computational cost as well.

\subsection{Overview of Main Results}

In Section \ref{sec:md}, we propose the Minkowski difference-based LSM (MD-LSM) for two point sets, and provide an algorithm of finding the largest linearly separable subsets of arbitrary two linearly inseparable sets. Then, we show the relationship between the linear separability degree of two linearly inseparable sets and the best classification accuracy provided by all possible linear models. Moreover, we provide the alternative versions of MD-LSM to reduce the cost of computing the MD-LSM of each hidden layer after each training epoch. 

In Section \ref{sec:parameter}, we study the effects of activation function and network size on the linear separability of hidden layers.  First, the hidden layer of a neural networks is regarded as the pseudo-linear mapping (PLM) ({\it cf.} \cite{2020Memory}). Then, we provide a sufficient condition that a PLM changes the relative position between one Minkowski difference (MD) point and the splitting hyperplane. Moreover, based on the random matrix theory, we have proven that the increase of network size can enlarge the probability of increasing the linear separability degree of hidden layer outputs.

In Section \ref{sec:experiment}, we conduct the numerical experiments to validate the theoretical findings of this paper on some well-known networks including multilayer perceptron (MLP) \cite{bishop1994neural}, convolutional neural network (CNN) \cite{lecun1998gradient}, deep brief network (DBN) \cite{hinton2006fast}, ResNet \cite{he2016deep}, VGGNet \cite{simonyan2014very}, AlexNet \cite{krizhevsky2017imagenet}, vision transformer (ViT) \cite{dosovitskiy2020image} and GoogLeNet \cite{szegedy2015going} in classification tasks, respectively. 

In Section \ref{supp:lda}, we discuss the difference between the proposed MD-LSM and Fisher LDA, and then explain why we use MD-LSMs to evaluate the linear separability degree of hidden layer outputs rather than the Rayleigh quotient in Fisher LDA. In Section \ref{supp:proof}, we prove the main results of this paper, and the last section concludes the paper. In the  appendix, we briefly sketch the recent works on theoretical analysis of deep networks (part \ref{supp:related}), and then provide the complete experimental report (part \ref{supp:experiment}). 

\section{Minkowski Difference Based Linear Separability Measure (MD-LSM)} \label{sec:md}

In this section, we present the concept of MD-LSM for two point sets and then give an algorithm to find the maximum linearly separable subset of two linearly inseparable point sets. Moreover, we provide some alternative versions of MD-LSMs with low computation cost.
 
\subsection{Minkowski Difference and Maximum Linearly Separable Subset}

First, we introduce the concept of Minkowski difference (MD), which has been widely used for the applications in many areas such as data classification \cite{mampaey2012efficient,takeda2013unified}, motion planning for robots \cite{lozano1979algorithm}, real-time collision detection \cite{ericson2004real} and computer graphics \cite{ghosh1993unified,ghosh1990solution}.
\begin{definition}[Minkowski Difference]
	Let ${\cal A}=\left\{ \mathbf{a}_1,\cdots,\mathbf{a}_I \right\} \subset \mathbb{R}^N$ and ${\cal B}=\left\{ \mathbf{b}_1,\cdots,\mathbf{b}_J \right\} \subset \mathbb{R}^N$ be two point sets. Then, the Minkowski difference between the two sets is defined as 
	\begin{equation*}
		\mathrm{MD}({\cal A},{\cal B}) := \big\{ \mathbf{m}_{ij}:= \mathbf{a}_i-\mathbf{b}_j\in \mathbb{R}^N\ | \; \mathbf{a}_i\in {\cal A},\;\mathbf{b}_j\in {\cal B} \big\}.
	\end{equation*}
\end{definition}
 \noindent Based on Minkowski difference, we convert the linear separability of two point sets into the relative position relationship between a point set and a hyperplane that passes the origin. 
\begin{theorem}\label{thm:md}
	Two points sets ${\cal A},{\cal B}\in\mathbb{R}^N$ are linearly separable if and only if there exists a vector $\bm{\omega}\in\mathbb{R}^N$ such that all points of $\mathrm{MD}({\cal A},{\cal B})$ locates in one side of the hyperplane $\bm{\omega}^T\mathbf{x}=0$, $\mathbf{x}\in\mathbb{R}^N$.
\end{theorem}
 
 \noindent We note that the hyperplane $\bm{\omega}^T {\bf m} = 0$ parallels to the separating hyperplane between the two classes ${\cal A}$ and ${\cal B}$ if they are linearly separable. Additionally, if the two sets ${\cal A}$ and ${\cal B}$ are linearly inseparable, some points of $\mathrm{MD}({\cal A},{\cal B})$ will lie in one side of the hyperplane and the rest lie in the other side: 
\vspace{-0.5mm}
\begin{definition}[Minor and Major Sides]
	Given a hyperplane $\bm{\omega}^T {\bf m} = 0$, if more than half points of $\mathrm{MD}({\cal A},{\cal B})$ lie in one side of $\bm{\omega}^T {\bf m} =0$, then this side is said to be the major side of the hyperplane; and accordingly, the other side of $\bm{\omega}^T {\bf m} =0$ is said to be the minor side of the hyperplane. 
\end{definition}
 \noindent Furthermore, denote ${\rm major}_{\bm{\omega}}(\mathrm{MD}({\cal A},{\cal B}))$ (resp. ${\rm minor}_{\bm{\omega}}(\mathrm{MD}({\cal A},{\cal B}))$) as the subset of $\mathrm{MD}({\cal A},{\cal B})$ containing all points of $\mathrm{MD}({\cal A},{\cal B})$ that locate in the major (resp. minor) side of $\bm{\omega}^T {\bf m} = 0$. If some points ${\bf m}_{ij}$ located at the minor side of the hyperplane $\bm{\omega}^T {\bf m} = 0$, we can remove the corresponding original points ${\bf a}_i$ (resp. ${\bf b}_j$) from ${\cal A}$ (resp. ${\cal B}$) to form two new sets ${\cal A}_1$ and ${\cal B}_1$, which are linearly separable. 

\begin{definition}
The set ${\rm MaxLS}({\cal A}, {\cal B})$ is said to be the maximum linearly-separable subset of ${\cal A}\cup{\cal B}$ if  
	\begin{equation*}
		{\rm MaxLS}({\cal A}, {\cal B}) :={\cal A}_\circ \cup {\cal B}_\circ=\mathop{\arg\max}_{{\cal A}' \subset {\cal A}, {\cal B}' \subset {\cal B}} |{\cal A}'|+|{\cal B}'| \quad {\rm s.t.} \quad  \mbox{${\cal A}'$ and ${\cal B}'$ are linearly separable}.
	\end{equation*}
\end{definition}
\noindent Namely, ${\rm MaxLS}({\cal A}, {\cal B})$ is the largest-size subset of ${\cal A}\cup {\cal B}$ such that ${\cal A}_\circ$ and ${\cal B}_\circ$ are linearly separable. It is noteworthy that ${\rm MaxLS}({\cal A}, {\cal B})$ could not be unique.

\begin{figure}[htbp] 
	\centering
	\includegraphics[height=4cm]{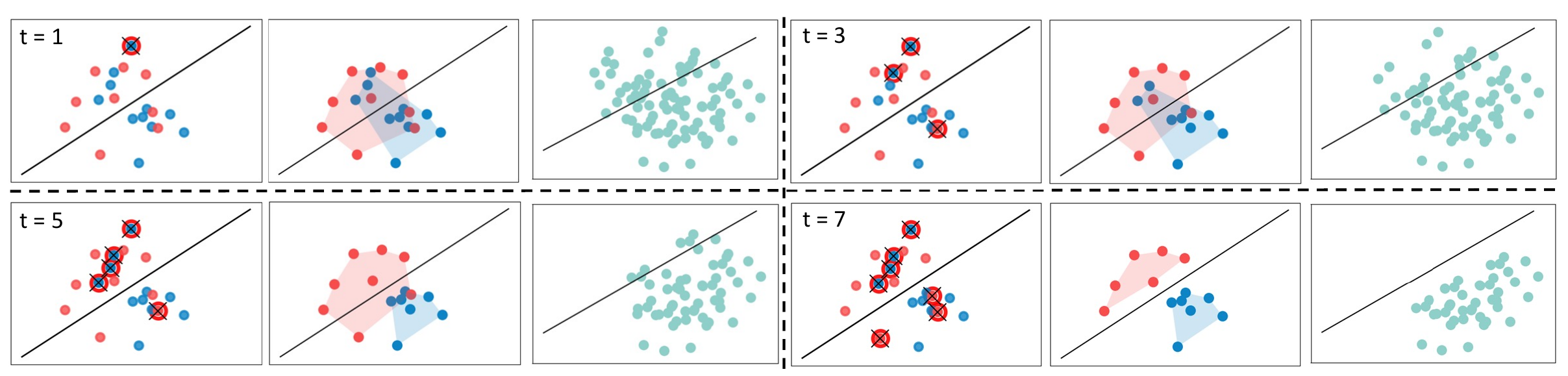}  
	\caption{The workflow of obtaining the maximum linearly separable subset. Left: two point sets with the marks that signify the one to be removed in the iterations. Middle: two convex hulls of the rest points in the two sets. Right: Minkowski difference of the rest points}\label{fig:maxls}
\end{figure}

\begin{remark}
The maximum linearly-separable subset ${\rm MaxLS}({\cal A}, {\cal B})$ of two sets ${\cal A}$ and ${\cal B}$ can be obtained in an iterative way: (1) Given $P\times Q$ points ${\bf m}_{i_1,j_1} ,\cdots, {\bf m}_{i_P,j_Q} $ lie in the minor side of $\bm{\omega}^T {\bf m} = 0$, build a undirected bipartite graph $G({\cal V}_0,{\cal E}_0)$ with ${\cal V}_0 = \{{\bf a}_{i_1},\cdots,{\bf a}_{i_P},{\bf b}_{j_1},\cdots,{\bf b}_{j_Q}\}$ and ${\cal E} = \{ ({\bf a}_{i_1},{\bf b}_{j_1}),\cdots, ({\bf a}_{i_P},{\bf b}_{j_Q})  \}$. (2) Remove one vertex $v_1$ with the largest degree from ${\cal V}_0$ and update ${\cal V}_1 =  {\cal V}_0 \setminus \{v_1\}$. (3) Eliminate the edges associated with the vertex $v_1$ and update ${\cal E}_1 \leftarrow {\cal E}_0$. (4) Repeat the steps (2)-(3) until ${\cal E}_t = \emptyset$. (5) Remove the points in ${\cal V}_0 \setminus {\cal V}_t$ from the original sets ${\cal A}$ and ${\cal B}$, and the rest form the desired ${\rm MaxLS}({\cal A}, {\cal B})$. Figure \ref{fig:maxls} exhibits a numerical example of this algorithm as well.

\end{remark}

\subsection{MD-based Linear Separability Measure (MD-LSM)}
 
Following Theorem \ref{thm:md}, the ratio of the numbers of the points $\mathbf{m}_{ij}\in \mathrm{MD}({\cal A},{\cal B})$ that respectively locate in the two sides of the hyperplane can be treated as a criterion to measure the linear separability degree between ${\cal A}$ and ${\cal B}$: 
\begin{equation}\label{eq:ls*}
	{\rm LS}_* ({\cal A},{\cal B}) := \max_{\bm{\omega}\in\mathbb{R}^N}\left\{\max \left\{  \frac{\sum\limits_{i\leq I,j\leq J} \mathbf{1}(\bm{\omega}^T  \mathbf{m}_{ij}>0)}{| \mathrm{MD}({\cal A},{\cal B}) |} , \frac{\sum\limits_{i\leq I,j\leq J} \mathbf{1}(\bm{\omega}^T  \mathbf{m}_{ij}<0)}{| \mathrm{MD}({\cal A},{\cal B}) |}    \right\} \right\},
\end{equation}
where $| \mathrm{MD}({\cal A},{\cal B}) |$ is the cardinality of $\mathrm{MD}({\cal A},{\cal B})$ and ${\bf 1}(\mathcal{E})$ is the indicator function w.r.t. the event $\mathcal{E}$. Denote ${\rm ACC}_{ \mathbf{w},{\bf b}  }({\cal A},{\cal B})$ as the classification accuracy of the linear model ${\bf y} = \langle \mathbf{w},{\bf x} \rangle +{\bf b}$ on the point sets ${\cal A} \cup {\cal B}$, and denote ${\rm ACC}_{\rm line}({\cal A},{\cal B}) : = \max_{ \mathbf{w},{\bf b} \in\mathbb{R }^N} \{{\rm ACC}_{ \mathbf{w},{\bf b}  }({\cal A},{\cal B}) \} $
as the maximum classification accuracy of all possible linear models. The following theorem illustrates the relationship between ${\rm LS}_* ({\cal A},{\cal B})$ and the maximum linearly separable subsets:
\begin{theorem}\label{thm:md-ls}
	Given two point sets ${\cal A}$ and ${\cal B}$, then   
	\begin{equation*}
		{\rm ACC}_{\rm line}({\cal A},{\cal B})=\frac{|{\cal A}_\circ| + |{\cal B}_\circ|}{|{\cal A}|+|{\cal B}|} \geq {\rm LS}_*({\cal A},{\cal B}) \geq \max \left\{  \frac{|{\cal A}_\circ|}{|{\cal A}|} , \frac{|{\cal B}_\circ|}{|{\cal B}|}  \right \}, 
	\end{equation*}
	and the equality holds if and only if the sets ${\cal A}$ and ${\cal B}$ are linearly separable. 
\end{theorem}
\noindent This result also reveals that the relationship between the linear separability and the classification accuracy of linear models. Although the classification accuracy can be used to evaluate the performance of a network, it is still challenging to track the behavior of each hidden layer during its training phase. This theorem provides a way of evaluating the linear separability of the outputs of each hidden layer during the training phase.

\subsection{Computation of MD-LSM}
 
Since the computation of ${\rm LS} _*({\cal A},{\cal B})$ is NP-hard, we consider a variant of ${\rm LS}_* ({\cal A},{\cal B})$: 
\begin{equation}\label{eq:ls0}
	{\rm LS}_0 ({\cal A},{\cal B}) := \max_{\bm{\omega}\in\mathbb{R}^N}\left\{\max \left\{  \frac{\sum\limits_{i\leq I,j\leq J} {\rm sgn}(\bm{\omega}^T  \mathbf{m}_{ij})}{| \mathrm{MD}({\cal A},{\cal B}) |} , \frac{\sum\limits_{i\leq I,j\leq J}{\rm sgn}(-\bm{\omega}^T  \mathbf{m}_{ij})}{| \mathrm{MD}({\cal A},{\cal B}) |}    \right\} \right\} ,
\end{equation} 
where ${\rm sgn}(\cdot)$ is a sign function. The following lemma demonstrates that ${\rm LS}_* ({\cal A},{\cal B})$ and ${\rm LS}_0 ({\cal A},{\cal B})$ are equivalent from the perspective of maximum linearly separable subset: 
\begin{lemma}
	Given two sets ${\cal A},{\cal B}$, let $\bm{\omega}_*$ and $\bm{\omega}_0$ be the weight vectors achieving the maximum operations in \eqref{eq:ls*} and \eqref{eq:ls0}, respectively. Then, it holds that ${\rm major}_{\bm{\omega}_*}(\mathrm{MD}({\cal A},{\cal B})) = {\rm major}_{\bm{\omega}_0}(\mathrm{MD}({\cal A},{\cal B}))$.
\end{lemma}
\noindent The proof of this lemma is direct, so we omit it here. It shows that the points lying in the major sides of the two hyperplanes $\bm{\omega}_*^T {\bf m} =0$ and $\bm{\omega}_0^T {\bf m} =0$ are the same. Unfortunately, it is still NP-hard to solve the optimization problem \eqref{eq:ls0}. Instead, we consider the following term: 
\begin{equation}\label{eq:ls1}
	{\rm LS}_1 ({\cal A},{\cal B}) :=  \max_{\bm{\omega}\in\mathbb{R}^N}\left\{ { \left | \sum\limits_{i\leq I,j\leq J} \bm{\omega}^T  \mathbf{m}_{ij} \right| } \Big/  { \sum\limits_{i\leq I,j\leq J}\left |  \bm{\omega}^T  \mathbf{m}_{ij} \right|} \right\}.
\end{equation}
The numerator $|\sum_{i,j} \bm{\omega}^T  \mathbf{m}_{ij}|$ is the absolute value of the sum of the directed distances from the points of $\mathrm{MD}({\cal A},{\cal B}) $ to the hyperplane $\bm{\omega}^T \mathbf{m} = 0$. If all points of $\mathrm{MD}({\cal A},{\cal B}) $ locate in one side of $\bm{\omega}^T \mathbf{m} = 0$, {\it i.e.,} the two sets ${\cal A},{\cal B}$ are linearly separable, it holds that ${\rm LS}_1 ({\cal A},{\cal B}) =1$. In contrast, if the value of ${\rm LS}_1 ({\cal A},{\cal B})$ is close to zero, the convex hulls of the two sets ${\cal A},{\cal B}$ overlap heavily.

Because of the existence of absolute value operation, it is still difficult to solve the optimization problem \eqref{eq:ls1}. Thus, we consider its alternative version:
\begin{equation}\label{eq:ls2}
	{\rm LS}_2({\cal A},{\cal B}) :=  \max_{\bm{\omega}\in\mathbb{R}^N}\left\{ { \left ( \sum\limits_{i\leq I,j\leq J} \bm{\omega}^T  \mathbf{m}_{ij} \right)^2 } \;  \Big/ \;  {  \sum\limits_{i\leq I,j\leq J}\big ( \bm{\omega}^T  \mathbf{m}_{ij}\big)^2 } \right\}.
\end{equation}
Let $\widetilde{\mathbf{m}}:= \sum_{i\leq I,j\leq J} \mathbf{m}_{ij} $ be the sum of all points of $\mathrm{MD}({\cal A},{\cal B}) $ and define 
$${\bf M} = [\mathbf{m}_{11} , \cdots,\mathbf{m}_{1J},\cdots, \mathbf{m}_{i1},\cdots,\mathbf{m}_{iJ},\cdots, \mathbf{m}_{I1}, \cdots,\mathbf{m}_{IJ}]_{N\times IJ}$$ 
as the matrix with the points of $\mathrm{MD}({\cal A},{\cal B}) $ being its column vectors. Then, the optimization problem \eqref{eq:ls2} can be rewritten as 
 \begin{align}\label{eq:ls2-rewrite1}
	{\rm LS}_2({\cal A},{\cal B})= \max_{\bm{\omega}\in\mathbb{R}^N}\left\{ \frac{  \bm{\omega}^T \widetilde{\mathbf{m}} \widetilde{\mathbf{m}}^T   \bm{\omega} }{  \bm{\omega}^T \mathbf{M}\mathbf{M}^T \bm{\omega} } \right\}, 
	  \end{align}
which has an equivalent form:  
\begin{align}\label{eq:ls2-rewrite2}
	&  \max_{\bm{\omega}\in\mathbb{R}^N}   \;\; \bm{\omega}^T \widetilde{\mathbf{m}} \widetilde{\mathbf{m}}^T   \bm{\omega} \quad   \mbox{s.t.} \quad  \bm{\omega}^T \mathbf{M}\mathbf{M}^T \bm{\omega} =1 \quad   \Longleftrightarrow  \quad \max_{\overline{\bm{\omega}}\in\mathbb{R}^N}   \;\; \overline{\bm{\omega}}^T  \bm{\Sigma} \overline{\bm{\omega}} \quad    \mbox{s.t.} \quad  \overline{\bm{\omega}}^T \overline{\bm{\omega}} =1,
\end{align}
where $\bm{\Sigma} :=  ( \mathbf{M}\mathbf{M}^T)^{-\frac{1}{2}} \widetilde{\mathbf{m}} \widetilde{\mathbf{m}}^T ( \mathbf{M}\mathbf{M}^T)^{-\frac{1}{2}}$ and $\overline{\bm{\omega}} : = ( \mathbf{M}\mathbf{M}^T)^{1/2}\bm{\omega} $. Then, ${\rm LS}_2({\cal A},{\cal B})$ should be the largest eigenvalue of $\bm{\Sigma}$ and the solution $\widehat{\bm{\omega}}= ( \mathbf{M}\mathbf{M}^T)^{-\frac{1}{2}}  \overline{\bm{\omega}}^\dag $, where $\overline{\bm{\omega}}^\dag$ is the eigenvector associated with the largest eigenvalue of  $\bm{\Sigma}$. Then, we obtain a closed form of the solution to \eqref{eq:ls2}. Furthermore, the resultant $\widehat{\bm{\omega}}$ will be substituted into Eqs. \eqref{eq:ls*}--\eqref{eq:ls1} to achieve the approximate calculation of the MD-LSMs ${\rm LS}_*$, ${\rm LS}_0$ and ${\rm LS}_1$, respectively. We note that, in the case of large sample size or high network width or both of them, it is still time-consuming to perform the eigenvalue decomposition on the hidden-layer outputs ${\rm hid}({\cal X})$ after each time of updating network weights. Thus, we adopt a more efficient way to obtain the approximate values of ${\rm LS}_*$, ${\rm LS}_0$, ${\rm LS}_1$ and ${\rm LS}_2$ in the subsequent numerical experiments ({\it cf.} Remark \ref{rem:approx}).

In addition, we present a theorem to show that the linear transformation cannot effect the linear separability of original point sets:  
\begin{theorem}[Linear Transfomation]\label{thm:linear.map}
	Let ${\cal A}=\left\{ \mathbf{a}_1,\cdots,\mathbf{a}_I \right\} \subset \mathbb{R}^N$ and ${\cal B}=\left\{ \mathbf{b}_1,\cdots,\mathbf{b}_J \right\} \subset \mathbb{R}^N$ be two point sets. Denote ${\bf V} = [{\bf v}_1,\cdots,{\bf v}_H]_{N\times H}$ as the weight matrix, where ${\bf v}_h = (v_{h1},\cdots,v_{hN})^T \in \mathbb{R}^N$. Define ${\bf V}({\cal A}) =\{{\bf V}^T \mathbf{a}_i: 1 \leq i\leq I   \}$ and ${\bf V}({\cal B}) =\{{\bf V}^T \mathbf{b}_j: 1 \leq j\leq J\}$. Then, it holds that ${\rm LS}_i({\cal A},{\cal B}) = {\rm LS}_i({\bf V}({\cal A}),{\bf V}({\cal B}))$ ($ i = *,0,1,2$).
\end{theorem}
\noindent This theorem demonstrates that the linear transformation ${\bf V}({\bf x}) = {\bf V}^T {\bf x}$ cannot change the linear separability degree of the original point sets. Therefore, it is necessary to equip the nodes of hidden layers with non-linear activation functions. At the end of this section, we define the MD-LSMs for multiple point sets:
\begin{definition}[MD-LSMs for Multi-class Classification]\label{def:multiple}
	Given $S$ point sets ${\cal A}_1,\cdots,{\cal A}_S$, denote ${\cal A}^c_s = \bigcup_{t  \in \{1,\cdots,S\} \setminus \{s\}} {\cal A}_t $. Then, the MD-LSM for the $S$ points sets is defined as:
	\begin{equation*}
		{\rm MultiLS}_i\big ({\cal A}_1,\cdots,{\cal A}_S\big)= \sum_{s=1}^{S} \frac{|{\cal A}_s|\cdot  {\rm LS}_i({\cal A}_s,{\cal A}^c_s)}{ \sum_{s=1}^S\big|{\cal A}_s\big|   }, \quad  i = *,0,1,2. 
	\end{equation*}
\end{definition}
\noindent In the one-vs-rest (OvR) way, we break down an $S$-class classification task into $S$ binary classification tasks and then compute the individual ${\rm LS}_i({\cal A}_s,{\cal A}^c_s)$ of each task. Then, the MD-LSM ${\rm MultiLS}_i\big ({\cal A}_1,\cdots,{\cal A}_S\big)$ of the $S$-class sample sets is expressed as a sum of ${\rm LS}_i({\cal A}_s,{\cal A}^c_s)$ weighted by the ratio of the size of ${\cal A}_s$ to the size of all samples.


\section{Effects of Activation Function and Network Size}\label{sec:parameter}
 
In this section, we study the effects of activation function and network size on the linear separability degree of hidden layer outputs. 
 
\subsection{Activation Functions}
 
\begin{figure}[htbp]
	\centering
	 
	\subfigure[Sigmoid]{
		\includegraphics[width=0.22\textwidth]{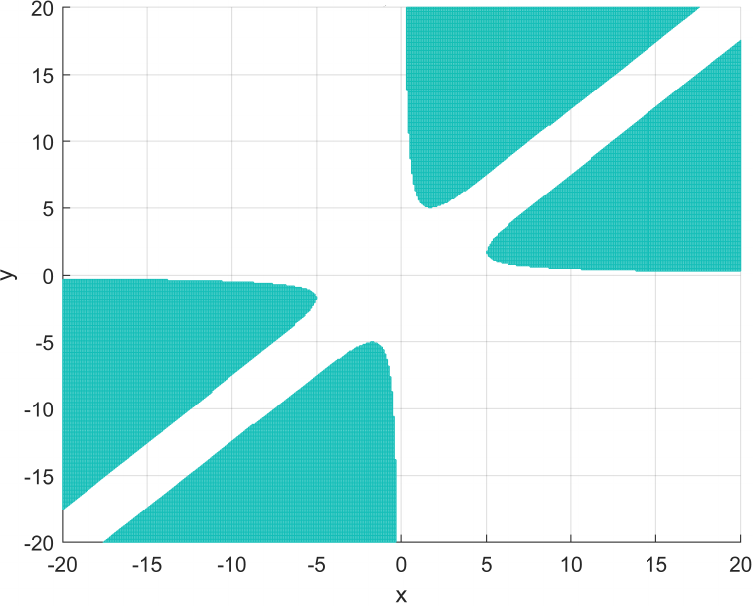}
	}	
	\subfigure[Tanh]{
		\includegraphics[width=0.22\textwidth]{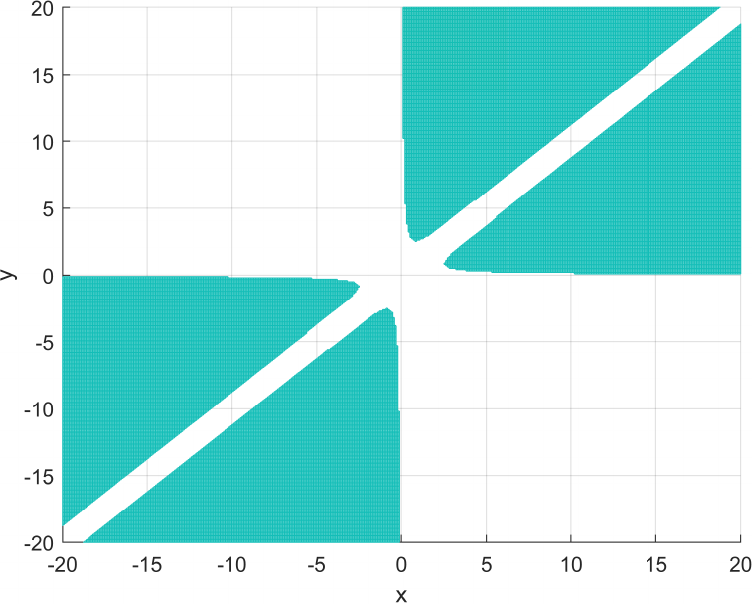}
	}	
	\subfigure[Arctan]{
		\includegraphics[width=0.22\textwidth]{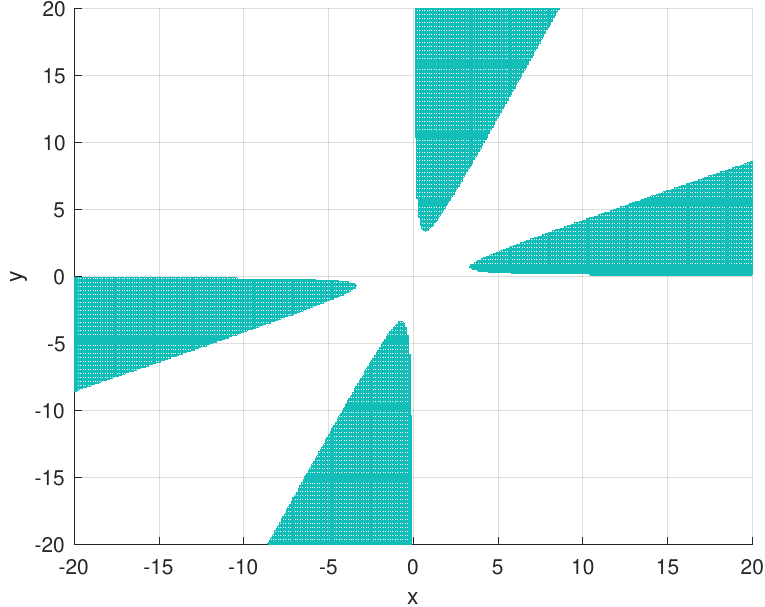}
	}	
	\subfigure[Softsign]{
		\includegraphics[width=0.22\textwidth]{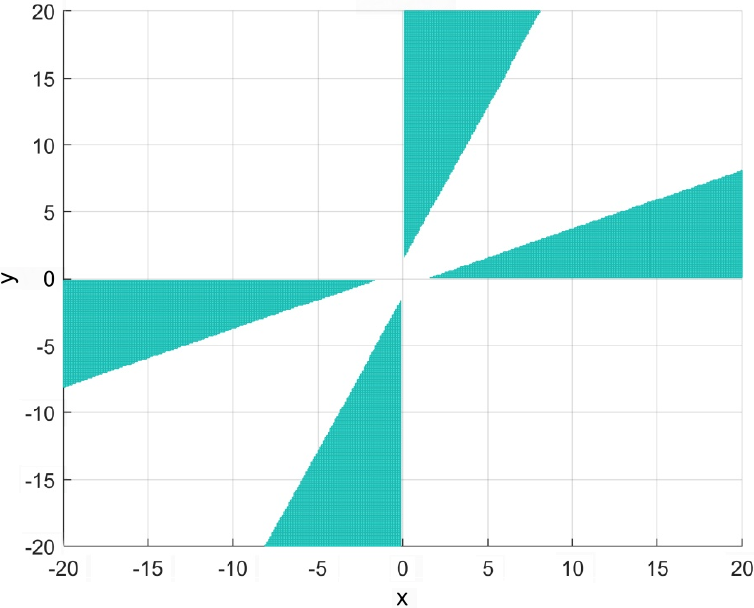}
	}	 
	\caption{The ranges of $F_\sigma(x,y)$ for different activation functions.}
	\label{fig:activation} 
\end{figure}

Consider a hidden layer of a neural network that contains $H$ hidden nodes, and let $\sigma: \mathbb{R} \rightarrow \mathbb{R}$ be an activation function. Given an input $\mathbf{x} = (x_1,\cdots,x_N)^T \in \mathbb{R}^N$, denote ${\bf V} = [{\bf v}_1,\cdots,{\bf v}_H]_{N\times H}$ as the weight matrix, where ${\bf v}_h = (v_{h1},\cdots,v_{hN})^T \in \mathbb{R}^N$ is the weight vector between the input and the $h$-th hidden node. Then, the hidden layer of neural networks can be regarded as a pseudo-linear mapping (PLM) in the following form: 
\begin{equation*}
	{\bf V}_\sigma({\bf x}): =
	\left(  \sigma(\langle {\bf v}_1, {\bf x} \rangle),\cdots, \sigma(\langle {\bf v}_H, {\bf x} \rangle) \right)^T.
\end{equation*}
Especially, if ${\bf V}$ is a random matrix, the mapping ${\bf V}_\sigma({\bf x})$ is called the random PLM accordingly. Denote ${\rm MD}( {\bf V}_\sigma({\cal A}), {\bf V}_\sigma({\cal B})) = \{{\bf n}_{ij}: = {\bf V}_\sigma({\bf a}_i) - {\bf V}_\sigma({\bf b}_j)  \;| \;  1\leq i\leq I,\; 1\leq j\leq J \}$ as the Minkowski difference of two transformed sets ${\bf V}_\sigma({\cal A})$ and ${\bf V}_\sigma({\cal B})$. The following theorem shows that PLMs have the capability of changing the linear separability of the original point sets ${\cal A}$ and ${\cal B}$.
\begin{theorem}\label{thm:change}
	Let ${\bf V}_\sigma: \mathbb{R}^N \rightarrow \mathbb{R}^H$ be a pseudo-linear mapping and $\sigma(x):\mathbb{R} \rightarrow \mathbb{R}$ be a second order derivable function with non-negative first derivative. Assume that an MD point ${\bf m}_{ij} = {\bf a}_i- {\bf b}_j$ lies in the major (resp. minor) side of the hyperplane $\bm{\omega}^T {\bf m} = 0$. Define 
\begin{equation*}
		F_\sigma(x,y) := \frac{\big[\sigma''(x) - \sigma''(y)\big] (x-y)}{\big[\sigma'(x) + \sigma'(y)\big] },\;\; (x,y)\in\mathbb{R}\times \mathbb{R}.
\end{equation*}
If the relation $F_\sigma(\langle  {\bf a}_i, {\bf v}_h \rangle,\langle  {\bf b}_j, {\bf v}_h \rangle)>2$ holds for any $h\in\{1,2,\cdots,H\}$, then the MD point ${\bf n}_{ij}= {\bf V}_\sigma({\bf a}_i) - {\bf V}_\sigma({\bf b}_j)$ lies in the minor (resp. major) side of the hyperplane $(\bm{\omega}^T{\bf V}) {\bf n} =0$. 
\end{theorem}
\noindent This result interprets that if the weights ${\bf v}_1,\cdots,{\bf v}_H$ satisfy that $F_\sigma(\langle  {\bf a}_i, {\bf v}_h \rangle,\langle  {\bf b}_j, {\bf v}_h \rangle)>2$ for any $h\in\{1,2,\cdots,H\}$, the MD point ${\bf n}_{ij}$ has a different relatively positional relation with the hyperplane $(\bm{\omega}^T{\bf V}) {\bf n} =0$. In other words, if the original MD point ${\bf m}_{ij}$ lies in the major side of $\bm{\omega}^T {\bf m} =0$, the transformed MD point ${\bf n}_{ij}$ lies in the minor side of $(\bm{\omega}^T{\bf V}) {\bf n} =0$. As shown in Fig. \ref{fig:activation}, there is a wide choice of the activation function $\sigma$, such as sigmoid, tanh, arctan and softsign. 
 
%
\subsection{Network Size}
 
By treating the multiple hidden layers as the composition of multiple PLMs, we consider the effect of the network size (including width and depth) on the linear separability. Denote 
$$ {\bf V}_\sigma^{L}({\bf a}) := \prod_{l=1}^L {\bf V}_\sigma^{(l)}({\bf a}) =  {\bf V}_\sigma^{(L)} \circ {\bf V}_\sigma^{(L-1)} \circ \cdots\circ  {\bf V}_\sigma^{(1)}({\bf a}),$$ 
where $ {\bf V}_\sigma^{(1)},\cdots,  {\bf V}_\sigma^{(L)}$ are independent random PLMs with ${\bf V}^{(l)} \in \mathbb{R}^{H_{l-1}\times H_l}$, and the operation $\circ$ stands for the composition of two PLMs. Denote $ {\bf V}_\sigma^{L}({\cal A}) = \{   {\bf V}_\sigma^{L}({\bf a}) : {\bf a}\in {\cal A}  \}$ as the set of the $L$-th hidden layer outputs w.r.t. the input set ${\cal A}$. For any $1\leq  l\leq L$, let 
$${\rm MD}( {\bf V}^{(l)}_\sigma({\cal A}), {\bf V}^{(l)}_\sigma({\cal B})) := \{{\bf n}^{(l)}_{ij}: = {\bf V}^{(l)}_\sigma({\bf a}_i) - {\bf V}^{(l)}_\sigma({\bf b}_j)  \;| \;  1\leq i\leq I,\; 1\leq j\leq J \}$$ 
be the Minkowski difference of two transformed sets ${\bf V}^{(l)}_\sigma({\cal A})$ and ${\bf V}^{(l)}_\sigma({\cal B})$. Let $\widetilde{\mathbf{n}}^{(l)} = \sum_{i\leq I,j\leq J}  \mathbf{n}^{(l)}_{ij}$ be the sum of all hidden-layer outputs and $\mathbf{N}^{(l)} = [\mathbf{n}^{(l)}_1, \cdots, \mathbf{n}^{(l)}_{IJ}]_{H_l\times (IJ)} $ be the matrix with the points of ${\rm MD}( {\bf V}^{(l)}_\sigma({\cal A}), {\bf V}^{(l)}_\sigma({\cal B}))$ being its column vectors. For any $1\leq l\leq L$, set $$\bm{\Omega}^{(l)}:=[ \mathbf{N}^{(l)}(\mathbf{N}^{(l)})^T]^{-\frac{1}{2}} \widetilde{\mathbf{n}}^{(l)} (\widetilde{\mathbf{n}}^{(l)})^T [ \mathbf{N}^{(l)}(\mathbf{N}^{(l)})^T]^{-\frac{1}{2}}.$$

 \noindent Some empirical evidences have shown that if the number of hidden nodes is larger than the dimension of the original data, the linear separability degree of the hidden layer outputs is likely to be larger than that of the original data. By using Tropp's tail inequality for random matrices \cite[Theorem 1.3]{tropp2012user}, we study the relationship between the linear separability and the number of hidden nodes:
\begin{theorem}[Network Width]\label{thm:width}
If $\mathbb{E} \bm{\Omega}^{(1)}= 0 $ and $(\bm{\Omega}^{(1)})^2  \preceq {\bf A}^2$, then   
	\begin{align}\label{eq:node}
		\mathbb{P} \{  {\rm LS}_2({\bf V}^{(1)}_\sigma({\cal A}),{\bf V}^{(1)}_\sigma({\cal B})) > {\rm LS}_2({\cal A},{\cal B})    \}\leq H \cdot {\rm e}^{ - \frac{( {\rm LS}_2({\cal A},{\cal B}))^2}{8 \sigma^2}},  
	\end{align} 
where $\sigma^2 := \| \bm{A}^2 \|_s$, ${\bf A} \preceq {\bf B}$ means that ${\bf B} - {\bf A}$ is a positive semidefinite matrix and $\|\cdot\|_s$ stands for the spectral norm.
\end{theorem}
\noindent This result addresses that with probability at most $H{\rm e}^{ - ( {\rm LS}_2({\cal A},{\cal B}))^2/8 \sigma^2}$, the value of
${\rm LS}_2({\bf V}^{(1)}_\sigma({\cal A}),{\bf V}^{(1)}_\sigma({\cal B}))$ is larger than that of ${\rm LS}_2({\cal A},{\cal B})$. Namely, if we would like to improve the linear separability degree of the original sets ${\cal A}$ and ${\cal B}$, one feasible way is to map them into a higher-dimensional space via non-linear mappings, {\it i.e.,} to increase the network width. Moreover, if the linear separability degree of ${\cal A}$ and ${\cal B}$ reaches a relatively high level, the probability that the random PLM ${\bf V}^{(1)}_\sigma(\cdot)$ can further improve their linear separability degree becomes low. %
 
\begin{corollary}[Network Depth]\label{cor:depth}
	Assume that $\mathbb{E} \bm{\Omega}^{(l)} = 0 $ and $ (\bm{\Omega}^{(l)})^2  \preceq ({\bf A}^{(l)})^2$ holds for any $1\leq l\leq L$. Then,    
	\begin{align}\label{eq:layer}
		\mathbb{P} \{  {\rm LS}_2({\bf V}^{L}_\sigma({\cal A}),{\bf V}^{L}_\sigma({\cal B})) > {\rm LS}_2({\cal A},{\cal B})    \}   \leq  LH \cdot {\rm e}^{ -  C_0^2/(8 \sigma_{0}^2)},  
	\end{align}
where $\sigma_{0}^2 = \max_l \| (\bm{A}^{(l)})^2 \|_s$ and $C_0 := \min_{l}  {\rm LS}_2({\bf V}^{(l)}_\sigma({\cal A}),{\bf V}^{(l)}_\sigma({\cal B}))  $.
\end{corollary}
\noindent This result suggests that adding more hidden layers is also a feasible way to increase the linear separability degree of the data points when they pass through a neural network.

\begin{remark}
We note that the results \eqref{eq:node} and \eqref{eq:layer} are not of the optimal forms. For example, it is expected to find a monotonically decreasing function $\rho(H)$ w.r.t. the network width $H$ such that 
\begin{align}\label{eq:desired}
\mathbb{P} \big\{  {\rm LS}_2({\bf V}^{(1)}_\sigma({\cal A}),{\bf V}^{(1)}_\sigma({\cal B})) \leq {\rm LS}_2({\cal A},{\cal B})    \big\}\leq \rho(H),  
\end{align}
which can be equivalently interpreted as follows: with probability at least $1-\rho(H)$, it holds that $ {\rm LS}_2({\bf V}^{(1)}_\sigma({\cal A}),{\bf V}^{(1)}_\sigma({\cal B})) > {\rm LS}_2({\cal A},{\cal B})$. Unfortunately, it is still technically difficult to obtain the appropriate small-ball probability inequalities for the largest eigenvalues of random matrices to achieve the probability inequality of the form \eqref{eq:desired}.
\end{remark}

\section{Numerical Experiments}\label{sec:experiment}

In this section, we conduct the experiments to examine the synchronicity between the linear separability degree of hidden layer outputs and the network training performance of several popular networks including MLP, CNN, DBN, ResNet-20, VGGNet, AlexNet, ViT and GoogLeNet-V1. The experimental results also validate the theoretical findings in Theorem \ref{thm:width} and Corollary \ref{cor:depth}. All experiments are processed in the DELL\textsuperscript{\textregistered} PowerEdge\textsuperscript{\textregistered} T640 Tower Server with two Intel\textsuperscript{\textregistered} Xeon\textsuperscript{\textregistered} 20-core processors, 128 GB RAM and a NVIDIA\textsuperscript{\textregistered} Tesla\textsuperscript{\textregistered} V100 32GB GPU.

 \begin{figure}[htbp]
	\centering
	\subfigure[MLP-5 (ReLU)]{
		\includegraphics[width=0.31\textwidth]{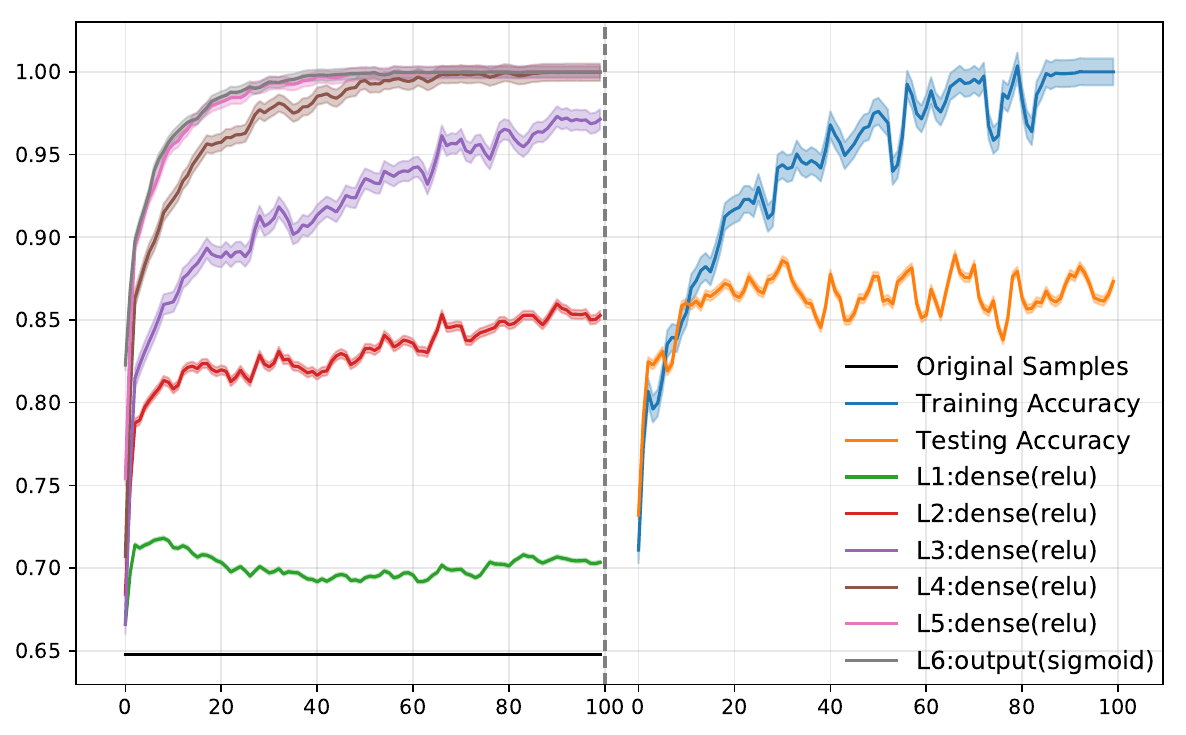}\label{fig:mlp5-relu}
	}
	\subfigure[MLP-5 (Sigmoid)]{
		\includegraphics[width=0.31\textwidth]{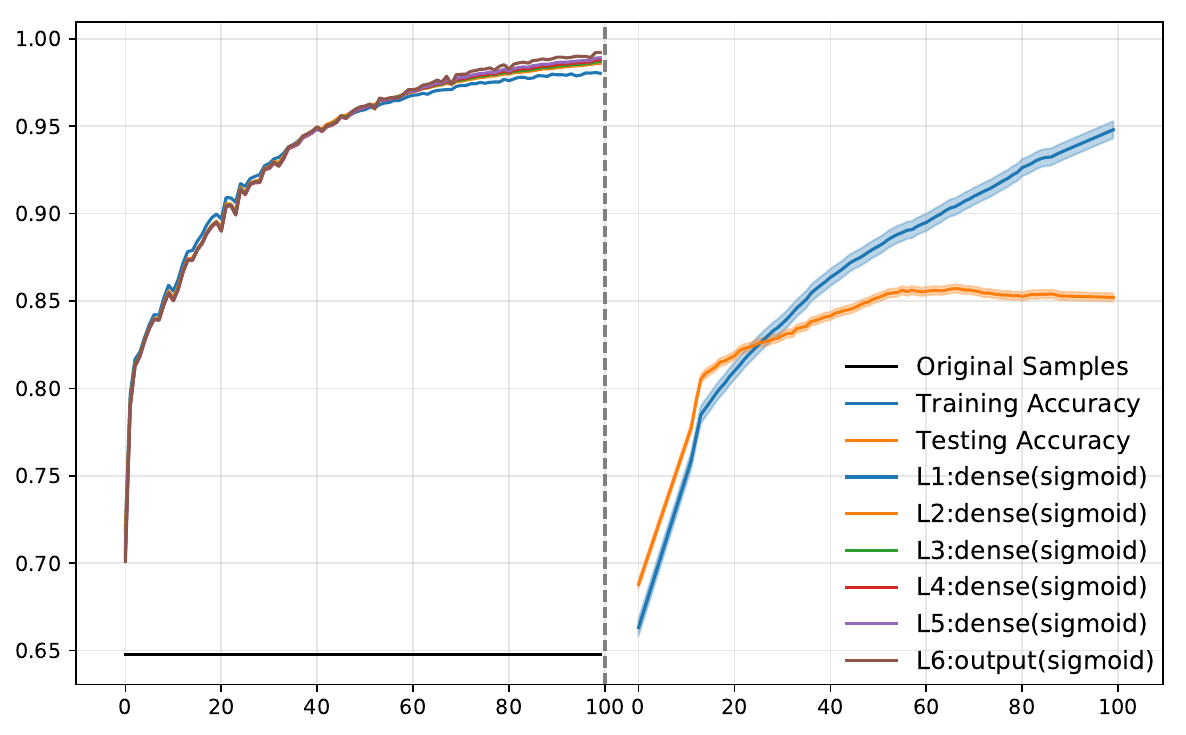}\label{fig:mlp5-sig}
	}
	\subfigure[MLP-10 (ReLU)]{
		\includegraphics[width=0.31\textwidth]{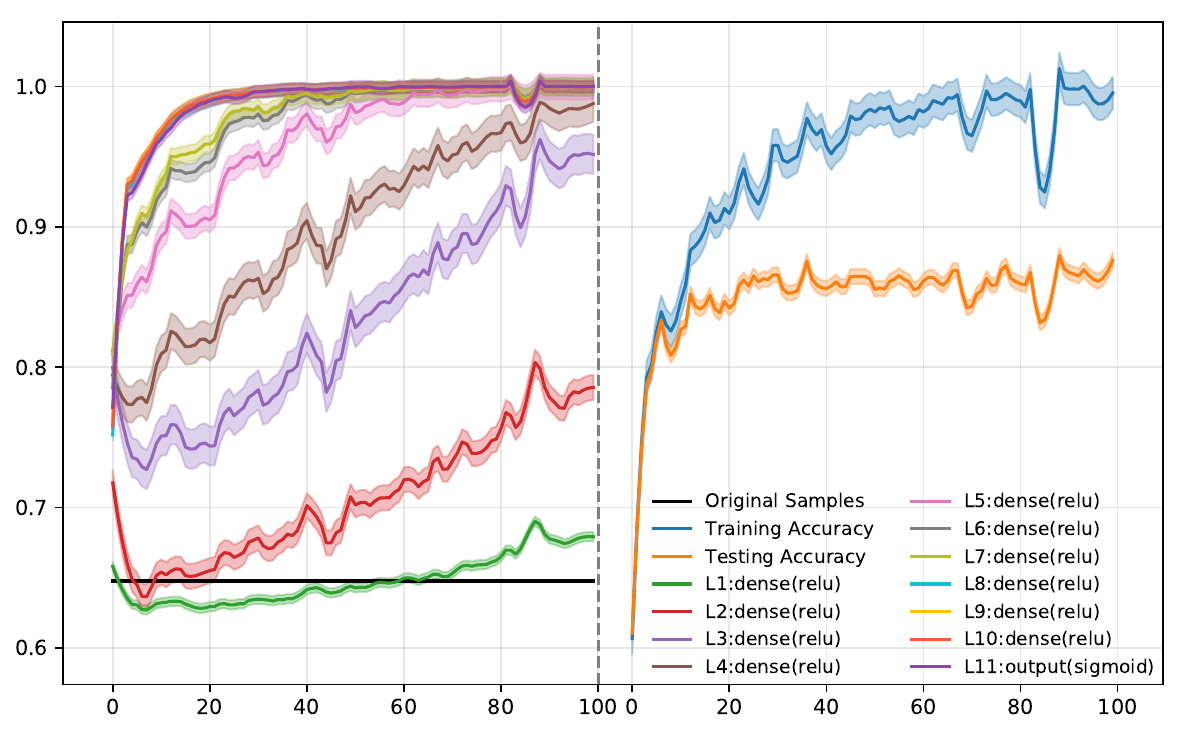}\label{fig:mlp10-relu}
	}
	\subfigure[MLP-10 (Sigmoid)]{
		\includegraphics[width=0.31\textwidth]{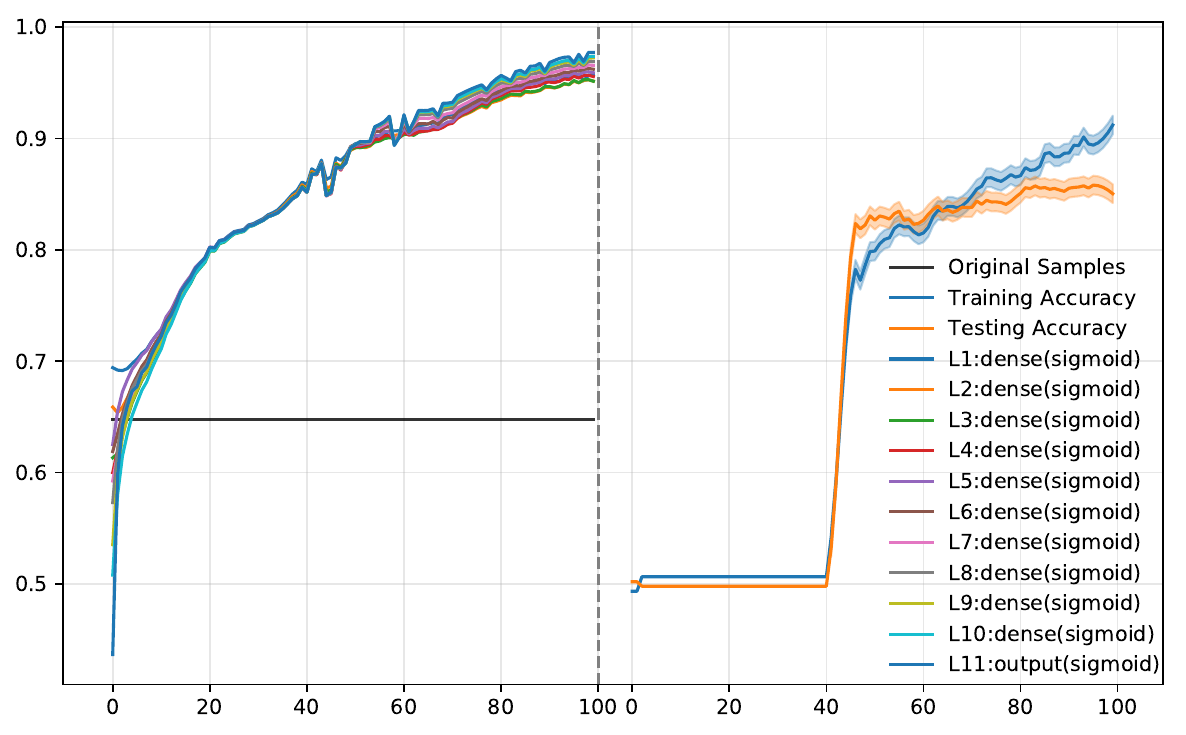}\label{fig:mlp10-sig}
	}
	\subfigure[CNN]{
		\includegraphics[width=0.31\textwidth]{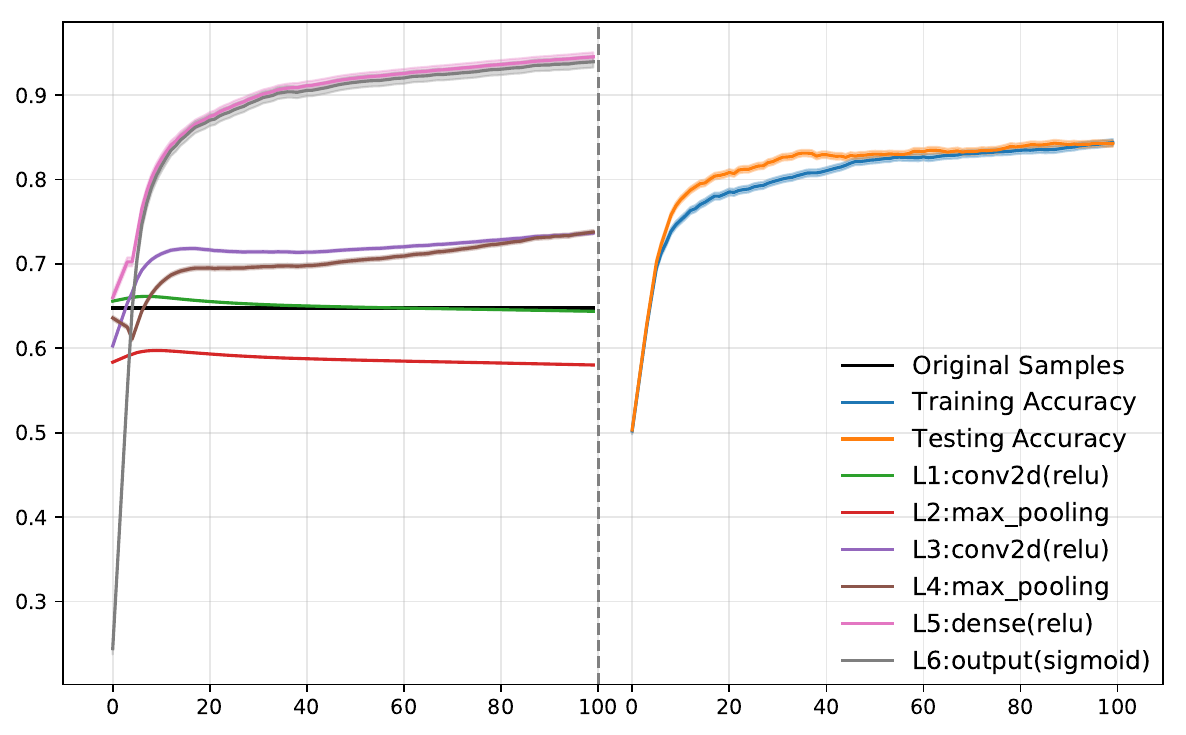}\label{fig:cnn}
	}
	\subfigure[AlexNet]{
		\includegraphics[width=0.31\textwidth]{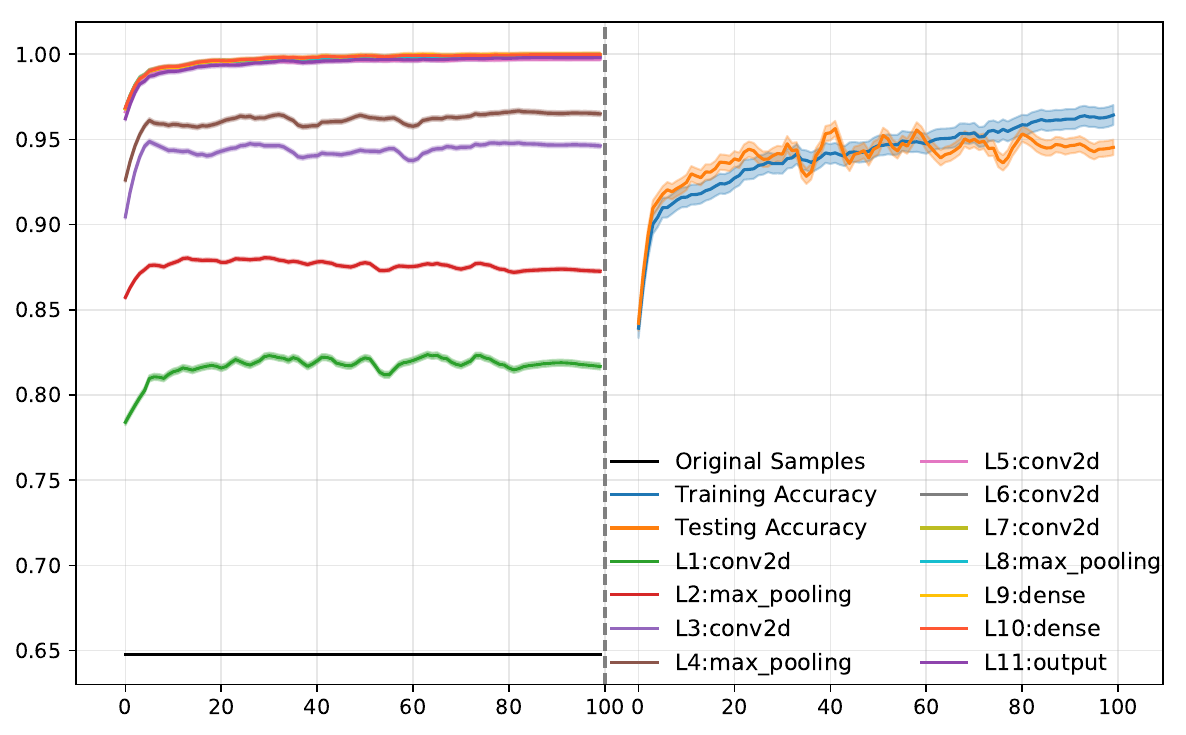}\label{fig:alexnet}
	}
	\subfigure[GoogLeNet-V1]{
		\includegraphics[width=0.31\textwidth]{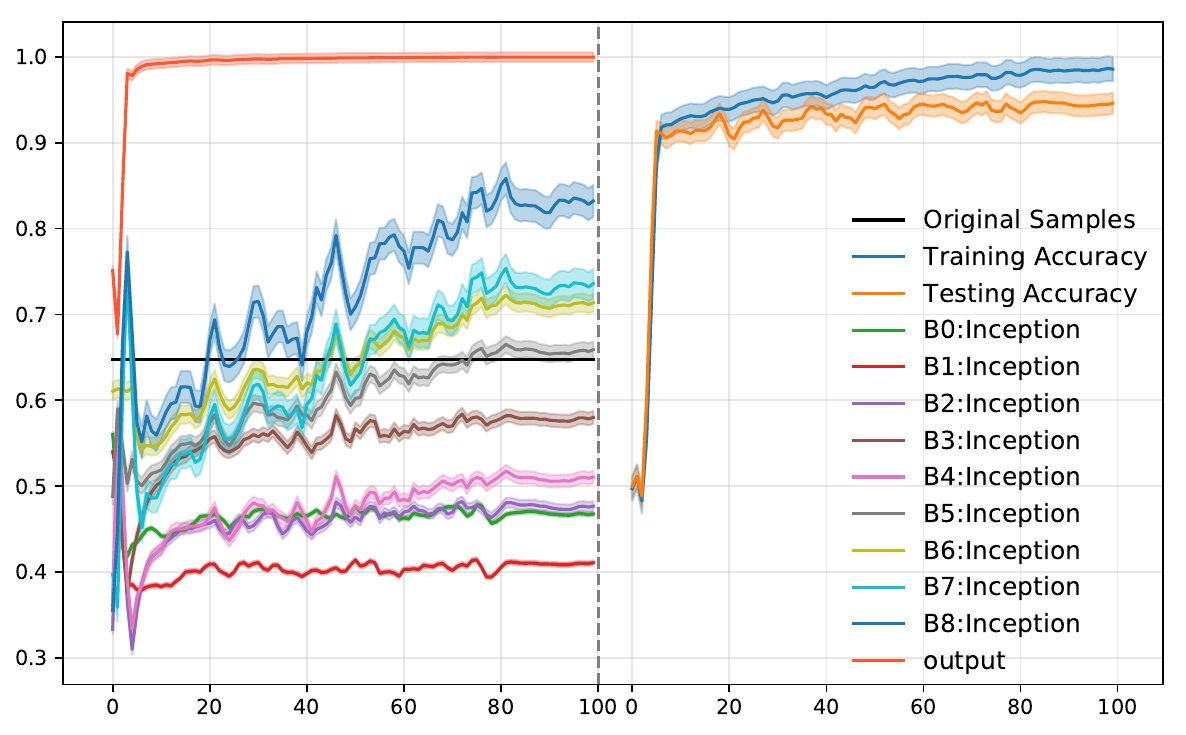}\label{fig:googlenet}
	}
	\subfigure[ResNet-20]{
		\includegraphics[width=0.31\textwidth]{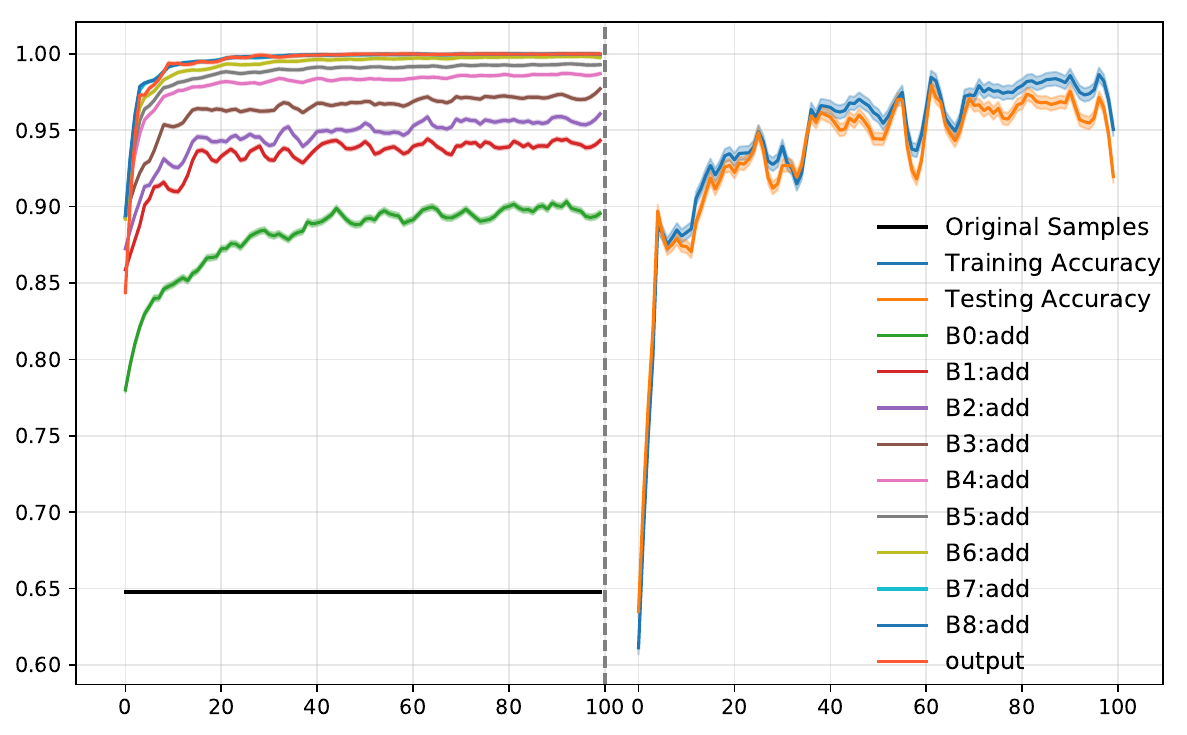}\label{fig:resnet}
	}
	\subfigure[VGGNet]{
		\includegraphics[width=0.31\textwidth]{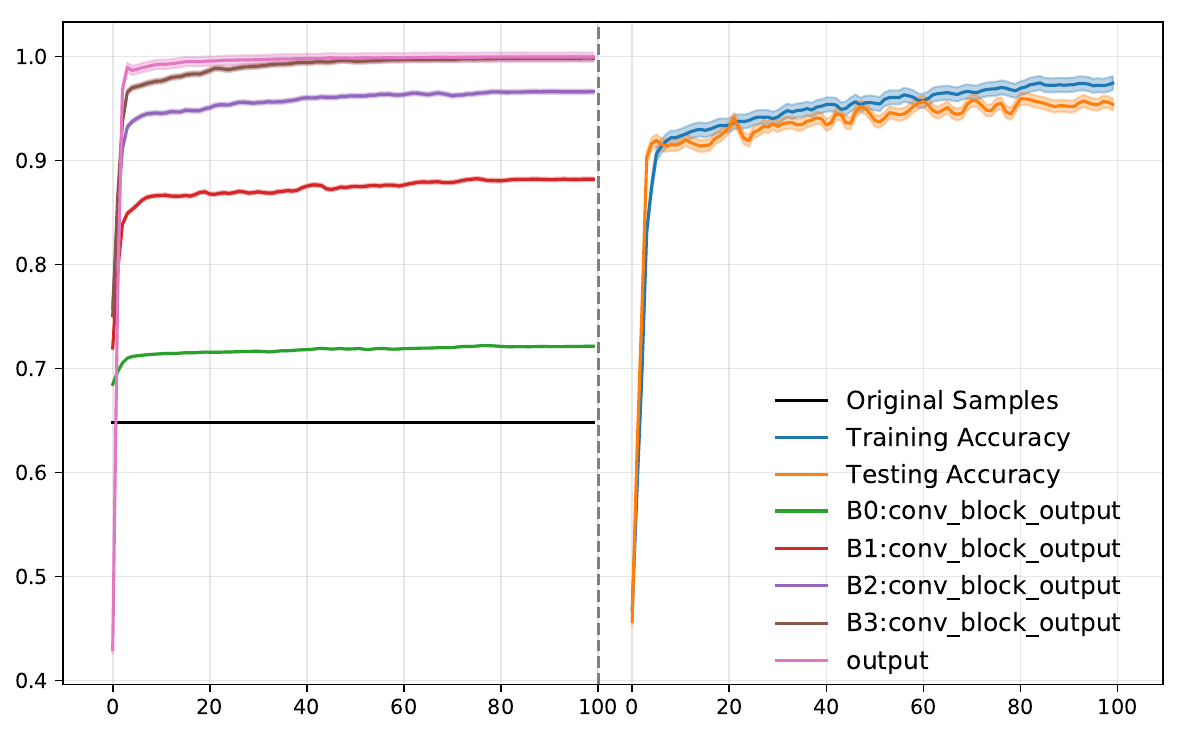}\label{fig:vggnet}
	}
	\subfigure[ViT]{
		\includegraphics[width=0.31\textwidth]{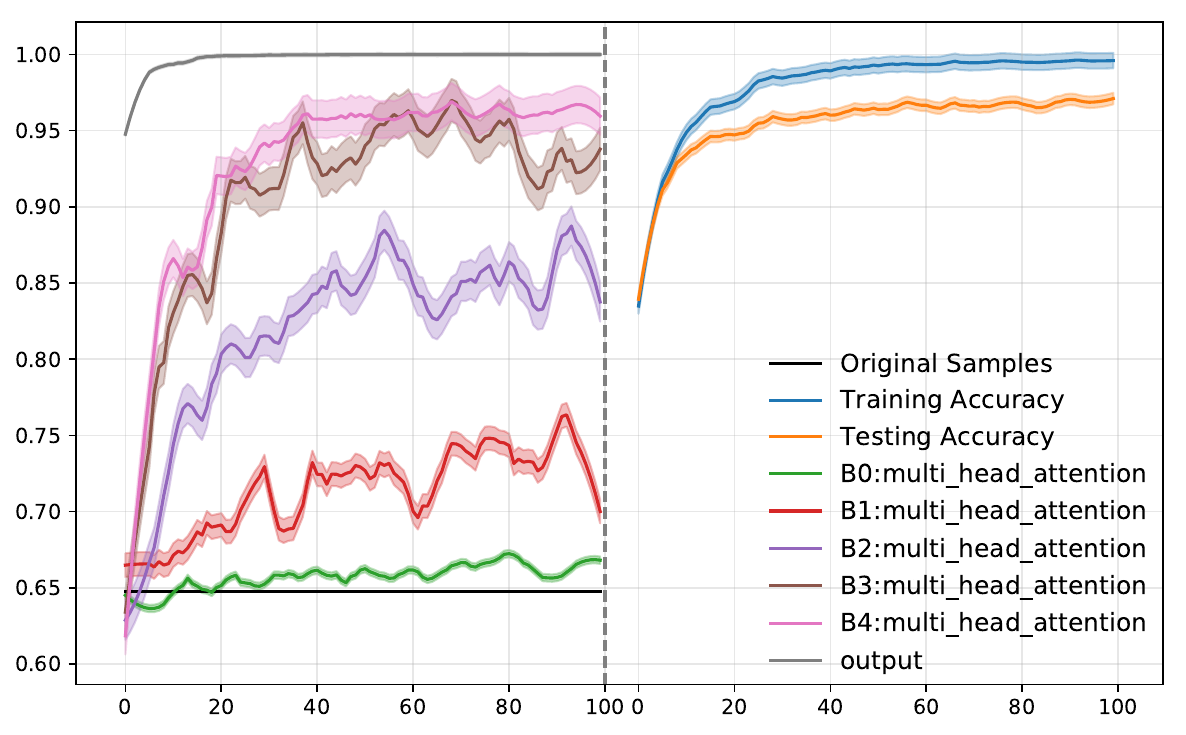}\label{fig:vit}
	}
	\subfigure[MLP (Ten-Class)]{
		\includegraphics[width=0.31\textwidth]{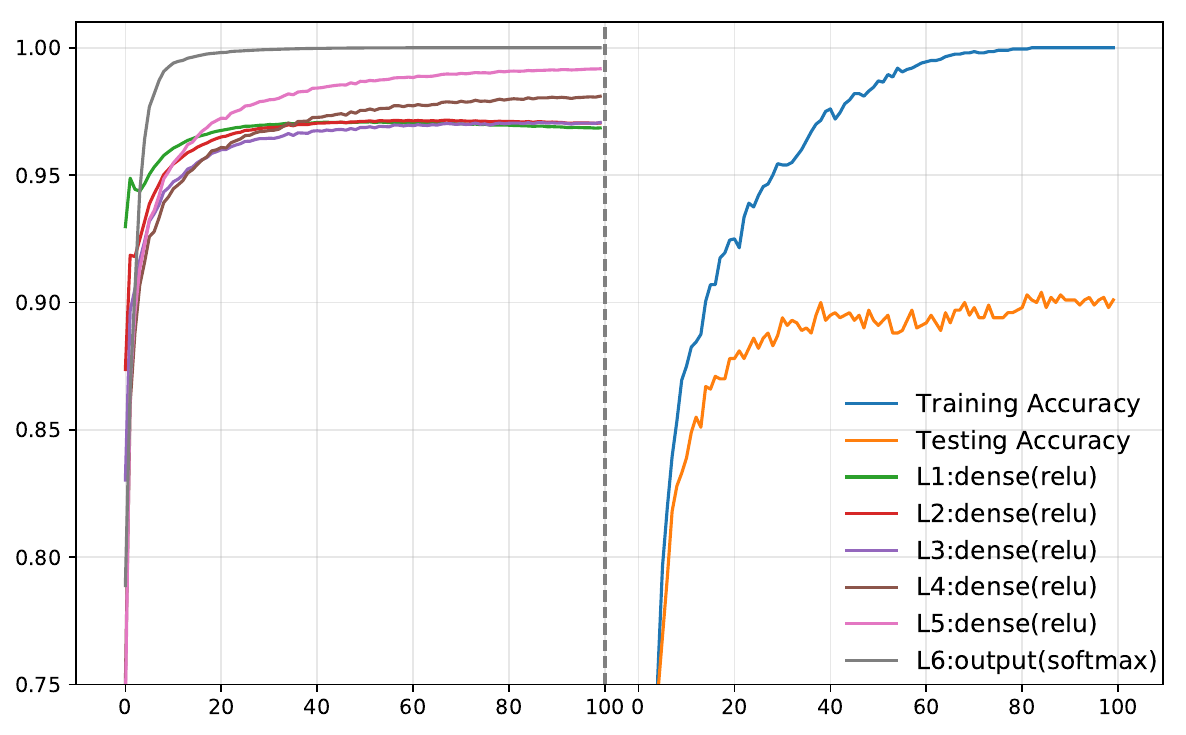}\label{fig:mlp-relu-ten}
	}
	\subfigure[DBN]{
		\includegraphics[width=0.31\textwidth]{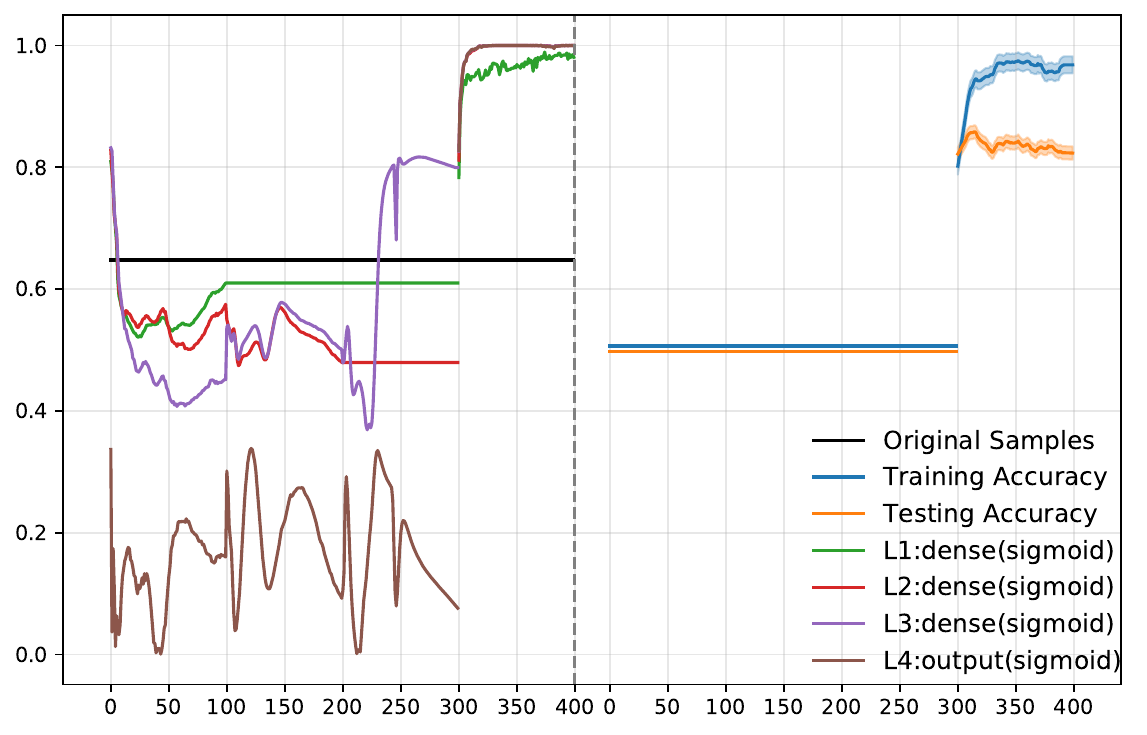}\label{fig:dbn}
	} 	
	\caption{ Synchronicity between the ${\rm LS}_1$ value (the left of each subfigure) of hidden layer outputs and training (or testing) accuracy (the right of each subfigure) of different neural networks, where ${\rm L1}$ and ${\rm B1}$ stand for the $1$st hidden layer and the $1$st block, respectively. Since the ranges of ${\rm LS}_1$, training accuracy and testing accuracy are all the interval $[0,1]$, the corresponding curves share the same $y$ label in each subfigure.} 
	\label{fig:experiment}
\end{figure}
\subsection{Experiment Setting}

Two categories (airplane and automobile) in CIFAR-10 dataset \cite{2012Learning} are selected to form the binary classification task. The SGD method with minibatch is used to updated the network weights within $100$ training epochs (including the fine-tuning phase of DBN). Since the structures of MLP, CNN and DBN are not powerful enough to obtain a good training performance by using all samples of the two categories within the limited epochs, we randomly select $2000$ (resp. $1000$) samples from the training (resp. testing) data of the two categories for training (resp. testing) the three kinds of networks. Moreover, since the network size of MLP, CNN and DBN is not large, we directly use the selected $2000$ training samples to compute the MD-LSMs of their hidden layer outputs. In contrast, we use all training (resp. testing) data of the two categories to train (resp. test) ResNet, VGGNet, AlexNet, ViT and GoogLeNet. Since the dimension of hidden layer outputs of these deep networks is high, in view of the computational burden, we randomly select $500$ samples from the two categories to compute MD-LSMs for these networks after each training epoch. 
\begin{remark}[Approximate Calculation of MD-LSMs]\label{rem:approx}
	Recall the optimization problem \eqref{eq:ls2-rewrite2} for computing ${\rm LS}_2$.  Unfortunately, it is still very time-consuming to implement the eigenvalue decomposition of $\bm{\Sigma}$ after each training epoch, especially when the dimension of $\bm{\Sigma}$ is high. Thus, in the numerical experiments, the weight vectors achieving the MD-LSMs \eqref{eq:ls*}, \eqref{eq:ls0}, \eqref{eq:ls1} and \eqref{eq:ls2} are approximately computed after each training epoch via the following optimization problem: 
	\begin{align}\label{eq:approx.weight}
		\max_{\bm{\omega}\in\mathbb{R}^N}   \;\; \bm{\omega}^T \widetilde{\mathbf{m}} \widetilde{\mathbf{m}}^T   \bm{\omega} \quad   \mbox{s.t.} \quad \bm{\omega}^T\bm{\omega} =1,
	\end{align}
whose solution is $\widehat{\bm{\omega}} = \frac{\widetilde{\mathbf{m}}}{ \|\widetilde{\mathbf{m}}\|}$. Compared with the original form \eqref{eq:ls2-rewrite2}, this approximate one actually is derived by setting $ \mathbf{M}\mathbf{M}^T = {\bf I}$.
\end{remark}

\subsection{Experimental Results and Discussion}

In Fig. \ref{fig:experiment}, we illustrate the synchronicity between linear separability of hidden layers and training performance in the classification tasks. Since ${\rm LS}_0$, ${\rm LS}_1$ and ${\rm LS}_2$ basically have the same experimental results, we only draw the ${\rm LS}_1$ curves for all hidden layers of MLP, CNN and DBN and for the main blocks of AlexNet, GoogLeNet, ResNet, VGGNet and ViT, respectively. The complete experimental report, containing ${\rm LS}_0$, ${\rm LS}_1$ and ${\rm LS}_2$ curves for all hidden layers of these networks, is arranged in the appendix (part \ref{supp:experiment}). Moreover, we also provide the detailed structures of these neural networks with the name of each hidden layer to facilitate the interpretation of experimental results. %

{\bf [Binary Classification]} As shown in Figs. \ref{fig:mlp5-relu}--\ref{fig:vit}, there is an obvious synchronicity between the ${\rm LS}_1$ curves of hidden layers and the accuracy curves: 1) when the training accuracy increases, the ${\rm LS}_1$ value of the outputs of each hidden layer (or main block) increases synchronously; 2) when some fluctuations appear in the ${\rm LS}_1$ curves, the training and the testing accuracy curves have the fluctuations occurring nearby the corresponding epochs accordingly; 3) especially for the neural networks with relatively shallow structures, such as MLP and CNN ({\it cf}. Figs. \ref{fig:mlp5-relu}--\ref{fig:cnn}), the magnitude of the fluctuations in the ${\rm LS}_1$ curves is merely proportional to that of the fluctuations in the training and the testing accuracy curves.

{\bf [Network Size]} The experimental results, given in Figs. \ref{fig:mlp5-relu}--\ref{fig:mlp-relu-ten}, also reflect two facts: 1) in most cases, the linear separability of the hidden layers (or blocks) is stronger than that of the original data after a few training epochs; and 2) the hidden layers (or blocks), which are closer to the output layer, have higher linear separability. These phenomena coincide with the theoretical findings in Theorem \ref{thm:width} and Corollary \ref{cor:depth}, which have proven that the large network size is more likely to provide high linear separability of the hidden layers.

{\bf [Multi-class Classification]} We also consider the linear separability of MLP for ten-class classification task. The experiment is conducted by using MLP to classify the MINST dataset \cite{lecun1998gradient}. We adopt the one-vs-rest (OvR) way to build ten MLPs with the same structure. After each training epoch, we compute the MD-LSMs of all hidden layer outputs of each CNN in the way mentioned in Definition \ref{def:multiple}. As shown in Fig. \ref{fig:mlp-relu-ten}, we obtain the same experiment observations as binary classification tasks and support the theoretical findings as well.

{\bf [Layer-wise Pretraining]} We use the linear separability to analyze the layer-wise pretraining strategy, implemented by using restricted Boltzmann machine (RBM), of DBN in binary classification task. During the pretraining phase, the weights of each hidden layers are updated $100$ times. After that, all weights of the network are fine-tuned in $100$ epochs by using the SGD method. RBM is an energy minimization method and the layer-wise pretraining aims to find the suitable weights such that the energy between two adjacent hidden layers decays to the minimal status. As shown in Fig. \ref{fig:dbn}, the RBM-based pretraining not only fails to increase the linear separability degree of each hidden layer but also makes it lower than the original data. However, the subsequent fine-tuning process can quickly improve the linear separability of hidden layers and meanwhile the network achieves a good performance. This phenomenon implies that the RBM-based pretraining makes each hidden layer capture useful features, which are not beneficial to increasing the linear separability degree yet. The reason is that the RBM method is independent of the label information, while the evaluation of linear separability is label-based. Interestingly, during the pretraining phase, the training and the testing accuracies are both round the value of $0.5$. This phenomenon implies that the network always maintains the largest randomness and diversity, which will benefit to the subsequent fine-tuning process.

\section{Difference between MD-LSM and Fisher LDA}\label{supp:lda}
In this section, we make a comparison between ${\rm LS}_2({\cal A},{\cal B})$ and Fisher LDA. First, we present a theorem to show their differences, and then explain why we use MD-LSM to evaluate the linear separability degree of two sets instead of Fisher LDA.

Let $\bm{\mu}_a = \frac{1}{I} \sum_{i=1}^I  {\bf a}_i $ and $\bm{\mu}_b = \frac{1}{J} \sum_{j=1}^J  {\bf b}_j $ be the centers of the sets ${\cal A}$ and ${\cal B}$, respectively. Let ${\bf A}_c$ (resp. ${\bf B}_c$) be the matrix associated with the set ${\cal A}$ (resp. ${\cal B}$) whose columns consist of the mean shifted data points:
\begin{equation*}
	{\bf A}_c :=[{\bf a}_1-\bm{\mu}_a,\cdots,{\bf a}_{I}-\bm{\mu}_a] \;\; \mbox{and}\;\;   {\bf B}_c :=[{\bf b}_1-\bm{\mu}_b,\cdots,{\bf b}_{J}-\bm{\mu}_b].
\end{equation*}
Denote ${\bf S}_w = {\bf A}_c{\bf A}_c^T  + {\bf B}_c{\bf B}_c^T$ and ${\bf S}_b= (\bm{\mu}_a-\bm{\mu}_b) (\bm{\mu}_a-\bm{\mu}_b)^T$. The objective function of Fisher LDA can also be treated as an LSM:
\begin{equation}\label{eq:ls-lda}
	J_\omega =\max_{\bm{\omega}} \frac{\bm{\omega}^T {\bf S}_b\bm{\omega}}{\bm{\omega}^T{\bf S}_w\bm{\omega}}.
\end{equation}


The following results show the difference between MD-LSM and LDA-LSM.
\begin{proposition}\label{thm:LDA}
	Given two point sets ${\cal A} = \{ {\bf a}_1,\cdots,{\bf a}_I\}$ and ${\cal B} = \{ {\bf b}_1,\cdots,{\bf b}_J\}$, then it holds that
	\begin{eqnarray*}
		I^2J^2 {\bf S}_b & = & \widetilde{\bf m}\widetilde{\bf m}^T ;\\ 
		{\bf S}_w &=& {\bf A}  {\bf A} ^T  + {\bf B}  {\bf B} ^T  - \Big(I (\widehat{\mathbb{E}} {\bf a})(\widehat{\mathbb{E}} {\bf a})^T     + J (\widehat{\mathbb{E}} {\bf b})(\widehat{\mathbb{E}} {\bf b})^T \Big);\\
		{\bf M} {\bf M}^T  
		& = & J {\bf A} {\bf A}^T  +  I {\bf B} {\bf B}^T - IJ \cdot \widehat{\mathbb{E}}\{ {\bf a}{\bf b}^T +{\bf b}{\bf a}^T  \},
	\end{eqnarray*}
	where $\widehat{\mathbb{E}}$ stands for the sample mean with
	\begin{eqnarray*}
		\widehat{\mathbb{E}}{\bf a} &= &\frac{1}{I}\sum_{i=1}^I {\bf a}_i; \\
		\widehat{\mathbb{E}}{\bf b} &= & \frac{1}{J} \sum_{j=1}^J {\bf b}_j;\\
		\widehat{\mathbb{E}}\{ {\bf a}{\bf b}^T +{\bf b}{\bf a}^T  \} & =& \frac{1}{IJ}\mathop{\sum_{1\leq i\leq I}}_{1\leq j\leq J}(  {\bf a}_i {\bf b}_j^T + {\bf b}_j {\bf a}_i^T  ) .
	\end{eqnarray*}

\end{proposition}
As demonstrated above, since ${\bf S}_w$ differs from ${\bf M} {\bf M}^T$, the hyperplane $\bm{\omega}^T {\bf m} = 0$ achieving ${\rm LS}_2({\cal A},{\cal B})$ is different from the one achieving $J_\omega$. When we use the approximate way, mentioned in Remark \ref{rem:approx}, to compute the weight for the MD-LSMs, the corresponding optimization objective function coincides with that of Fisher LDA with ${\bf S}_b = {\bf I}$ ({\it cf.} Eq. \eqref{eq:ls-lda}). In spite of the same weight vector $\bm{\omega}$ derived from the approximated form, the linear separability degree is still evaluated in different forms after substituting $\bm{\omega}$ into the expressions of MD-LSMs (including ${\rm LS}_*$, ${\rm LS}_0$, ${\rm LS}_1$ and ${\rm LS}_2$) and $J_\omega$, respectively. 

\begin{figure}[htbp]
	\centering
	\subfigure[$ {\rm LS}_0=1$, $ {\rm LS}_1=1$, $ {\rm LS}_2=2.33\times 10^{3}$, $J_\omega=0.28$]{
		\includegraphics[width=0.3\textwidth]{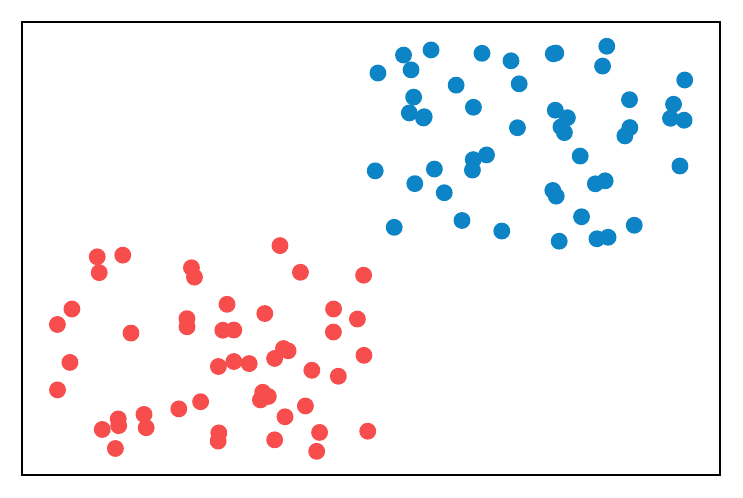}
	}\qquad
	\subfigure[$ {\rm LS}_0=0.95$, $ {\rm LS}_1=0.99$, $ {\rm LS}_2=1.96 \times 10^{3}$, $J_\omega=0.07$]{
		\includegraphics[width=0.3\textwidth]{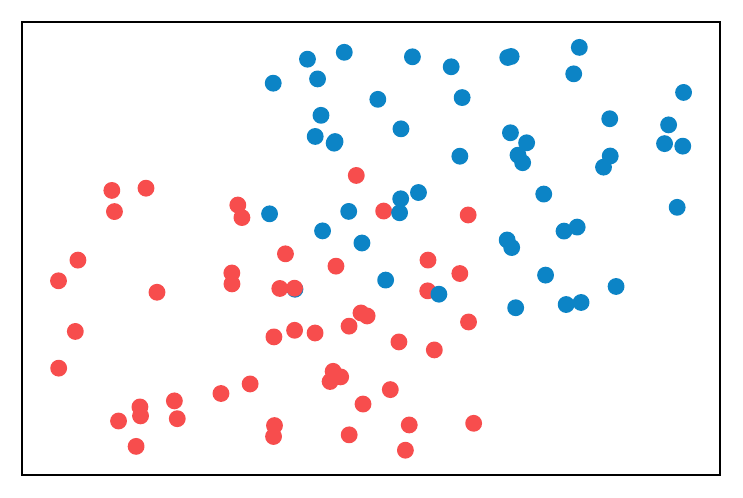}
	}
	\subfigure[$ {\rm LS}_0=0.68$, $ {\rm LS}_1=0.81$, $ {\rm LS}_2=1.21 \times 10^{3}$, $J_\omega=0.02$]{
		\includegraphics[width=0.3\textwidth]{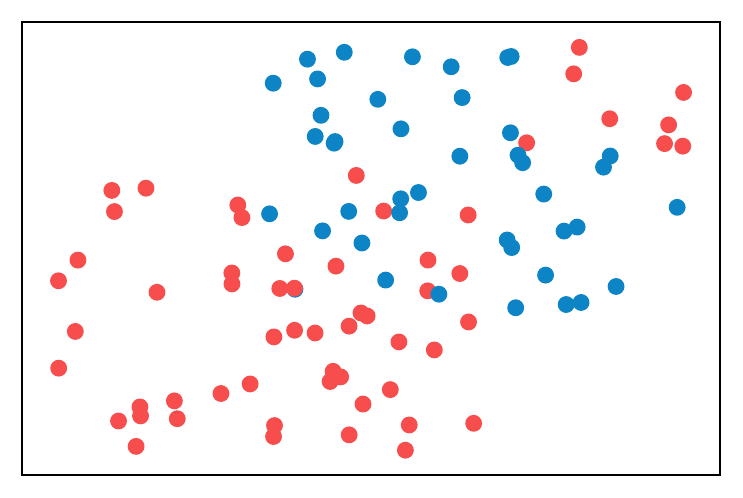}
	}\qquad
	\subfigure[$ {\rm LS}_0=0.21$, $ {\rm LS}_1=0.30$, $ {\rm LS}_2=1.53\times 10^2$, $J_\omega=1.30\times 10^{-3}$]{
		\includegraphics[width=0.3\textwidth]{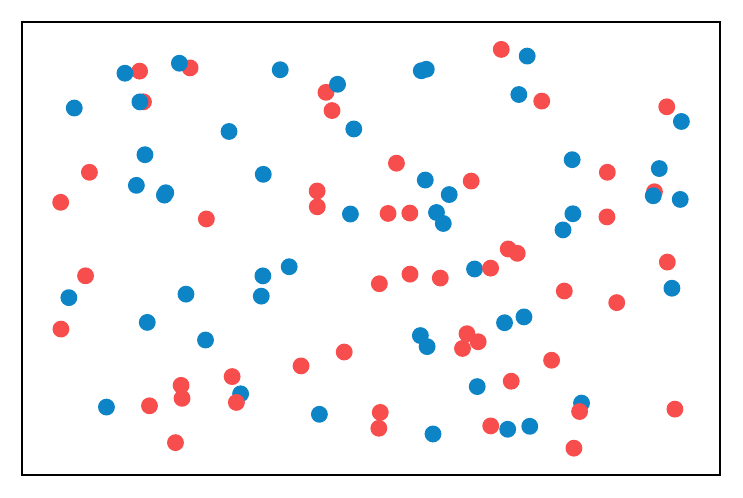}
	}
	%
	\caption{Evaluation of Linear Separability Degree of Different Point Sets Based on MD-LSMs and Fisher LDA.}
	\label{fig:lda}
\end{figure}

\begin{figure}[htbp]
\centering	
%
%
\subfigure[$ {\rm LS}_0$]{
	\includegraphics[width=0.43\textwidth]{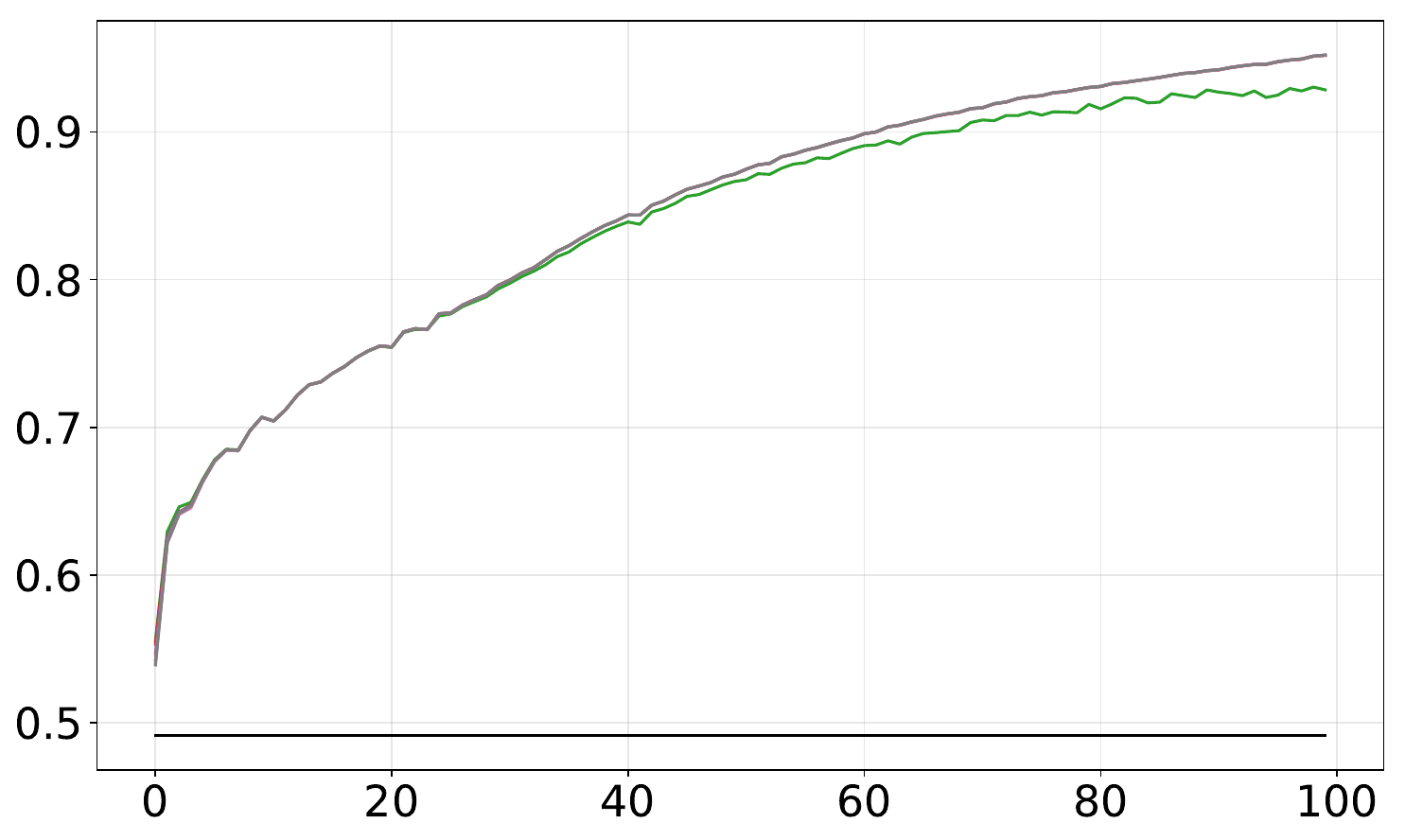}
}
\subfigure[$ {\rm LS}_1$]{
	\includegraphics[width=0.43\textwidth]{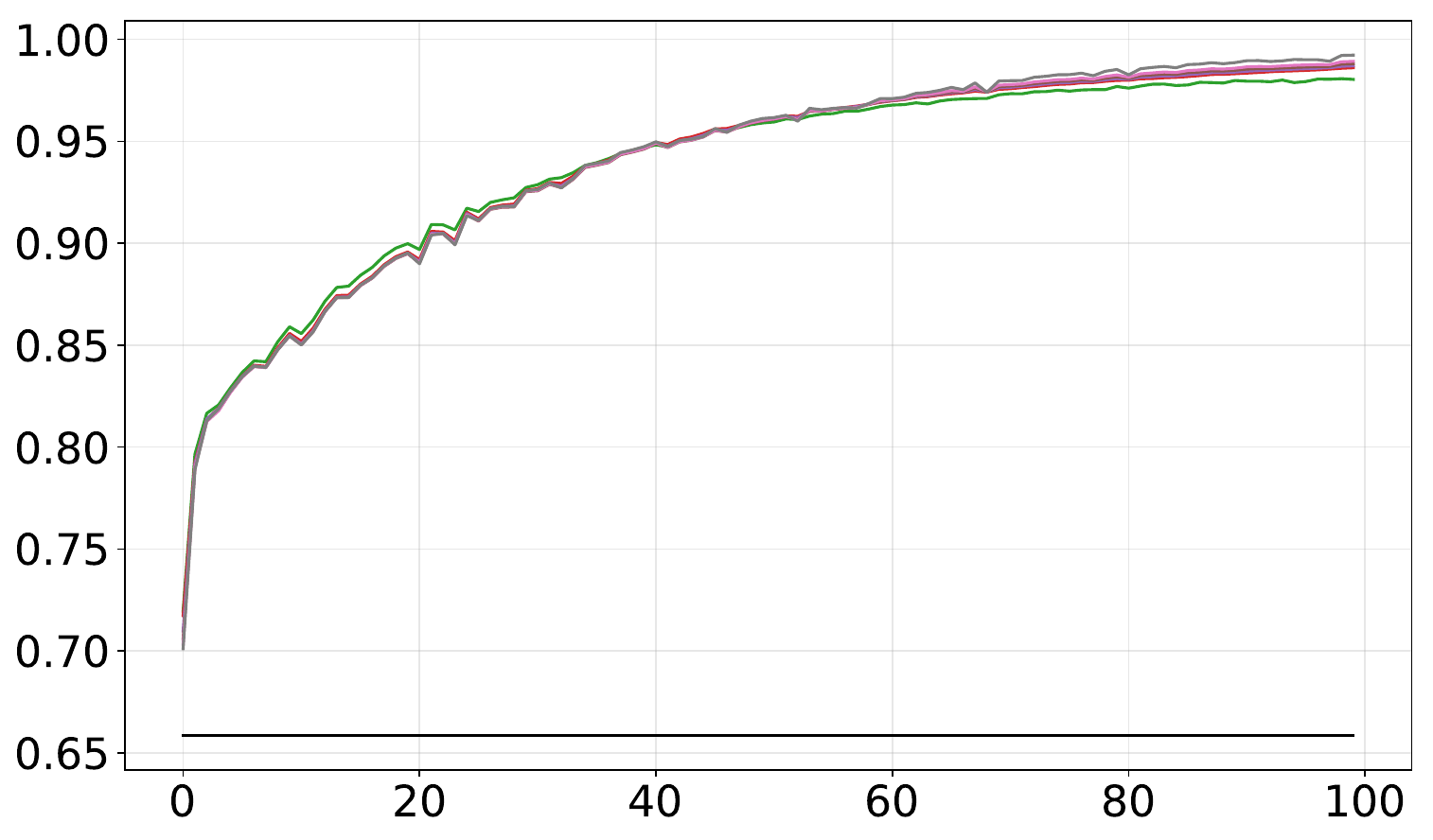}
}
\subfigure[$ {\rm LS}_2$]{
	\includegraphics[width=0.43\textwidth]{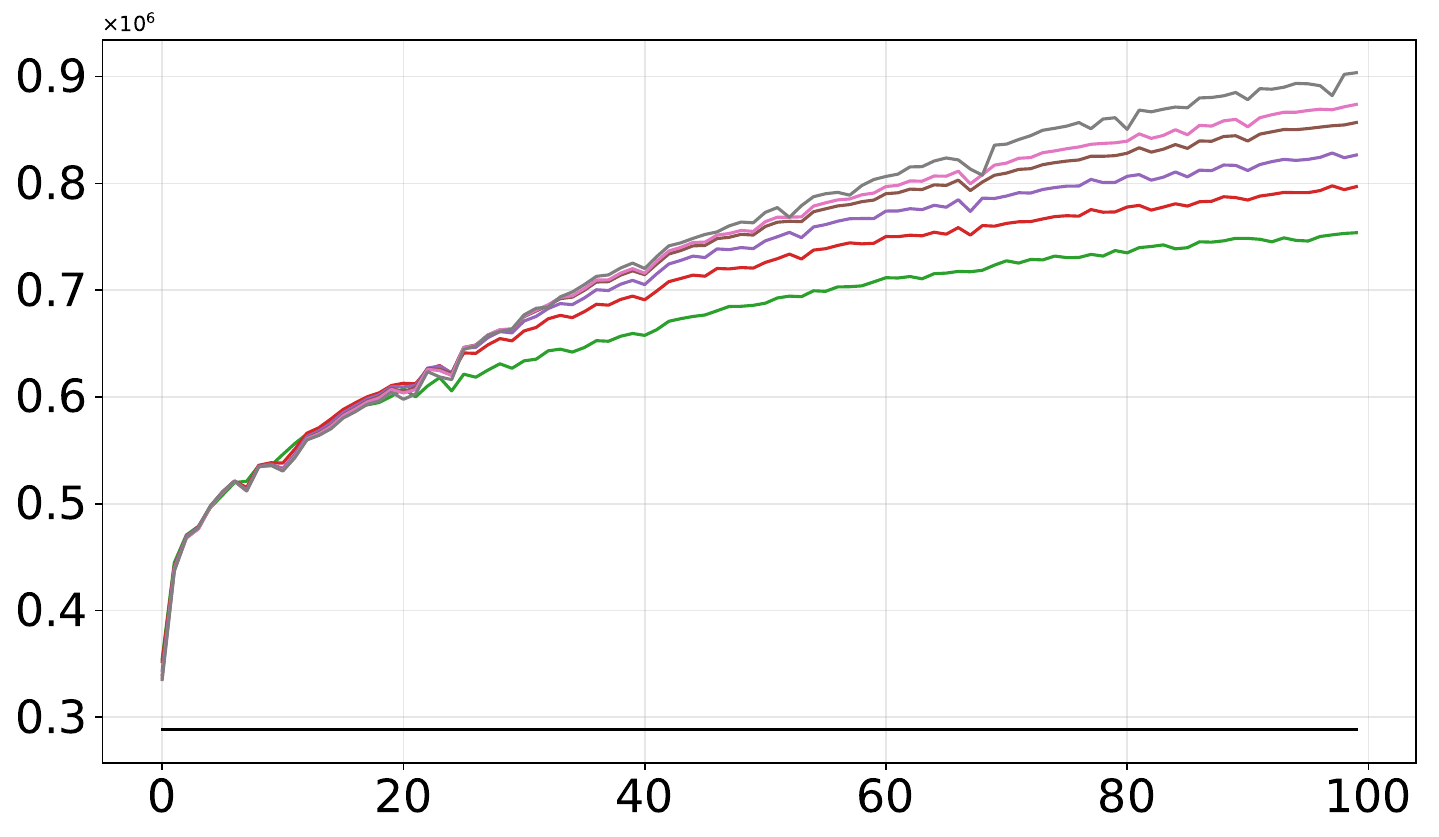}
}
\subfigure[$J_\omega$]{
	\includegraphics[width=0.43\textwidth]{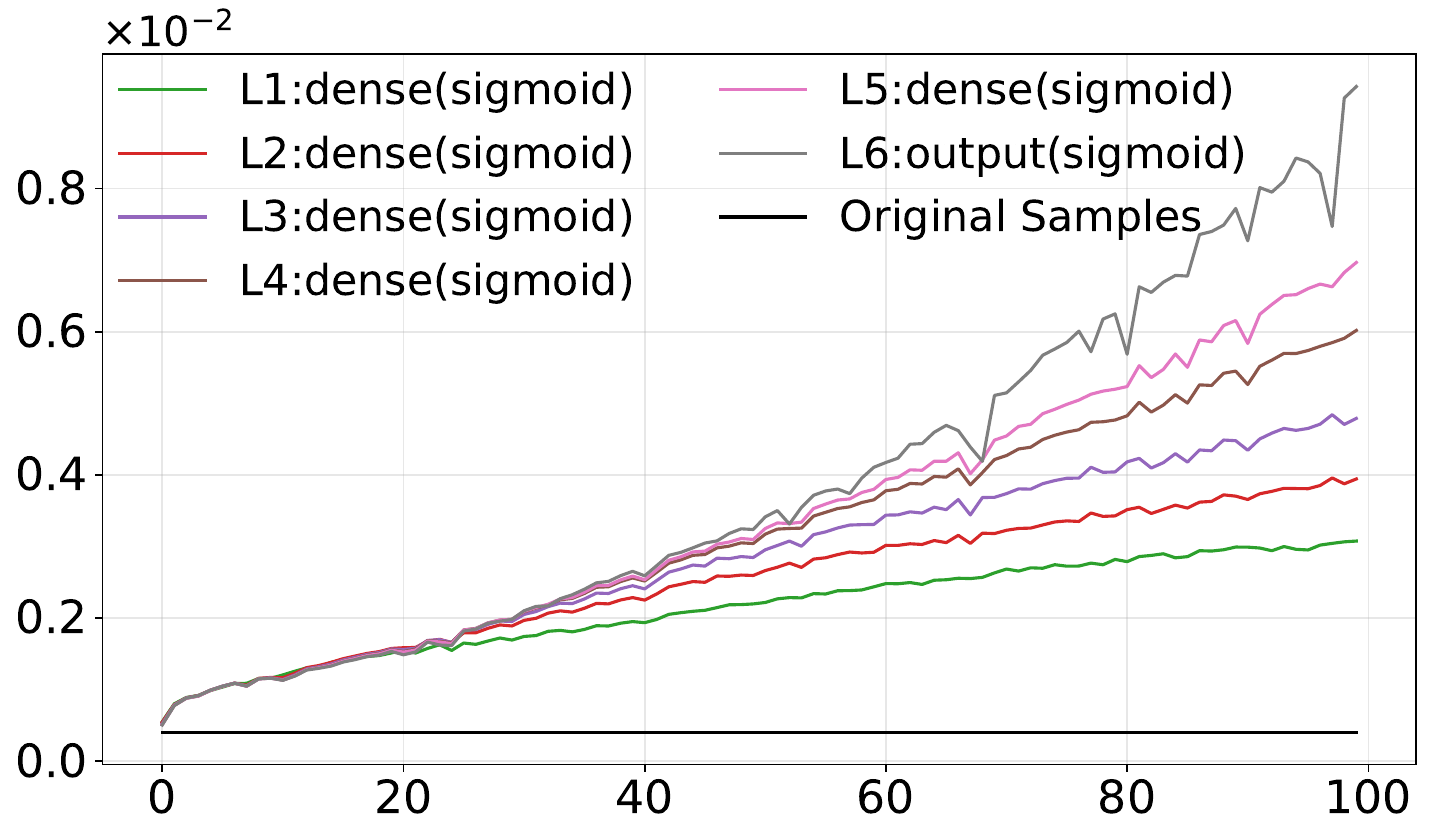}
}
\caption{Curves of MD-LSMs and $J_\omega$ for Hidden Layers of MLP-5.}
\label{fig:MLP-LDA}				
\end{figure}

\begin{figure}[htbp]
\centering	

\subfigure[$ {\rm LS}_0$]{
	\includegraphics[width=0.43\textwidth]{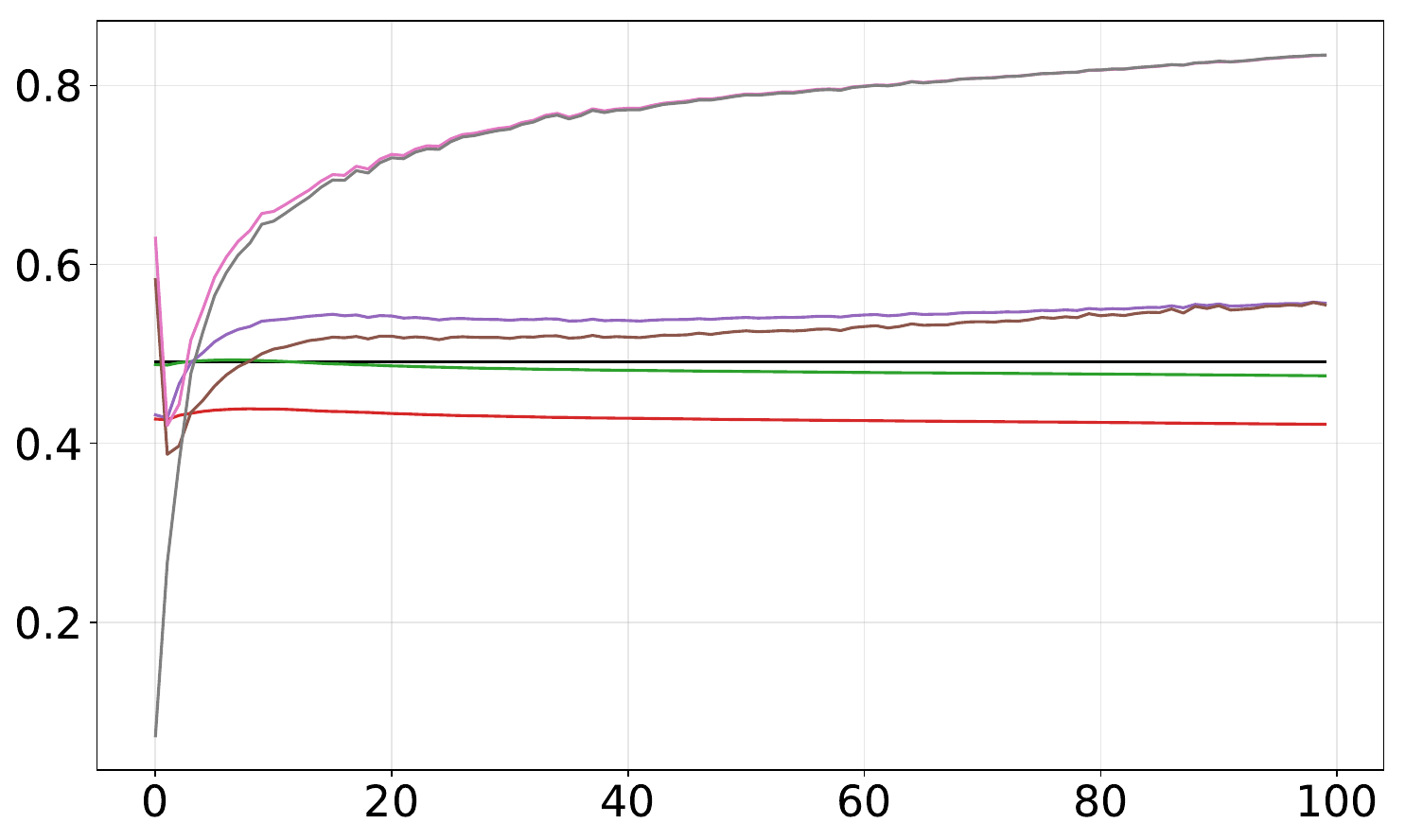}
}
\subfigure[$ {\rm LS}_1$]{
	\includegraphics[width=0.43\textwidth]{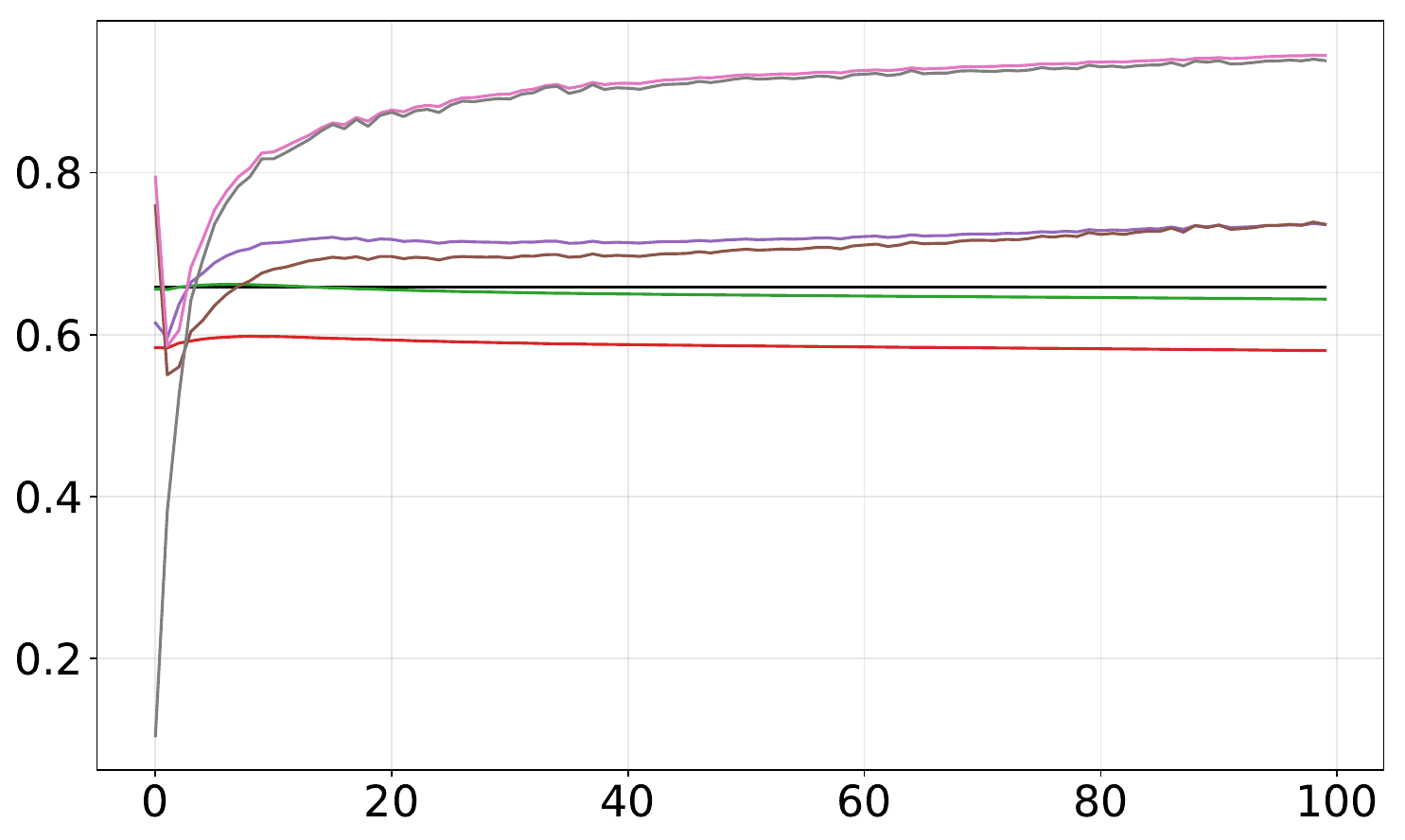}
}
\subfigure[$ {\rm LS}_2$]{
	\includegraphics[width=0.43\textwidth]{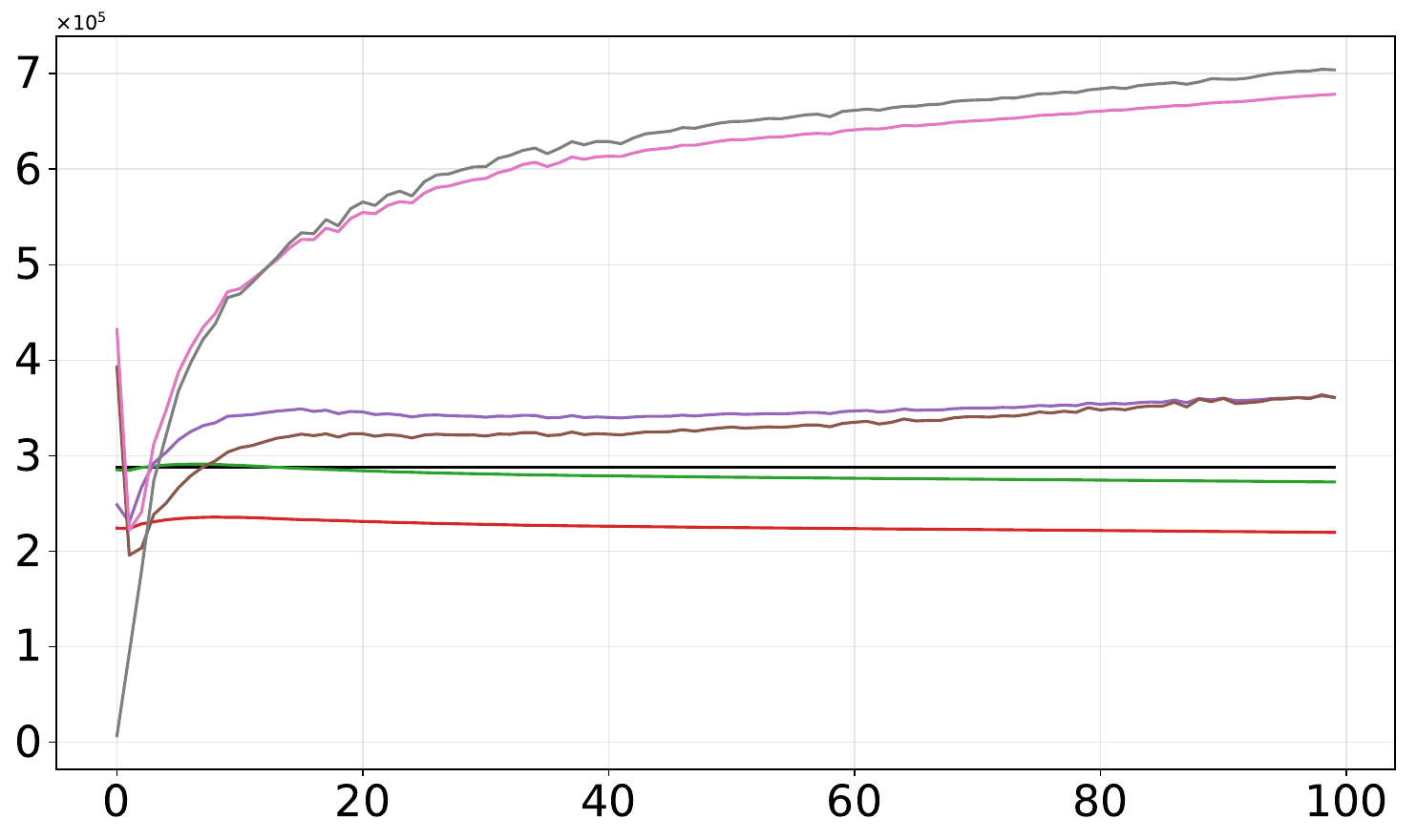}
}
\subfigure[$J_\omega$]{
	\includegraphics[width=0.43\textwidth]{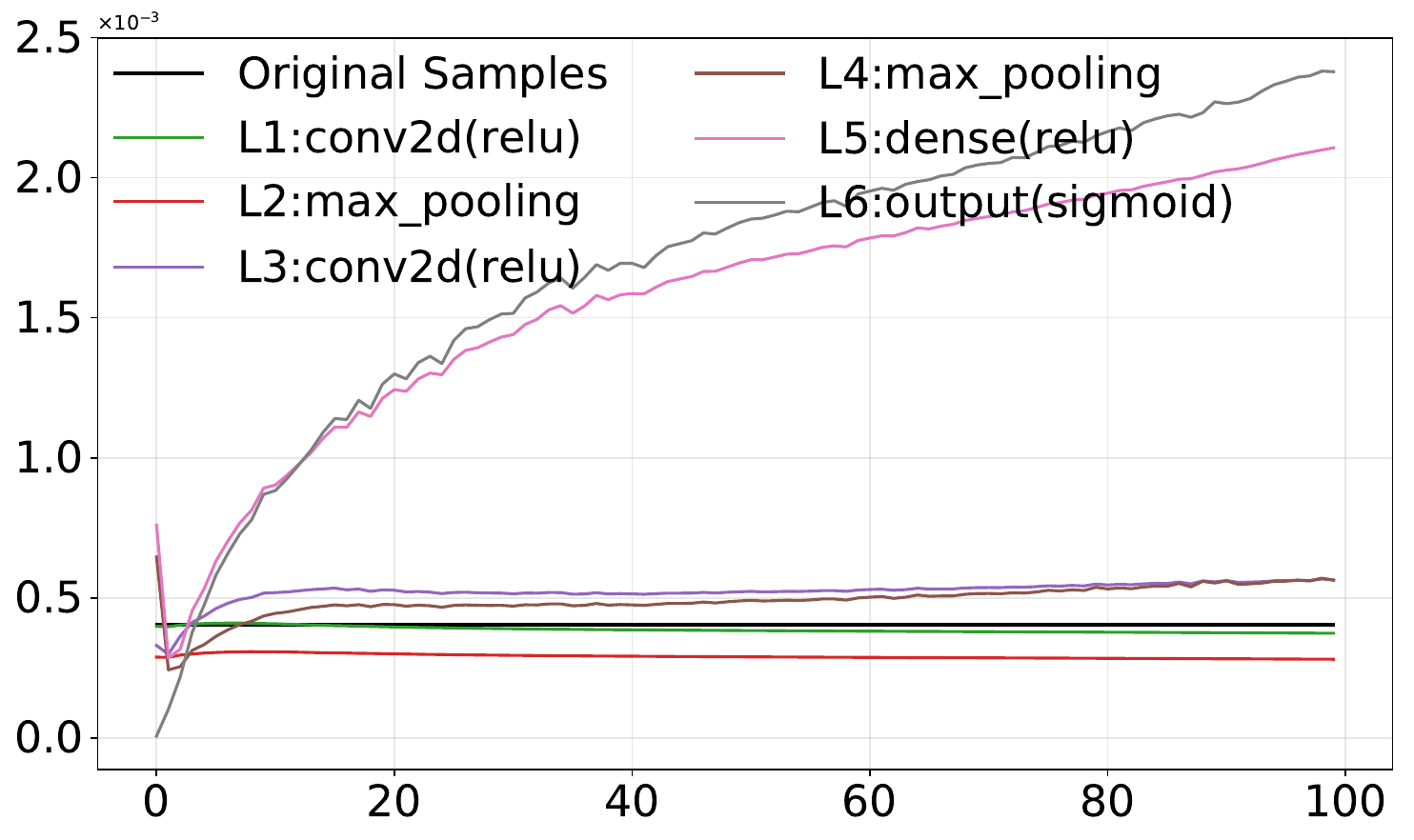}
}

\caption{Curves of MD-LSMs and $J_\omega$ for Hidden Layers of CNN}
\label{fig:CNN-LDA}
\end{figure}

\begin{figure}[htbp]
\centering	
\subfigure[$ {\rm LS}_0$]{
	\includegraphics[width=0.43\textwidth]{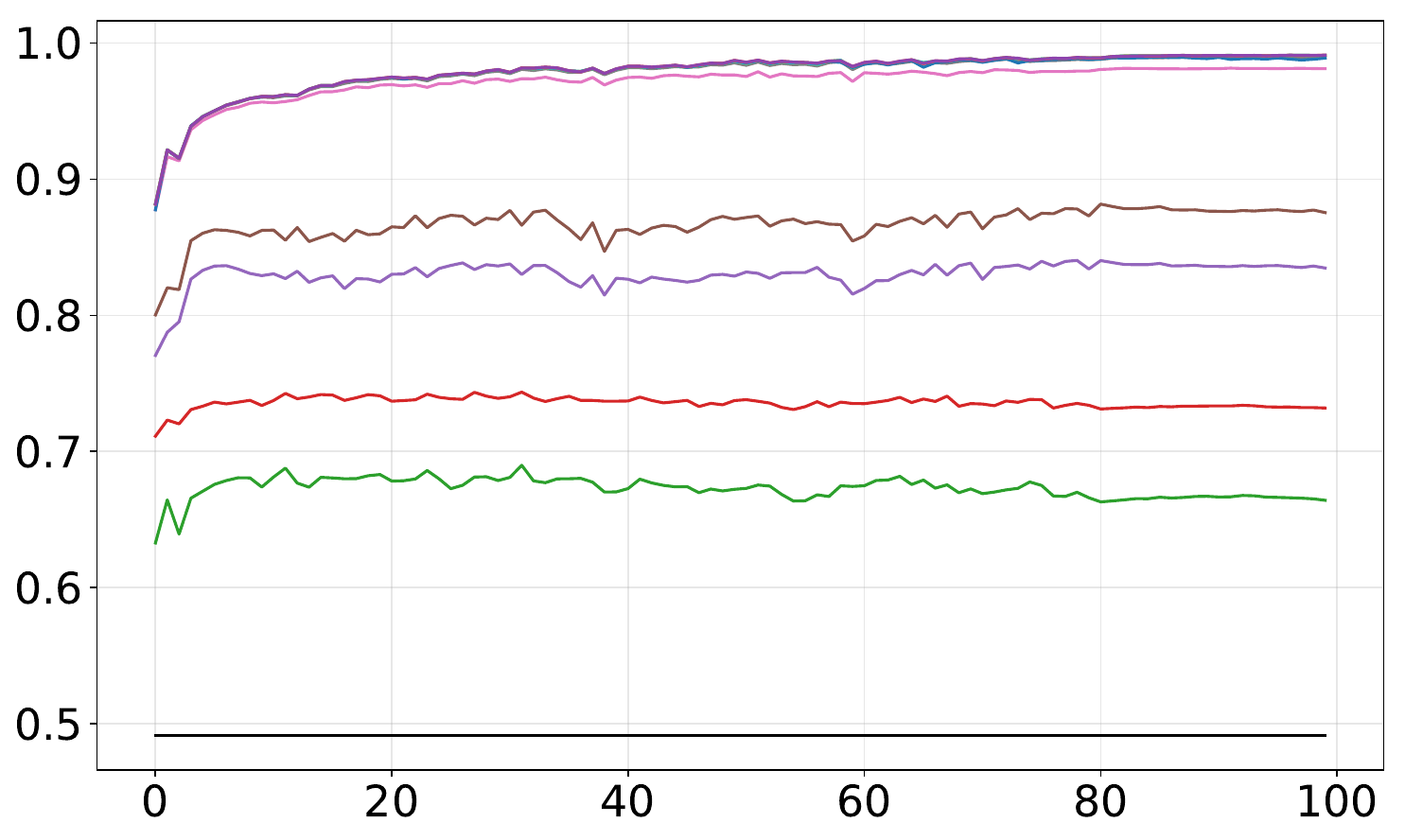}
}
\subfigure[$ {\rm LS}_1$]{
	\includegraphics[width=0.43\textwidth]{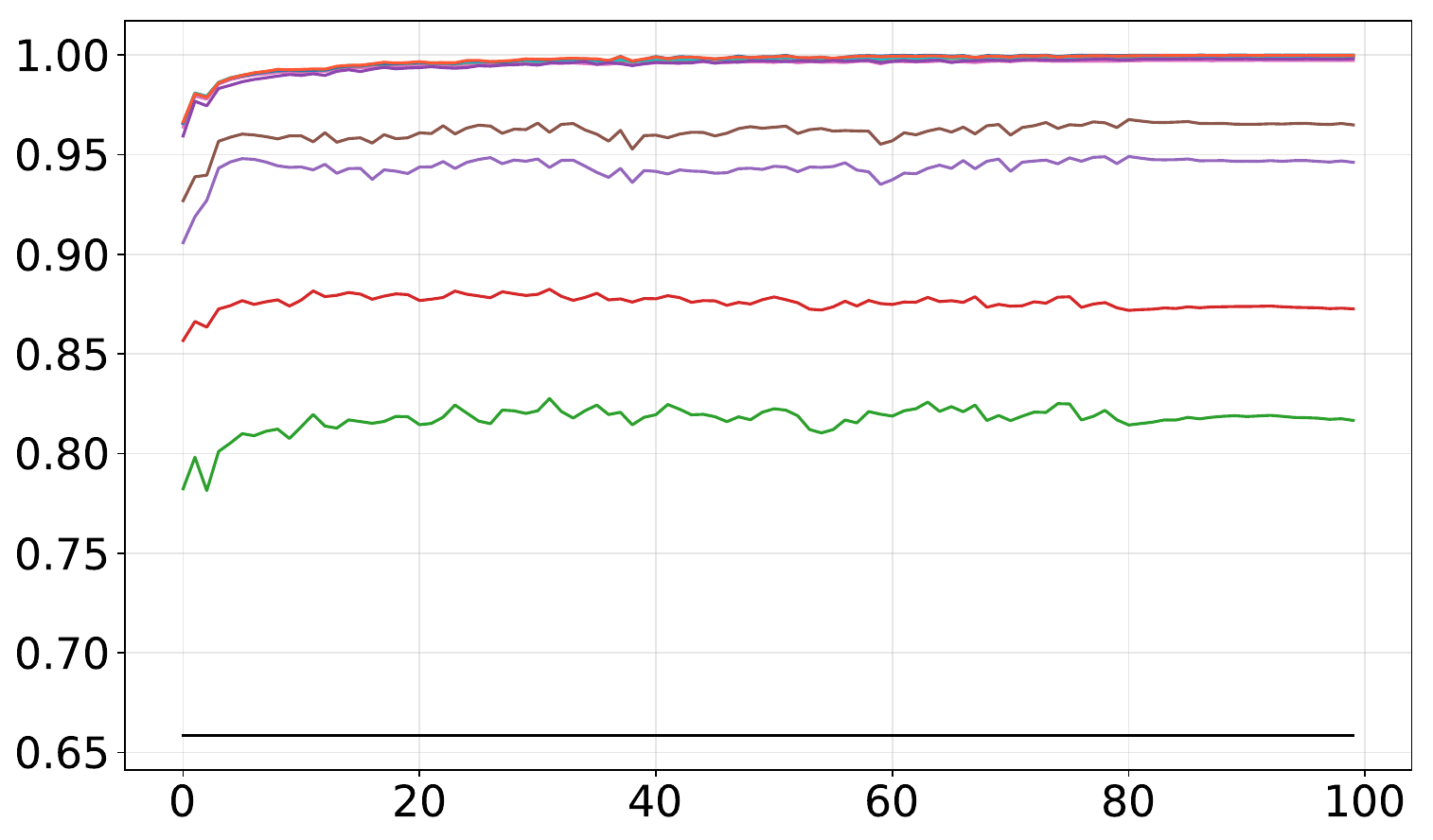}
}
\subfigure[$ {\rm LS}_2$]{
	\includegraphics[width=0.43\textwidth]{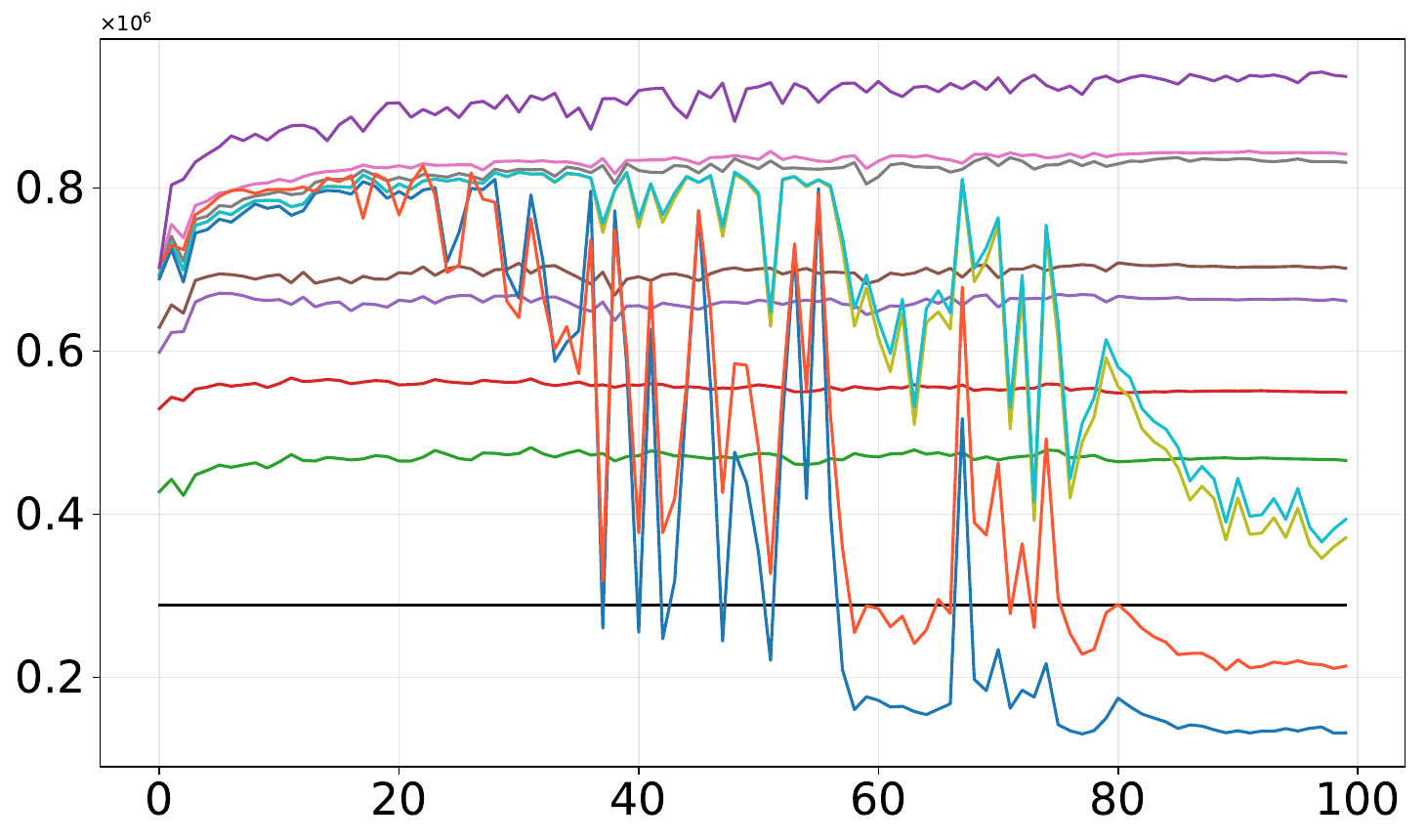}
}
\subfigure[$J_\omega$]{
	\includegraphics[width=0.43\textwidth]{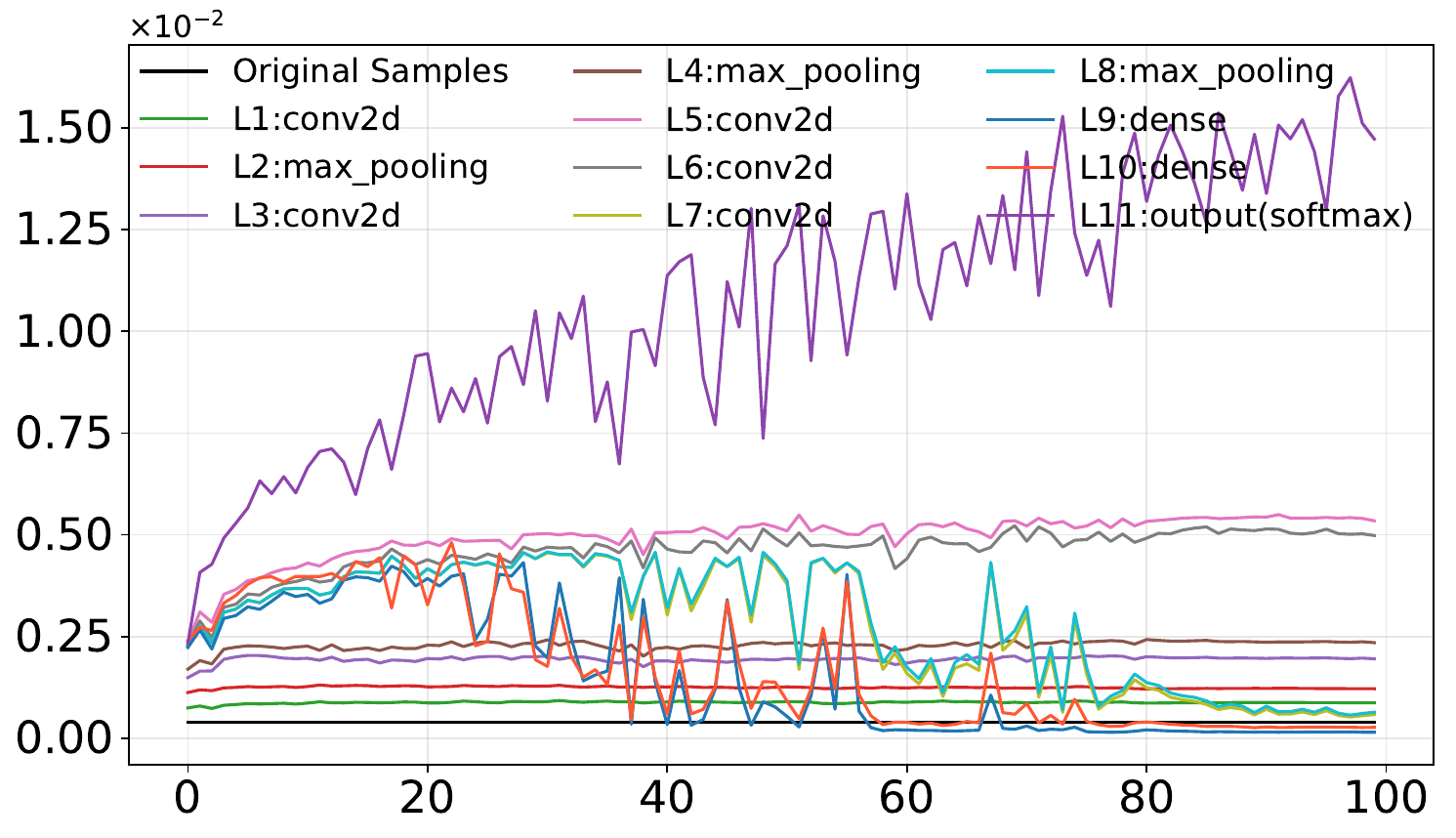}
}

\caption{Curves of MD-LSMs and $J_\omega$ for Hidden Layers of AlexNet (Softmax)}
\label{fig:AlexNet-LDA}	

\end{figure}

\begin{figure}[htbp]
\centering	

\subfigure[$ {\rm LS}_0$]{
	\includegraphics[width=0.43\textwidth]{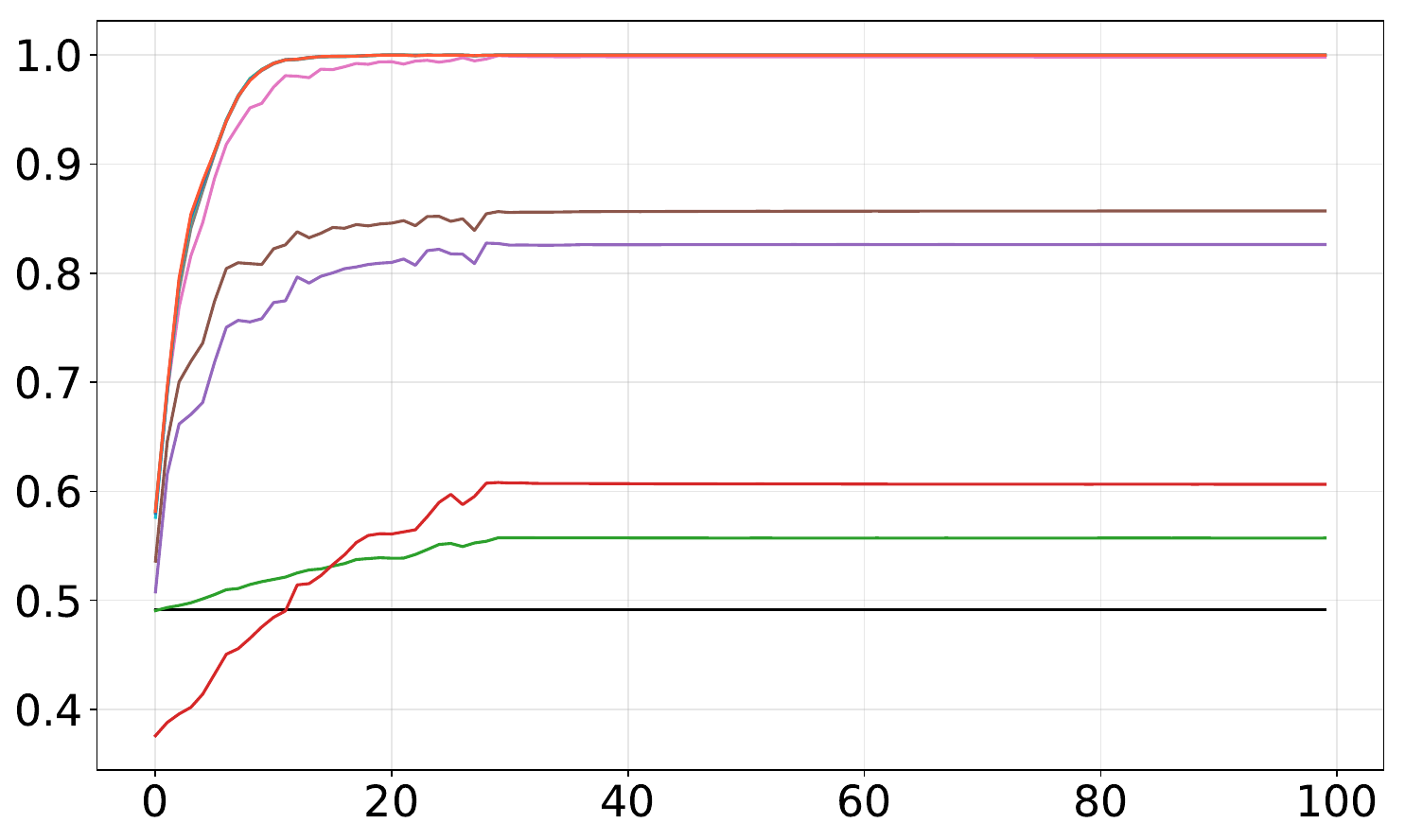}
}
\subfigure[$ {\rm LS}_1$]{
	\includegraphics[width=0.43\textwidth]{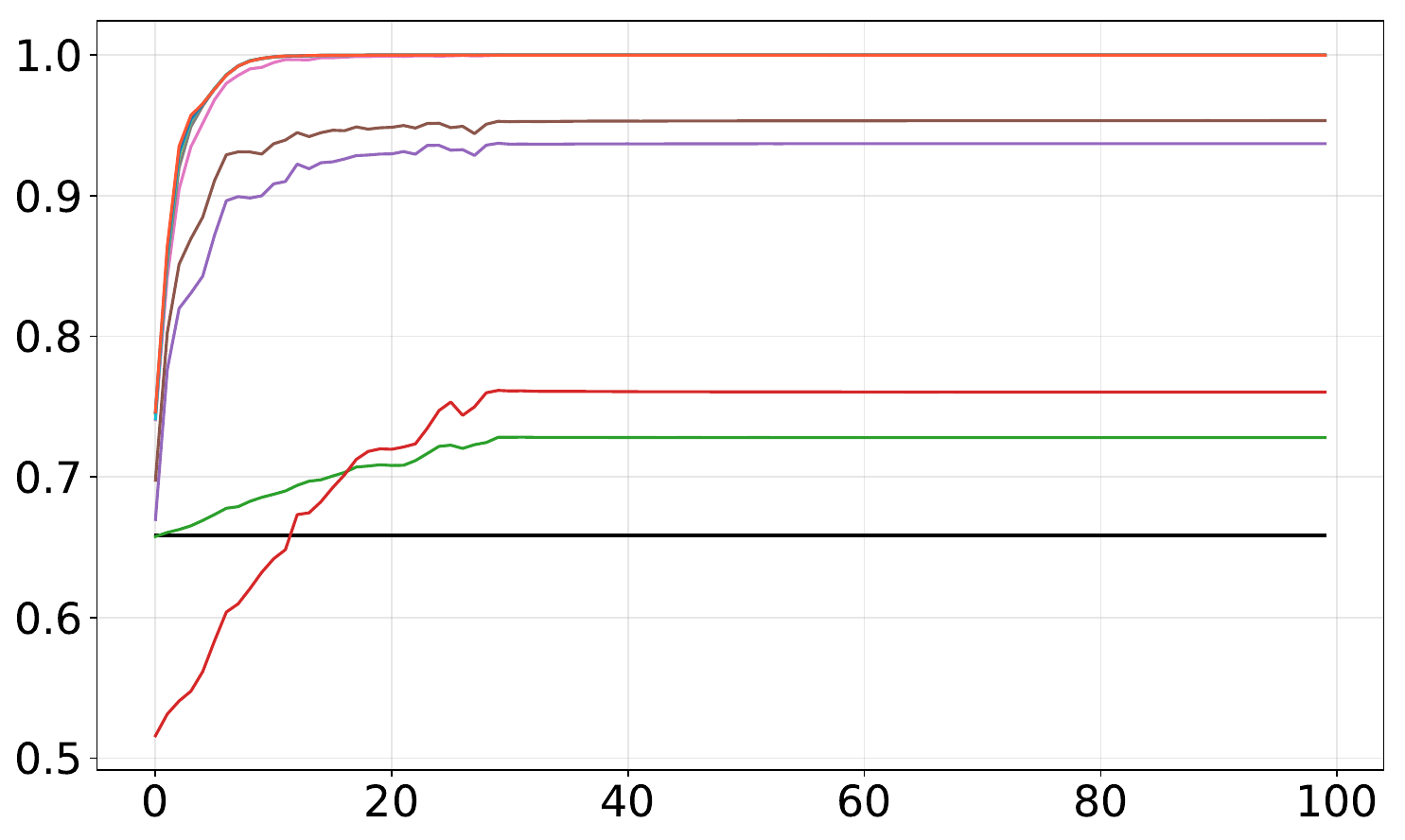}
}
\subfigure[$ {\rm LS}_2$]{
	\includegraphics[width=0.43\textwidth]{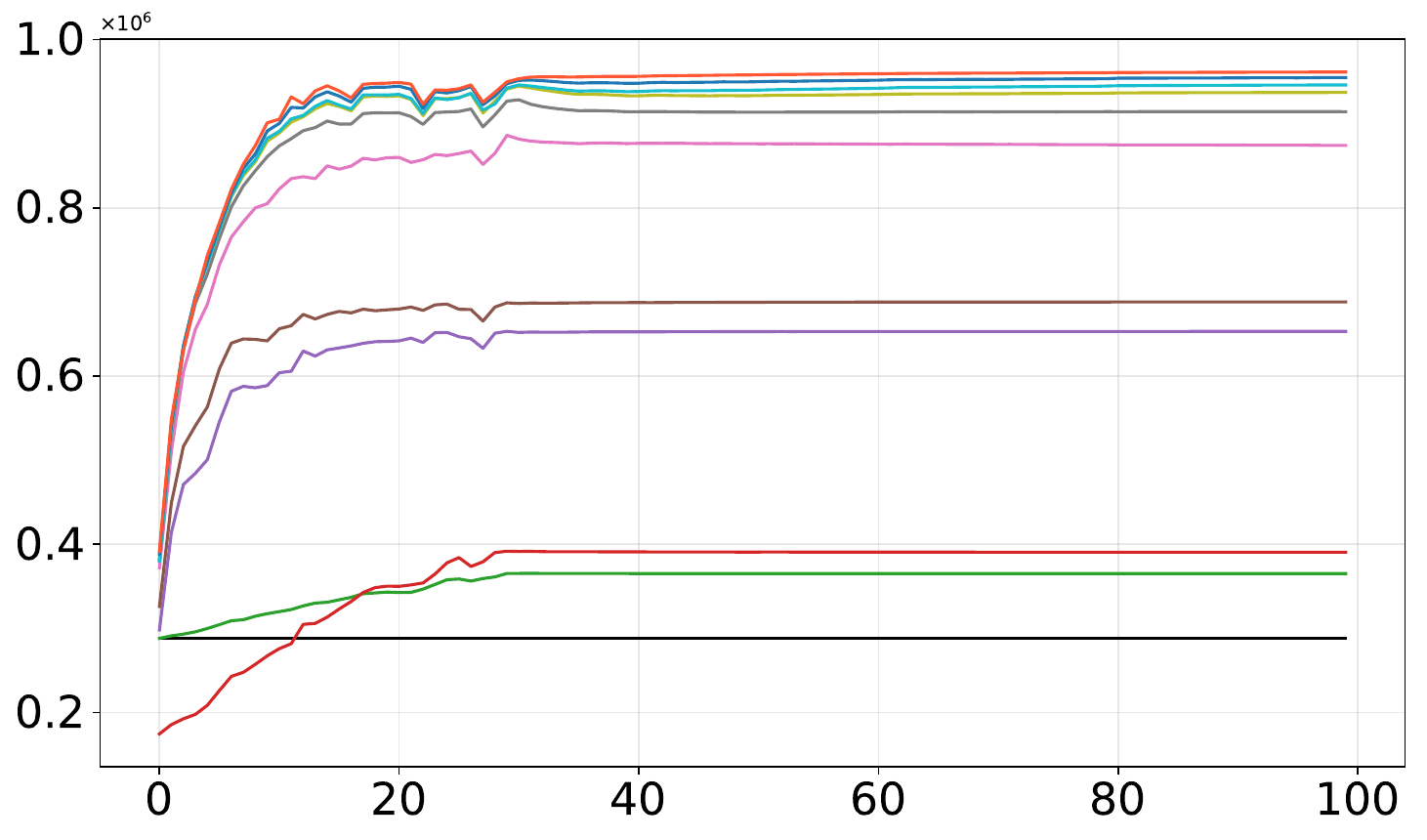}
}
\subfigure[$J_\omega$]{
	\includegraphics[width=0.43\textwidth]{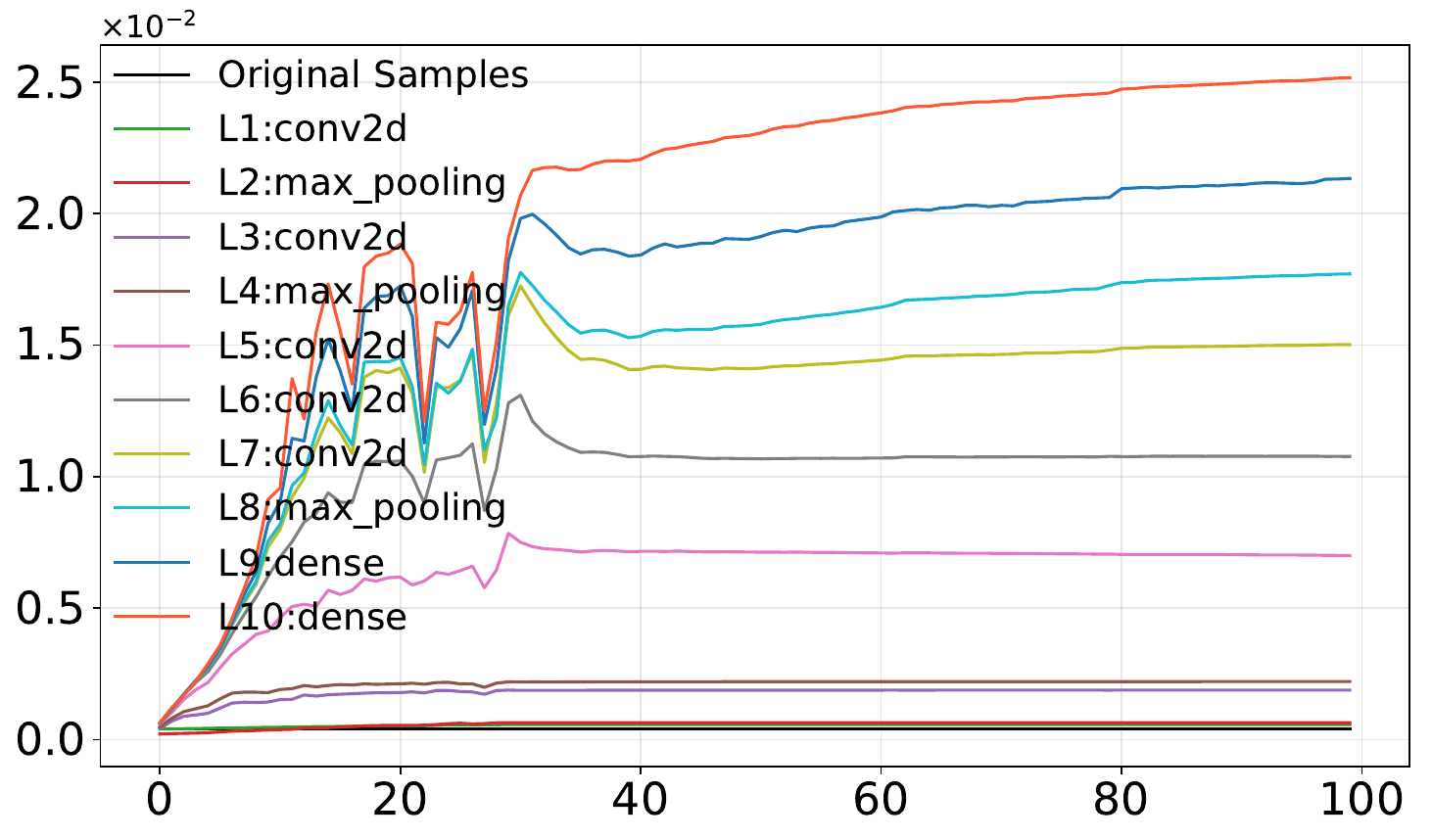}
}
\caption{Curves of MD-LSMs and $J_\omega$ for Hidden Layers of AlexNet (Sigmoid)}
\label{fig:AlexNet-LDA-no_output}	

\end{figure}

Moreover, the ranges of ${\rm LS}_0$ and ${\rm LS}_1$ are the interval $(0,1]$; and ${\rm LS}_0 ={\rm LS}_1 = 1$ holds if and only if the two sets are linearly separable ({\it cf.} Fig. \ref{fig:lda}). In contrast, the ranges of ${\rm LS}_2$ and $J_\omega$ are the interval $(0,+\infty)$, and they only provide the relative reference values for the linear separability. Thus, it is difficult to estimate the linear separability degree of two sets only based on the values of ${\rm LS}_2$ and $J_\omega$. Moreover, as shown in Figs. \ref{fig:MLP-LDA} -- \ref{fig:AlexNet-LDA-no_output}, there are fewer large fluctuations appearing in the curves of ${\rm LS}_0$ and ${\rm LS}_1$ than in the curves of ${\rm LS}_2$ and $J_\omega$. Interestingly, the curve shapes of ${\rm LS}_0$, ${\rm LS}_1$ and ${\rm LS}_2$ are the same, but they significantly differ from that of the curves of $J_\omega$. Therefore, we finally adopt the MD-LSMs as the measures of evaluating the linear separability degree of hidden layer outputs. 


%

\section{Proofs of Main Results}\label{supp:proof}

In this section, we give the proofs of Theorem \ref{thm:main}, Theorem \ref{thm:md}, Theorem \ref{thm:md-ls}, Theorem \ref{thm:linear.map}, Theorem \ref{thm:change}, Corollary \ref{cor:depth} and Proposition \ref{thm:LDA}, respectively.

\subsection{Proof of Theorem \ref{thm:main}}

{\bf Proof of Theorem \ref{thm:main}:} ``$\Longrightarrow$" If the classification accuracy increases, it means that more hidden-layer outputs can be correctly separated by using a hyperplane ${\bf w}^T {\bf x} + {\bf b} = 0$. Namely, the linear separability of hidden layer outputs increases. 

``$\Longleftarrow$" If the linear separability of the hidden layer outputs increases after updating the hidden-layer weights ${\bf V}' = {\bf V}+\triangle {\bf V}$, it means that there must exists a hyperplane $\bf{w} ^T {\bf s} + {\bf b} = 0$ such that more hidden-layer outputs can be correctly separated. Since the hyperplane $(\bf{w}')^T {\bf s} + {\bf b}' = 0$ can provide the highest training classification accuracy, the training performance of ${\rm net}'(\cdot)$ is better than that of ${\rm net}(\cdot)$. This completes the proof. \hfill$\blacksquare$

\subsection{Proof of Theorem \ref{thm:md}}

{\bf Proof of Theorem \ref{thm:md}:} ``$\Longrightarrow$": If ${\cal A}$ and ${\cal B}$ are linearly separable, there exists a vector $\bm{\omega}\in \mathbb{R}^N$ and a constant $c\in\mathbb{R}$ such that the relation $\mathbf{w}^T\mathbf{a} + c> \mathbf{w}^T\mathbf{b} + c$ holds for all $\mathbf{a}\in{\cal A}$ and $\mathbf{b}\in{\cal B}$. Then, we arrive at $\mathbf{w}^T(\mathbf{a}-\mathbf{b})> 0$ ($\forall \; \mathbf{a}\in{\cal A},\mathbf{b}\in{\cal B}$). Namely, all points of the Minkowski difference $\mathrm{MD}({\cal A},{\cal B})$ lie above the hyperplane $\mathbf{w}^T\mathbf{x}=0$.

``$\Longleftarrow$": Assume that all points of $\mathrm{MD}({\cal A},{\cal B})$ lie in one side of the hyperplane $\mathbf{w}^T\mathbf{x}=0$. Without loss of generality, we consider a vector $\mathbf{w}\in\mathbb{R}^N$ such that $\mathbf{w}^T(\mathbf{a} - \mathbf{b}) >0$ holds for any $\mathbf{a}\in\mathcal{A}$ and any $\mathbf{b}\in\mathcal{B}$. Define $\mathbf{a}^* := \arg\min _ {\mathbf{a}\in\mathcal{A}}\{ \mathbf{w}^T\mathbf{a} \} $ and $\mathbf{b}^\dag := \arg\max _ {\mathbf{b}\in\mathcal{B}}\{ \mathbf{w}^T\mathbf{b} \} $. Then, for all $\mathbf{a}\in\mathcal{A}$ and $\mathbf{b}\in\mathcal{B}$, it holds that
\begin{equation*}
	\mathbf{w}^T\mathbf{a} - \dfrac{\mathbf{w}^T\mathbf{a}^* + \mathbf{w}^T\mathbf{b}^\dag}{2}> 0 > \mathbf{w}^T\mathbf{b} - \dfrac{\mathbf{w}^T\mathbf{a}^* + \mathbf{w}^T\mathbf{b}^\dag}{2}.
\end{equation*}
Namely, the hyperplane $\mathbf{w}^T\mathbf{x} - \dfrac{\mathbf{w}^T\mathbf{a}^* + \mathbf{w}^T\mathbf{b}^\dag}{2} = 0$ ($x\in\mathbb{R}^N$) separates the set ${\cal A}$ from the set ${\cal B}$. This completes the proof. \hfill$\blacksquare$


\subsection{Proof of Theorem \ref{thm:md-ls}}

{\bf Proof of Theorem \ref{thm:md-ls}:} (1) The first equality is resulted from the definition of maximum linearly-separable subset. (2) It follows from ${\cal A}_\circ \subseteq {\cal A}$ and ${\cal B}_\circ \subseteq {\cal B}$ that 
\begin{eqnarray*}
	\frac{ 1}{ \frac{1}{|{\cal A}|} + \frac{1}{|{\cal B}|} } \geq  \frac{ 1}{ \frac{1}{|{\cal A}_\circ|} + \frac{1}{|{\cal B}_\circ|} } 
	\Longleftrightarrow 
	\frac{  |{\cal A}_\circ|+|{\cal B}_\circ| }{|{\cal A}|+|{\cal B}|}  \geq  \frac{|{\cal A}_\circ|\cdot|{\cal B}_\circ|}{|{\cal A}|\cdot|{\cal B}|} =  {\rm LS}_* ({\cal A},{\cal B}).
\end{eqnarray*}
The last equality holds because of the definition of ${\rm LS}_* ({\cal A},{\cal B})$. (3) Let $\bm{\omega}_*$ be the weight vector achieving the MD-LSM ${\rm LS}_* ({\cal A},{\cal B})$. Denote ${\cal A}_\circ^c = {\cal A}\setminus {\cal A}_\circ $ and ${\cal B}_\circ^c = {\cal B}\setminus {\cal B}_\circ $. Since ${\rm minor}_{\bm{\omega}_*}(\mathrm{MD}({\cal A},{\cal B})) \leq \min\big\{ |{\cal A}^c_\circ|\cdot |{\cal B} | ,  |{\cal B}^c_\circ|\cdot |{\cal A} | \big\}$, we arrive at
\begin{eqnarray*}
	& & 1 - \frac{{\rm minor}_{\bm{\omega}_*}(\mathrm{MD}({\cal A},{\cal B}))}{|{\cal A}|\cdot |{\cal B} |} \geq \max \left\{ 1- \frac{|{\cal A}^c_\circ|\cdot |{\cal B} |}{|{\cal A}|\cdot |{\cal B} |} , 1- \frac{|{\cal A}|\cdot |{\cal B}^c_\circ |}{|{\cal A}|\cdot |{\cal B} |} \right \} \\
	&\Longleftrightarrow  &  \frac{{\rm major}_{\bm{\omega}_*}(\mathrm{MD}({\cal A},{\cal B}))}{|{\cal A}|\cdot |{\cal B} |} \geq \max \left\{ \frac{|{\cal A}_\circ|\cdot |{\cal B} |}{|{\cal A}|\cdot |{\cal B} |} ,  \frac{|{\cal A}|\cdot |{\cal B}_\circ |}{|{\cal A}|\cdot |{\cal B} |} \right \} \\
	& \Longleftrightarrow  &  {\rm LS}_* ({\cal A},{\cal B}) \geq \max \left\{  \frac{|{\cal A}_\circ|}{|{\cal A}|} , \frac{|{\cal B}_\circ|}{|{\cal B}|}  \right \}.
\end{eqnarray*}
This completes the proof. \hfill$\blacksquare$


\subsection{Proof of Theorem \ref{thm:linear.map}}

{\bf Proof of Theorem \ref{thm:linear.map}:} Since ${\bf V}^T \mathbf{m}_{ij} = {\bf V}^T({\bf a}_i - {\bf b}_j)$, we have
\begin{align*}
	{\rm LS}_0({\bf V}({\cal A}),{\bf V}({\cal B})) =  & \max_{\bf{w}\in\mathbb{R}^H}\left\{\max \left\{  \frac{\sum\limits_{i\leq I,j\leq J} {\rm sgn}({\bf w}^T  {\bf V}^T \mathbf{m}_{ij})}{| \mathrm{MD}({\bf V}({\cal A}),{\bf V}({\cal B})) |} , \frac{\sum\limits_{i\leq I,j\leq J}{\rm sgn}(-{\bf w}^T  {\bf V}^T \mathbf{m}_{ij})}{| \mathrm{MD}({\bf V}({\cal A}),{\bf V}({\cal B})) |}    \right\} \right\}\nonumber\\
	=& \max_{\bm{\omega}\in\mathbb{R}^N}\left\{\max \left\{  \frac{\sum\limits_{i\leq I,j\leq J} {\rm sgn}(\bm{\omega}^T \mathbf{m}_{ij})}{| \mathrm{MD}({\cal A},{\cal B}) |} , \frac{\sum\limits_{i\leq I,j\leq J}{\rm sgn}(-\bm{\omega}^T \mathbf{m}_{ij})}{| \mathrm{MD}({\cal A},{\cal B}) |}    \right\}  \right\}\nonumber\\
	= & {\rm LS}_0({\cal A},{\cal B}).
\end{align*}
The rest can also be proven in the same way. This completes the proof. \hfill$\blacksquare$


\subsection{Proof of Theorem \ref{thm:change}}

{\bf Proof of Theorem \ref{thm:change}:}  Denote ${\bf m}_0 = {\bf a} - {\bf b}$ and ${\bf n}_0 = {\bf V}_\sigma({\bf a}) - {\bf V}_\sigma({\bf b})$. Without the loss of generality, we assume that the point ${\bf m}_0$ lies in the major side of the hyperplane $\bm{\omega}^T{\bf m} = 0$. Consider the following Taylor's expansion:
\begin{eqnarray*}
	{\bf V}_\sigma({\bf a})  &= &
	\left(
	\begin{array}{c}
		\sigma(\langle {\bf v}_1, {\bf a} \rangle)  \\
		\vdots \\
		\sigma(\langle {\bf v}_H, {\bf a} \rangle)
	\end{array}
	\right) =
	\left(
	\begin{array}{c}
		\sigma(\langle {\bf v}_1, {\bf b} \rangle) + \sigma'( \langle {\bf v}_1, {\bf b} \rangle) \cdot \langle {\bf v}_1, {\bf m}_0 \rangle  +  \frac{\sigma''(\langle \bm{\zeta}_1 , {\bf v}_1 \rangle)}{2} \cdot  {\bf v}_1^T {\bf m}_0{\bf m}_0^T {\bf v}_1\\
		\vdots \\
		\sigma(\langle {\bf v}_H, {\bf b} \rangle) + \sigma'( \langle {\bf v}_H, {\bf b} \rangle) \cdot \langle {\bf v}_H, {\bf m}_0  \rangle  +  \frac{\sigma''(\langle \bm{\zeta}_H , {\bf v}_H \rangle)}{2} \cdot  {\bf v}_H^T {\bf m}_0{\bf m}_0^T {\bf v}_H
	\end{array}
	\right)\nonumber\\
	&= & {\bf V}_\sigma({\bf b}) +
	\left(
	\begin{array}{c}
		\sigma'( \langle {\bf v}_1, {\bf b} \rangle) \cdot \langle {\bf v}_1, {\bf m}_0 \rangle  +  \frac{\sigma''(\langle \bm{\zeta}_1 , {\bf v}_1 \rangle)}{2} \cdot  {\bf v}_1^T {\bf m}_0{\bf m}_0^T {\bf v}_1\\
		\vdots \\
		\sigma'( \langle {\bf v}_H, {\bf b} \rangle) \cdot \langle {\bf v}_H, {\bf m}_0 \rangle  +  \frac{\sigma''(\langle \bm{\zeta}_H , {\bf v}_H \rangle)}{2} \cdot  {\bf v}_H^T {\bf m}_0{\bf m}_0^T {\bf v}_H
	\end{array}
	\right)\nonumber\\
	& \approx &  {\bf V}_\sigma({\bf b}) +
	\left(
	\begin{array}{c}
		\sigma'( \langle {\bf v}_1, {\bf b} \rangle) \cdot \langle {\bf v}_1, {\bf m}_0 \rangle  +  \frac{\sigma''(\langle  {\bf b} , {\bf v}_1 \rangle)}{2} \cdot  {\bf v}_1^T {\bf m}_0{\bf m}_0^T {\bf v}_1\\
		\vdots \\
		\sigma'( \langle {\bf v}_H, {\bf b} \rangle) \cdot \langle {\bf v}_H, {\bf m}_0 \rangle  +  \frac{\sigma''(\langle  {\bf b} , {\bf v}_H \rangle)}{2} \cdot  {\bf v}_H^T {\bf m}_0{\bf m}_0^T {\bf v}_H
	\end{array}
	\right),
\end{eqnarray*}
where $\bm{\zeta}_1,\cdots,\bm{\zeta}_H$ are $H$ points lying on the line segment between the two points ${\bf a}$ and ${\bf b}$. Alternatively, we also have
\begin{eqnarray*}
	{\bf V}_\sigma({\bf b})  & \approx &  {\bf V}_\sigma({\bf a}) +
	\left(
	\begin{array}{c}
		- \sigma'( \langle {\bf v}_1, {\bf a} \rangle) \cdot \langle {\bf v}_1, {\bf m} _0\rangle  +  \frac{\sigma''(\langle  {\bf a}, {\bf v}_1 \rangle)}{2} \cdot  {\bf v}_1^T {\bf m}_0{\bf m}_0^T {\bf v}_1\\
		\vdots \\
		- \sigma'( \langle {\bf v}_H, {\bf a} \rangle) \cdot \langle {\bf v}_H, {\bf m}_0 \rangle  +  \frac{\sigma''(\langle {\bf a} , {\bf v}_H \rangle)}{2} \cdot  {\bf v}_H^T {\bf m}_0{\bf m}_0^T {\bf v}_H
	\end{array}
	\right).
\end{eqnarray*}
Then, we arrive at
\begin{eqnarray*}
	{\bf n} _0& = & {\bf V}_\sigma({\bf a})  - {\bf V}_\sigma({\bf b}) \\
	& \approx &  \left(
	\begin{array}{c}
		\frac{  \sigma'( \langle {\bf v}_1, {\bf b} \rangle) + \sigma'( \langle {\bf v}_1, {\bf a} \rangle)}{2}   \cdot \langle {\bf v}_1, {\bf m}_0 \rangle  +  \frac{\sigma''(\langle  {\bf b}, {\bf v}_1 \rangle) - \sigma''(\langle  {\bf a}, {\bf v}_1 \rangle)}{4} \cdot  {\bf v}_1^T {\bf m}_0{\bf m}_0^T {\bf v}_1\\
		\vdots \\
		\frac{  \sigma'( \langle {\bf v}_H, {\bf b} \rangle) + \sigma'( \langle {\bf v}_H, {\bf a} \rangle)}{2} \cdot \langle {\bf v}_H, {\bf m}_0 \rangle  +  \frac{\sigma''(\langle  {\bf b}, {\bf v}_H \rangle) - \sigma''(\langle  {\bf a}, {\bf v}_H \rangle)}{4} \cdot  {\bf v}_H^T {\bf m}_0{\bf m}_0^T {\bf v}_H
	\end{array}
	\right)\\
	& =&  \left(
	\begin{array}{c}
		\frac{  \sigma'( \langle {\bf v}_1, {\bf b} \rangle) + \sigma'( \langle {\bf v}_1, {\bf a} \rangle)}{2}   \cdot \langle {\bf v}_1, {\bf m}_0 \rangle  \\
		\vdots \\
		\frac{  \sigma'( \langle {\bf v}_H, {\bf b} \rangle) + \sigma'( \langle {\bf v}_H, {\bf a} \rangle)}{2} \cdot \langle {\bf v}_H, {\bf m}_0 \rangle
	\end{array}
	\right) +  \left(
	\begin{array}{c}
		\frac{\sigma''(\langle  {\bf b}, {\bf v}_1 \rangle) - \sigma''(\langle  {\bf a}, {\bf v}_1 \rangle)}{4} \cdot  {\bf v}_1^T {\bf m}_0{\bf m}_0^T {\bf v}_1\\
		\vdots \\
		\frac{\sigma''(\langle  {\bf b}, {\bf v}_H \rangle) - \sigma''(\langle  {\bf a}, {\bf v}_H \rangle)}{4} \cdot  {\bf v}_H^T {\bf m}_0{\bf m}_0^T {\bf v}_H
	\end{array}
	\right)\\
	& = & \widetilde{{\bf V}}^T {\bf m}_0 + {\bf q}({\bf V},{\bf m}_0) \\
	&= &{\bf p}({\bf V},{\bf m}_0) + {\bf q}({\bf V},{\bf m}_0).
\end{eqnarray*}
with
\begin{equation*}
	\widetilde{{\bf V}} = \left[   \frac{ \big[ \sigma'( \langle {\bf v}_1, {\bf b} \rangle) + \sigma'( \langle {\bf v}_1, {\bf a} \rangle) \big] {\bf v}_1 }{2}    , \cdots,    \frac{ \big[ \sigma'( \langle {\bf v}_H, {\bf b} \rangle) + \sigma'( \langle {\bf v}_H, {\bf a} \rangle)\big]   {\bf v}_H }{2}  \right].
\end{equation*}

Since the first derivate $\sigma'(\cdot)$ is non-negative, the relative position of the point $\widetilde{{\bf V}}^T {\bf m}_0 $ will not be changed, {\it i.e.,} the point ${\bf p}({\bf V},{\bf m}_0)$ still lies in the major side of the hyperplane $(\bm{\omega}^T{\bf V}) \cdot {\bf n}_0 =0$. To change the relative position of ${\bf p}({\bf V},{\bf m}_0)$ w.r.t. the hyperplane $(\bm{\omega}^T{\bf V}) {\bf n}_0 =0$, the following condition should be satisfied: denoting ${\bf w}^T = \bm{\omega}^T{\bf V} $,
\begin{align*}
	&\left\{
	\begin{array}{r}
		|{\bf w}^T {\bf p}({\bf V},{\bf m}_0)| < |{\bf w}^T {\bf q}({\bf V},{\bf m}_0)|;\\
		{\bf p}^T({\bf V},{\bf m}_0) \cdot {\bf q}({\bf V},{\bf m}_0) <0.
	\end{array}
	\right.
\end{align*}
The second formula is equivalent to
\begin{align*}
	& {\bf p}^T({\bf V},{\bf m}_0) \cdot {\bf q}({\bf V},{\bf m}_0)  <  0 \nonumber\\
	\Longleftrightarrow  & \sum_{h=1}^H  \Big[  \sigma'( \langle {\bf v}_h, {\bf b} \rangle) + \sigma'( \langle {\bf v}_h, {\bf a} \rangle) \Big] \cdot \Big[\sigma''(\langle  {\bf b}, {\bf v}_h \rangle) - \sigma''(\langle  {\bf a}, {\bf v}_h \rangle) \Big] \cdot \frac{\langle {\bf v}_h, {\bf m}_0 \rangle^3}{8}      <  0 \nonumber\\
	%
	%
	\Longleftarrow  &    \Big[  \sigma'( \langle {\bf v}_h, {\bf b} \rangle) + \sigma'( \langle {\bf v}_h, {\bf a} \rangle) \Big] \cdot \Big[\sigma''(\langle  {\bf b}, {\bf v}_h \rangle) - \sigma''(\langle  {\bf a}, {\bf v}_h \rangle) \Big] \cdot \langle {\bf v}_h, {\bf m}_0 \rangle      <  0.
\end{align*}
Namely, the sufficient condition for the second relation is that the points ${\bf a},{\bf b}$ and the weights ${\bf v}_1,\cdots,{\bf v}_H$ satisfy that $F_1(  \langle  {\bf a}, {\bf v}_h \rangle, \langle  {\bf b}, {\bf v}_h \rangle ) > 0$ holds for any $1\leq h\leq H$, where
\begin{equation*}
	F_1(x,y) = \Big[\sigma'(x ) + \sigma'(y) \Big] \cdot \Big[\sigma''(x ) - \sigma''(y) \Big] \cdot (x-y), \;\; (x,y)\in\mathbb{R}\times \mathbb{R}.
\end{equation*}

%

Next, we also need to consider the first relation
$$|{\bf w}^T {\bf p}({\bf V},{\bf m}_0)| < |{\bf w}^T {\bf q}({\bf V},{\bf m}_0)|.$$
Since the sign of ${\bf m}_0^T {\bf v}_h$ is always opposite to that of $\sigma''(\langle  {\bf b}, {\bf v}_h \rangle) - \sigma''(\langle  {\bf a}, {\bf v}_h \rangle)$, we only need to consider the magnitudes of ${\bf w}^T {\bf p}({\bf V},{\bf m}_0)$ and $-{\bf w}^T {\bf q}({\bf V},{\bf m}_0)$. 
\begin{enumerate}[(i)]
	\item
	If the sign of $\langle {\bf v}_h, {\bf w} \rangle$ is same as that of $\langle {\bf v}_h, {\bf m}_0 \rangle$ for any $h\in\{1,\cdots,H\}$, then we have
	\begin{align*}
		& |{\bf w}^T {\bf p}({\bf V},{\bf m}_0)| < |{\bf w}^T {\bf q}({\bf V},{\bf m}_0)|\\
		\Longleftrightarrow& {\bf w}^T {\bf p}({\bf V},{\bf m}_0) <  - {\bf w}^T {\bf q}({\bf V},{\bf m}_0) \\
		\Longleftrightarrow & \sum_{h=1}^H \frac{  ({\bf v}_h^T {\bf w}) \big[\sigma'( \langle {\bf v}_h, {\bf b} \rangle) + \sigma'( \langle {\bf v}_h, {\bf a} \rangle)\big] ({\bf v}_h^T {\bf m}_0)}{2} \\
		& \qquad\qquad< - \sum_{h=1}^H \frac{ ({\bf v}_h^T {\bf w}) \big[\sigma''(\langle  {\bf b}, {\bf v}_h \rangle) - \sigma''(\langle  {\bf a}, {\bf v}_h \rangle)\big] ({\bf v}_h^T {\bf m}_0)^2 }{4},\quad \forall h\in\{1,2,\cdots,H\}\\
		\Longleftarrow   &\frac{  ({\bf v}_h^T {\bf w}) \big[\sigma'( \langle {\bf v}_h, {\bf b} \rangle) + \sigma'( \langle {\bf v}_h, {\bf a} \rangle)\big] ({\bf v}_h^T {\bf m}_0)}{2} \\
		& \qquad\qquad < \frac{ ({\bf v}_h^T {\bf w}) \big[\sigma''(\langle  {\bf a}, {\bf v}_h \rangle) - \sigma''(\langle  {\bf b}, {\bf v}_h \rangle)\big] ({\bf v}_h^T {\bf m}_0)^2 }{4},\quad \forall h\in\{1,2,\cdots,H\}\\
		\Longleftrightarrow   &2  \big[\sigma'( \langle {\bf v}_h, {\bf b} \rangle) + \sigma'( \langle {\bf v}_h, {\bf a} \rangle)\big]  <  \big[\sigma''(\langle  {\bf a}, {\bf v}_1 \rangle) - \sigma''(\langle  {\bf b}, {\bf v}_h \rangle)\big] ({\bf v}_h^T {\bf m}_0) ,\quad \forall h\in\{1,2,\cdots,H\}\\
		\Longleftrightarrow   &2   <  \frac{\big[\sigma''(\langle  {\bf a}, {\bf v}_h \rangle) - \sigma''(\langle  {\bf b}, {\bf v}_h \rangle)\big] ({\bf v}_h^T {\bf m}_0)}{\big[\sigma'( \langle {\bf v}_h, {\bf b} \rangle) + \sigma'( \langle {\bf v}_h, {\bf a} \rangle)\big] },\quad \forall h\in\{1,2,\cdots,H\}.
	\end{align*}

	\item If the sign of $\langle {\bf v}_h, {\bf w} \rangle$ is the same as that of $-\langle {\bf v}_h, {\bf m}_0 \rangle$ for any $h\in\{1,\cdots,H\}$, then we have
	\begin{align*}
		& |{\bf w}^T {\bf p}({\bf V},{\bf m}_0)| < |{\bf w}^T {\bf q}({\bf V},{\bf m}_0)|\\
		\Longleftrightarrow& -{\bf w}^T {\bf p}({\bf V},{\bf m}_0) < {\bf w}^T {\bf q}({\bf V},{\bf m}_0) \\
		\Longleftrightarrow &\sum_{h=1}^H \frac{  ({\bf v}_h^T {\bf w}) \big[\sigma'( \langle {\bf v}_h, {\bf b} \rangle) + \sigma'( \langle {\bf v}_h, {\bf a} \rangle)\big] ({\bf v}_h^T {\bf m}_0)}{2} \\
		& \qquad\qquad >- \sum_{h=1}^H \frac{ ({\bf v}_h^T {\bf w}) \big[\sigma''(\langle  {\bf b}, {\bf v}_h \rangle) - \sigma''(\langle  {\bf a}, {\bf v}_h \rangle)\big] ({\bf v}_h^T {\bf m}_0)^2 }{4}\\
		\Longleftarrow   &\frac{  ({\bf v}_h^T {\bf w}) \big[\sigma'( \langle {\bf v}_h, {\bf b} \rangle) + \sigma'( \langle {\bf v}_h, {\bf a} \rangle)\big] ({\bf v}_h^T {\bf m}_0)}{2}  \\
		& \qquad\qquad> \frac{ ({\bf v}_h^T {\bf w}) \big[\sigma''(\langle  {\bf a}, {\bf v}_h \rangle) - \sigma''(\langle  {\bf b}, {\bf v}_h \rangle)\big] ({\bf v}_h^T {\bf m}_0)^2 }{4},\quad \forall h\in\{1,2,\cdots,H\}\\
		\Longleftrightarrow   &2  \big[\sigma'( \langle {\bf v}_h, {\bf b} \rangle) + \sigma'( \langle {\bf v}_h, {\bf a} \rangle)\big]  <  \big[\sigma''(\langle  {\bf a}, {\bf v}_1 \rangle) - \sigma''(\langle  {\bf b}, {\bf v}_h \rangle)\big] ({\bf v}_h^T {\bf m}_0) ,\quad \forall h\in\{1,2,\cdots,H\}\\
		\Longleftrightarrow   &2   <  \frac{\big[\sigma''(\langle  {\bf a}, {\bf v}_h \rangle) - \sigma''(\langle  {\bf b}, {\bf v}_h \rangle)\big] ({\bf v}_h^T {\bf m}_0)}{\big[\sigma'( \langle {\bf v}_h, {\bf b} \rangle) + \sigma'( \langle {\bf v}_h, {\bf a} \rangle)\big] },\quad \forall h\in\{1,2,\cdots,H\}.
	\end{align*}
\end{enumerate}
To sum up, the sufficient condition for $ |{\bf w}^T {\bf p}({\bf V},{\bf m}_0)| < |{\bf w}^T {\bf q}({\bf V},{\bf m}_0)|$ is 
\begin{equation*}
	2   <  \frac{\big[\sigma''(\langle  {\bf a}, {\bf v}_h \rangle) - \sigma''(\langle  {\bf b}, {\bf v}_h \rangle)\big] ({\bf v}_h^T {\bf m}_0)}{\big[\sigma'( \langle {\bf v}_h, {\bf b} \rangle) + \sigma'( \langle {\bf v}_h, {\bf a} \rangle)\big] },\qquad \forall h \in \{1,2,\cdots,H\}.
\end{equation*}
Denote the function
\begin{equation*}
	F_\sigma(x,y) = \frac{\big[\sigma''(x) - \sigma''(y)\big] (x-y)}{\big[\sigma'(x) + \sigma'(y)\big] },\;\; (x,y)\in\mathbb{R}\times \mathbb{R}.
\end{equation*}
It is clear that the right hand side of the above formula is $F_\sigma(\langle  {\bf a}, {\bf v}_h \rangle,\langle  {\bf b}, {\bf v}_h \rangle)$. Then, the above issue is equivalent to finding the range such that $F_\sigma(x,y)>2$. Namely, as long as the points ${\bf a},{\bf b}$ and the weight vectors ${\bf v}_1,\cdots,{\bf v}_H$ satisfy that $F_\sigma(\langle  {\bf a}, {\bf v}_h \rangle,\langle  {\bf b}, {\bf v}_h \rangle)>2$ ($\forall h\in\{1,2,\cdots,H\}$), the point ${\bf n}_0 ={\bf V}_\sigma({\bf a}) -{\bf V}_\sigma({\bf b})$ lies in the minor side of the hyperplane ${\bf w}^T {\bf n} = (\bm{\omega}^T {\bf V}) {\bf n} = 0 $. Since $F_\sigma(\langle  {\bf a}, {\bf v}_h \rangle,\langle  {\bf b}, {\bf v}_h \rangle)>2$ implies that $F_1(  \langle  {\bf a}, {\bf v}_h \rangle, \langle  {\bf b}, {\bf v}_h \rangle ) > 0$, this completes the proof. \hfill $\blacksquare$



\subsection{Proof of Corollary \ref{cor:depth}}

{\bf Proof of Corollary \ref{cor:depth}:} For convenience, we briefly denote $a^L : = {\rm LS}_2({\bf V}^{L}_\sigma({\cal A}),{\bf V}^{L}_\sigma({\cal B}))$, $a^0 = {\rm LS}_2({\cal A},{\cal B}) $ and ${\bf V}^{(0)} = {\bf I}$. According to the law of total probability, we have 
\begin{eqnarray}\label{eq:layer.pr1}
	&&\mathbb{P} \{  a^{L} > a^0    \} \nonumber \\
	&= &
	\mathbb{P} \{  a^{L} > a^0 | a^{L-1} > a^0   \}  \cdot \mathbb{P} \{  a^{L-1} > a^0  \}  + \mathbb{P} \{  a^{L} > a^0 | a^{L-1}  \leq a^0   \}  \cdot \mathbb{P} \{  a^{L-1} \leq a^0  \} \nonumber\\
	&\leq & \mathbb{P} \{  a^{L-1} > a^0  \}  + \mathbb{P} \{  a^{L} > a^0 | a^{L-1}  \leq a^0   \}  \cdot \mathbb{P} \{  a^{L-1} \leq a^0  \} \nonumber\\
	&= & \mathbb{P} \{  a^{L-1} > a^0  \}  + \mathbb{P} \{  a^{L} > a^0 , a^{L-1}  \leq a^0   \}   \nonumber\\
	&\leq & \mathbb{P} \{  a^{L-1} > a^0  \}  + \mathbb{P} \{  a^{L} >  a^{L-1}    \}   \nonumber\\
	&= & \mathbb{P} \{  a^{L-1} > a^0  \}  +  \sum_{{\bf V}^{(1)}  ,\cdots, {\bf V}^{(L-1)}} \mathbb{P}({\bf V}^{(1)}  ,\cdots, {\bf V}^{(L-1)}) \cdot \mathbb{P} \{  a^{L} >  a^{L-1} | {\bf V}^{(1)}  ,\cdots, {\bf V}^{(L-1)}   \}   \nonumber\\
	& \leq & \mathbb{P} \{  a^{L-1} > a^0  \}  + \max_{{\bf V}^{(1)}  ,\cdots, {\bf V}^{(L-1)}} \mathbb{P} \{  a^{L} >  a^{L-1}    | {\bf V}^{(1)}  ,\cdots, {\bf V}^{(L-1)} \}\nonumber\\
	& & \vdots \nonumber\\
	& \leq & \sum_{l=1}^L   \max_{{\bf V}^{(1)}  ,\cdots, {\bf V}^{(l-1)}} \mathbb{P} \{  a^{l} >  a^{l-1}    | {\bf V}^{(1)}  ,\cdots, {\bf V}^{(l-1)} \}.
\end{eqnarray}
The combination of Eq. \eqref{eq:node} and Eq. \eqref{eq:layer.pr1} leads to the result \eqref{eq:layer}. This completes the proof. \hfill $\blacksquare$





\subsection{Proof of Proposition \ref{thm:LDA}}

{\bf Proof of Proposition \ref{thm:LDA}:} Denote ${\bf A} = [{\bf a}_1,\cdots,{\bf a}_I]$ and ${\bf B} = [{\bf b}_1,\cdots,{\bf b}_J]$. Let ${\bf 1} = (1,\cdots,1)^T$ be the vector whose components are all ones. Since $\bm{\mu}_a = \frac{1}{I}{\bf A} {\bf 1}$ and $[\bm{\mu}_a ,\cdots, \bm{\mu}_a]  = \frac{1}{I} {\bf A} {\bf 1}{\bf 1}^T$, we have
\begin{eqnarray*}
{\bf A}_c{\bf A}_c^T  & = & \left( {\bf A}  -  \frac{1}{I} {\bf A} {\bf 1}{\bf 1}^T \right) \left( {\bf A}  -  \frac{1}{I} {\bf A} {\bf 1}{\bf 1}^T \right)^T\\
& = &  {\bf A}{\bf A}^T  -  \frac{1}{I} {\bf A} {\bf 1}{\bf 1}^T{\bf A}^T - \frac{1}{I} {\bf A} {\bf 1}{\bf 1}^T{\bf A}^T + \frac{1}{I^2} {\bf A} {\bf 1}{\bf 1}^T {\bf 1}{\bf 1}^T{\bf A}^T\\
& = &  {\bf A}{\bf A}^T  -  \frac{1}{I} {\bf A} {\bf 1}{\bf 1}^T{\bf A}^T - \frac{1}{I} {\bf A} {\bf 1}{\bf 1}^T{\bf A}^T + \frac{1}{I} {\bf A} {\bf 1}{\bf 1}^T{\bf A}^T\\
& = &  {\bf A}{\bf A}^T  -  \frac{1}{I} {\bf A} {\bf 1}{\bf 1}^T{\bf A}^T .
\end{eqnarray*}
In the similar way, we also have ${\bf B}_c{\bf B}_c^T = {\bf B}{\bf B}^T  -  \frac{1}{J} {\bf B} {\bf 1}{\bf 1}^T{\bf B}^T$. Thus, the matrices ${\bf S}_w$ and ${\bf S}_b$ can be rewritten as
\begin{eqnarray*}
{\bf S}_w & = &  {\bf A}_c{\bf A}_c^T  + {\bf B}_c{\bf B}_c^T \\
& = &  {\bf A}  {\bf A} ^T  + {\bf B}  {\bf B} ^T  - \frac{1}{I}   {\bf A} {\bf 1} {\bf 1}^T    {\bf A} ^T    - \frac{1}{J}   {\bf B} {\bf 1} {\bf 1}^T    {\bf B} ^T;\\
{\bf S}_b & = &  (\bm{\mu}_a-\bm{\mu}_b) (\bm{\mu}_a-\bm{\mu}_b)^T\\
& = &  \frac{1}{I^2} {\bf A} {\bf 1} {\bf 1}^T {\bf A}^T  +  \frac{1}{J^2} {\bf B} {\bf 1}{\bf 1}^T{\bf B}^T - \frac{1}{IJ}   {\bf A} {\bf 1} {\bf 1}^T    {\bf B} ^T  - \frac{1}{IJ}   {\bf B} {\bf 1} {\bf 1}^T    {\bf A} ^T.
\end{eqnarray*}


Denote
\begin{eqnarray*}
{\rm D}({\bf A};J) &:= &\underbrace{[{\bf a}_1,\cdots, {\bf a}_I,\cdots, {\bf a}_1,\cdots, {\bf a}_I   ]}_{\mbox{$J$ groups of $\{ {\bf a}_1,\cdots, {\bf a}_I \}$  }} \in \mathbb{R}^{N\times IJ}; \\
{\rm D}({\bf B};I) & :=  &\underbrace{[{\bf b}_1,\cdots, {\bf b}_J,\cdots, {\bf b}_1,\cdots, {\bf b}_J   ]}_{\mbox{$I$ groups of $ \{{\bf b}_1,\cdots, {\bf b}_J \}$  }} \in \mathbb{R}^{N\times IJ}.
\end{eqnarray*}
Since ${\bf m}_{ij}={\bf a}_i - {\bf b}_j$, ${\bf M}$ can be rewritten as
\begin{eqnarray*}
{\bf M}  & = & [\mathbf{m}_{11},\cdots,\mathbf{m}_{1J},\cdots,\mathbf{m}_{i1},\cdots,\mathbf{m}_{iJ},\cdots, \mathbf{m}_{I1},\cdots,\mathbf{m}_{IJ}]_{N\times IJ} \\
& = & {\rm D}({\bf A};J) - {\rm D}({\bf B};I) .
\end{eqnarray*}
Then, we have
\begin{eqnarray*}
\widetilde{\bf m}\widetilde{\bf m}^T  & = &  \left(\sum_{ij} {\bf m}_{ij}\right) \left(\sum_{ij} {\bf m}_{ij}\right)^T \\
& = &   \big[{\rm D}({\bf A};J) {\bf 1} - {\rm D}({\bf B};I){\bf 1}\big] \cdot \big[{\rm D}({\bf A};J) {\bf 1} - {\rm D}({\bf B};I){\bf 1}\big]^T\\
& = & {\rm D}({\bf A};J) {\bf 1}  {\bf 1} ^T {\rm D}^T({\bf A};J)   +  {\rm D}({\bf B};I) {\bf 1}  {\bf 1} ^T {\rm D}^T({\bf B};I) \nonumber\\
&&\qquad\qquad\qquad\qquad-   {\rm D}({\bf B};I){\bf 1}    {\bf 1} ^T {\rm D}^T({\bf A};J) - {\rm D}({\bf A};J) {\bf 1} {\bf 1} ^T {\rm D}^T({\bf B};I) \\
& = & J^2 {\bf A}{\bf 1}  {\bf 1} ^T {\rm A}^T   +  I^2 {\bf B}{\bf 1}  {\bf 1} ^T {\rm B}^T  -  IJ {\bf B}{\bf 1}    {\bf 1} ^T {\bf A}^T  -  IJ  {\bf A}{\bf 1}    {\bf 1} ^T {\bf B}^T \\
& = & I^2J^2  \left(  \frac{1}{I^2} {\bf A} {\bf 1} {\bf 1}^T {\bf A}^T  +  \frac{1}{J^2} {\bf B} {\bf 1}{\bf 1}^T{\bf B}^T - \frac{1}{IJ}   {\bf A} {\bf 1} {\bf 1}^T    {\bf B} ^T  - \frac{1}{IJ}   {\bf B} {\bf 1} {\bf 1}^T    {\bf A} ^T \right).
\end{eqnarray*}
It is direct that
\begin{equation*}
I^2J^2 {\bf S}_b = \widetilde{\bf m}\widetilde{\bf m}^T,
\end{equation*}
which implies that the eigenvectors of the two matrices ${\bf S}_b$ and $\widetilde{\bf m}\widetilde{\bf m}^T$ have the same direction.

Moreover, let ${\bf a}$ and ${\bf b}$ stand for the random variables obeying the probability distributions on the sets ${\cal A}$ and ${\cal B}$, respectively. Since ${\bf A} {\bf 1} = I \cdot \widehat{\mathbb{E}}{\bf a} =\sum_{i=1}^I {\bf a}_i$ and ${\bf B} {\bf 1} = J \cdot \widehat{\mathbb{E}}{\bf b} = \sum_{j=1}^J {\bf b}_j$, we have
\begin{eqnarray*}
{\bf S}_w &= & {\bf A}  {\bf A} ^T  + {\bf B}  {\bf B} ^T  - \frac{1}{I}   {\bf A} {\bf 1} {\bf 1}^T    {\bf A} ^T    - \frac{1}{J}   {\bf B} {\bf 1} {\bf 1}^T    {\bf B} ^T\\
&=& {\bf A}  {\bf A} ^T  + {\bf B}  {\bf B} ^T  - \Big[I (\widehat{\mathbb{E}} {\bf a})(\widehat{\mathbb{E}} {\bf a})^T     + J (\widehat{\mathbb{E}} {\bf b})(\widehat{\mathbb{E}} {\bf b})^T \Big].
\end{eqnarray*}
Since $\sum_{i,j}(  {\bf a}_i {\bf b}_j^T + {\bf b}_j {\bf a}_i^T  ) = IJ \cdot \widehat{\mathbb{E}}\{ {\bf a}{\bf b}^T +{\bf b}{\bf a}^T  \}$, we have
\begin{eqnarray*}
{\bf M} {\bf M}^T  & = &  \big[{\rm D}({\bf A};J) - {\rm D}({\bf B};I) \big]   \big[{\rm D}({\bf A};J) - {\rm D}({\bf B};I) \big] ^T\\
& = &  J {\bf A} {\bf A}^T  +  I {\bf B} {\bf B}^T -  \sum_{i,j}(  {\bf a}_i {\bf b}_j^T + {\bf b}_j {\bf a}_i^T  )\\
& = & J {\bf A} {\bf A}^T  +  I {\bf B} {\bf B}^T - IJ \cdot \widehat{\mathbb{E}}\{ {\bf a}{\bf b}^T +{\bf b}{\bf a}^T  \}.
\end{eqnarray*}
This completes the proof. \hfill  $\blacksquare$


\section{Conclusion}
 
In general, the properties of deep neural networks are studied based on the backward inference from the network performance. Since neural networks have multilayer composite structures, it is technically difficult to analyze the function relation between the network performance and network structure parameters. To the best of our knowledge, it is still challenging to find applicable tools to directly analyze the effect of network structure parameters on the network performance.

In this paper, we first demonstrate that there exists a synchronicity between the linear separability of hidden layers and the training accuracy in the classification tasks. This finding suggests that the linear separability should be an applicable tool to layer-wisely explore the characteristics of deep networks. Then, we propose the MD-LSM ${\rm LS}_*$ to measure the linear separability degree of the outputs of each hidden layer. To alleviate the computation cost, we present some alternative versions of MD-LSMs including ${\rm LS}_0$, ${\rm LS}_1$ and ${\rm LS}_2$. The last one has a closed-form solution via eigenvalue decomposition. In the numerical experiment, we simplify the constraint of the optimization problem \eqref{eq:ls2} to achieve an approximation solution $\bm{\omega}$ to these MD-LSMs with a low computation cost ({\it cf.} Remark \ref{rem:approx}). 

We also analyze the relationship between the network structure parameters and the network performance. First, we provide a sufficient condition that a hidden layer can change the relative position relation between an MD point and the splitting hyperplane linear. Based on random matrix theory, we then demonstrate that the increase of network size can enlarge the probability of increasing the linear separability of hidden layers. The experimental results support our theoretical findings, and show that the synchronicity between the linear separability and network performance widely exists in the process of training popular deep networks (such as MLP, CNN, DBN, ResNet, VGGNet, AlexNet, ViT and GoogLeNet) in binary and multi-class classification tasks. In further works, we will use the MD-LSMs to measure the complexity of deep networks and analyze their generalization performance.



\appendix


\section{Appendix: Related Works}\label{supp:related}


Following the classical statistical learning theory techniques, it is expected to build a proper complexity measure of neural networks, and then to develop the relevant generalization bounds. Neyshabur {\it et al.} \cite{neyshabur2015norm} proposed a norm-based framework for measuring the capacity of neural networks, and the measure is expressed as an exponential form of the network depth. Golowich {\it et al.} \cite{golowich2020size} provided the generalization bound based on Rademacher complexity that is bounded by multiplying the weight norm of each layer, and thus such a bound is also of the exponential form of the network depth. Harvey {\it et al.} \cite{harvey2017nearly} used the Vapnik-Chervonenkis (VC) dimension as the complexity measure of the networks with piecewise linear activation functions, and then obtain the generalization bound which can be regarded as a linear function w.r.t. the network depth, the number of parameters and the number of nonlinear units. Bartlett \cite{bartlett2017spectrally} defined the margin-normalized spectral complexity, which is expressed as the product of the spectral norms of all weight matrices, and then provided the generalization bound based on covering number of neural networks. These results illustrates that the generalization performance of neural networks will become low when the network size increases. It accords with the traditional view of statistical learning theory but is still far away from the empirical obversion on the applications of deep neural networks.

By combining the spectral norm and the Frobenius norm of the weight matrices, Neyshabur {\it et al.} \cite{neyshabur2018pac} obtained the PAC-Bayes generalization bound of the networks with ReLU activation functions and then analyzed the stability of the network against the small perturbations on the weights. Under the PAC-Bayes framework, some works proposed the non-vacuous bounds of deep networks and these results were not dependent on the size or the dimension of parameters. Dziugaite {\it et al.} \cite{dziugaite2017computing} showed a data-dependent bound, derived from the result of \cite{mcallester1999pac}, for the deep classifiers trained by SGD method, and then optimized the resultant bound to find a posterior distribution starting from the local minima found by the previous SGD process. Zhou {\it et al.} \cite{zhou2018non} studied the generalization performance of compressed deep neural networks to explain a phenomenon that compressing a deep network only causes a slight performance change. These works usually assume that the prior or the posterior distribution of model parameters obeys Gaussian or sub-Gaussian, but this assumption may not always hold in practice.

The neural tangent kernel (NTK) method, proposed in \cite{jacot2018neural}, suggests that the process of training an infinitely wide neural network should be equivalently as a kernel regression process. Based on NTK, Du {\it et al.} \cite{du2019gradient} proved that the training error of infinitely wide neural networks of any depth will finally converge to zero. Arora {\it et al.} \cite{arora2019fine} provided a generalization bound for the two-layer ReLU network by bounding the corresponding NTK. However, since these works assume that the network width is infinite, it is still far away from the actual cases. Moreover, under the same assumption, some works also studied the theoretical properties of the over-parameterized neural networks ({\it e.g.,} two-layer ReLU networks) during the SGD training process \cite{li2018learning,du2018gradient,song2019quadratic}. To sum up, there still remains a big gap between the recent research achievement and the comprehensive explainability of deep neural networks.

\section{Appendix: Complete Experimental Report} \label{supp:experiment}

In this part, we provide the experimental results of three kinds of MD-LSMs: ${\rm LS}_0$, ${\rm LS}_1$ and ${\rm LS}_2$. In view of the complicated structures of VGGNet, ResNet-20, GoogLeNet-V1 and ViT, we also draw the structure diagrams to denote their hidden layers or main blocks. 
\vspace{-3mm}

\subsection{MLP}

\begin{figure}[htbp]
	\centering
	 
	\subfigure[$ {\rm LS}_0$]{
		\includegraphics[width=0.4\textwidth]{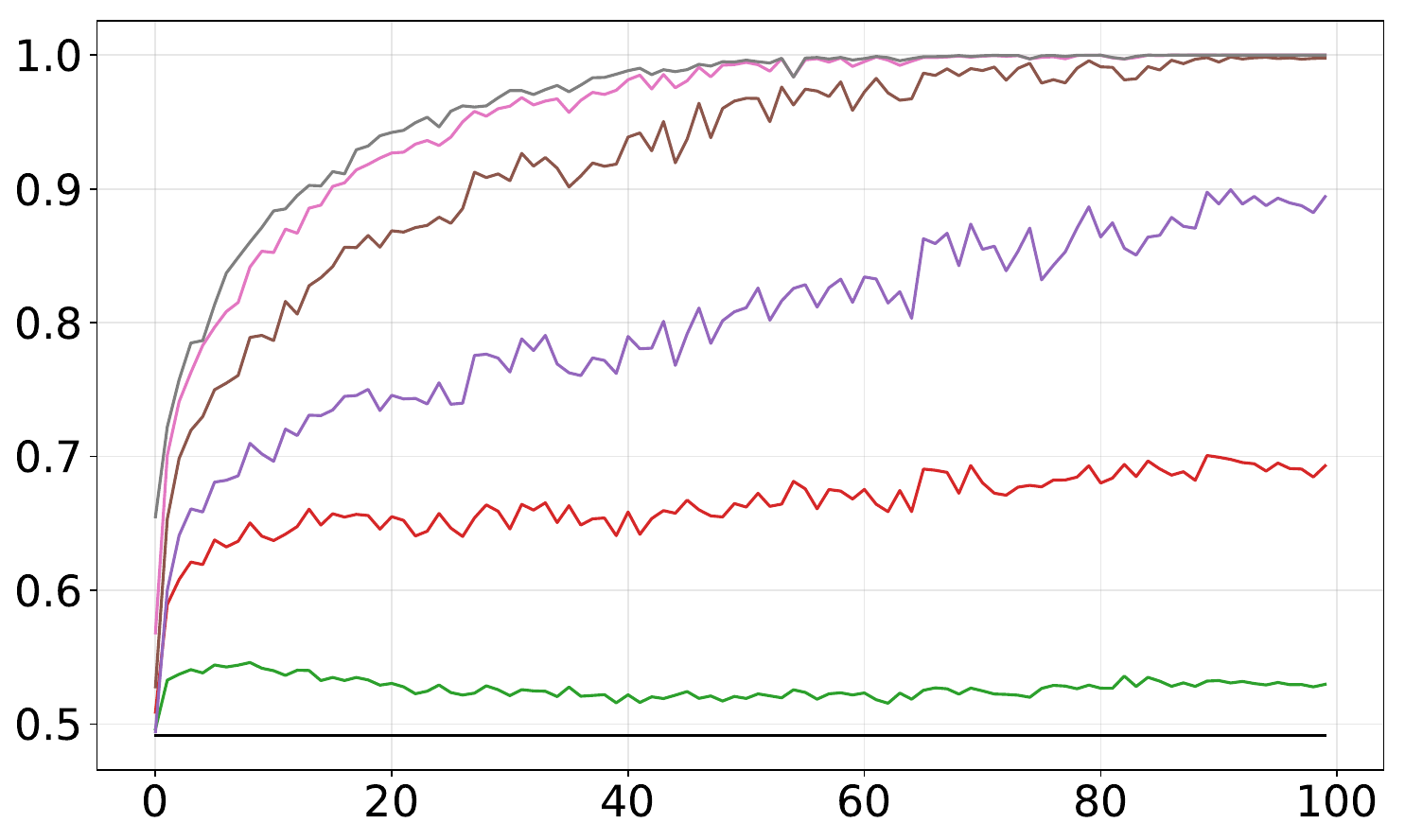}
	}
	\subfigure[$ {\rm LS}_1$]{
		\includegraphics[width=0.4\textwidth]{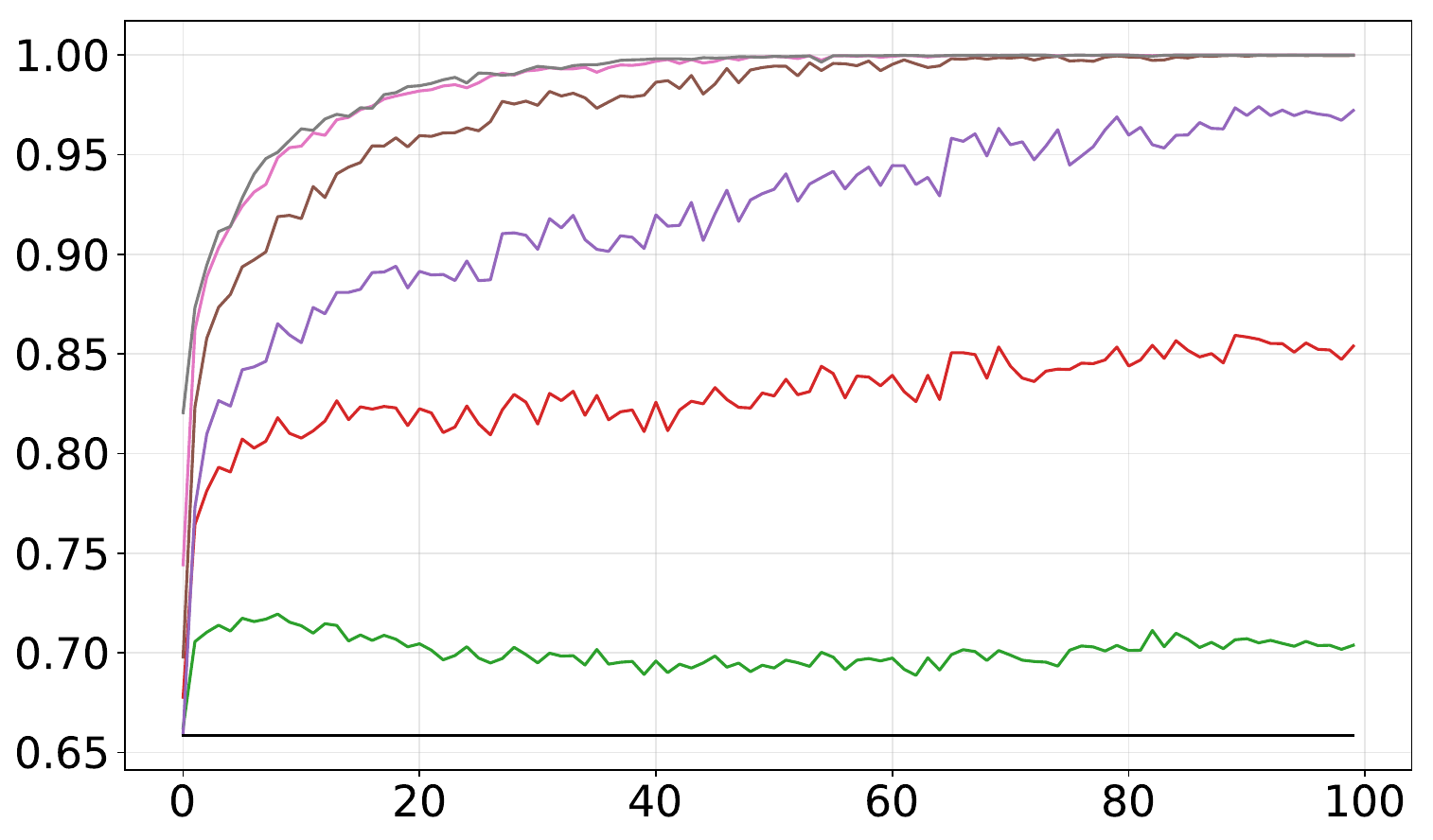}
	}
	\vskip -0.1in
	\subfigure[$ {\rm LS}_2$]{
		\includegraphics[width=0.4\textwidth]{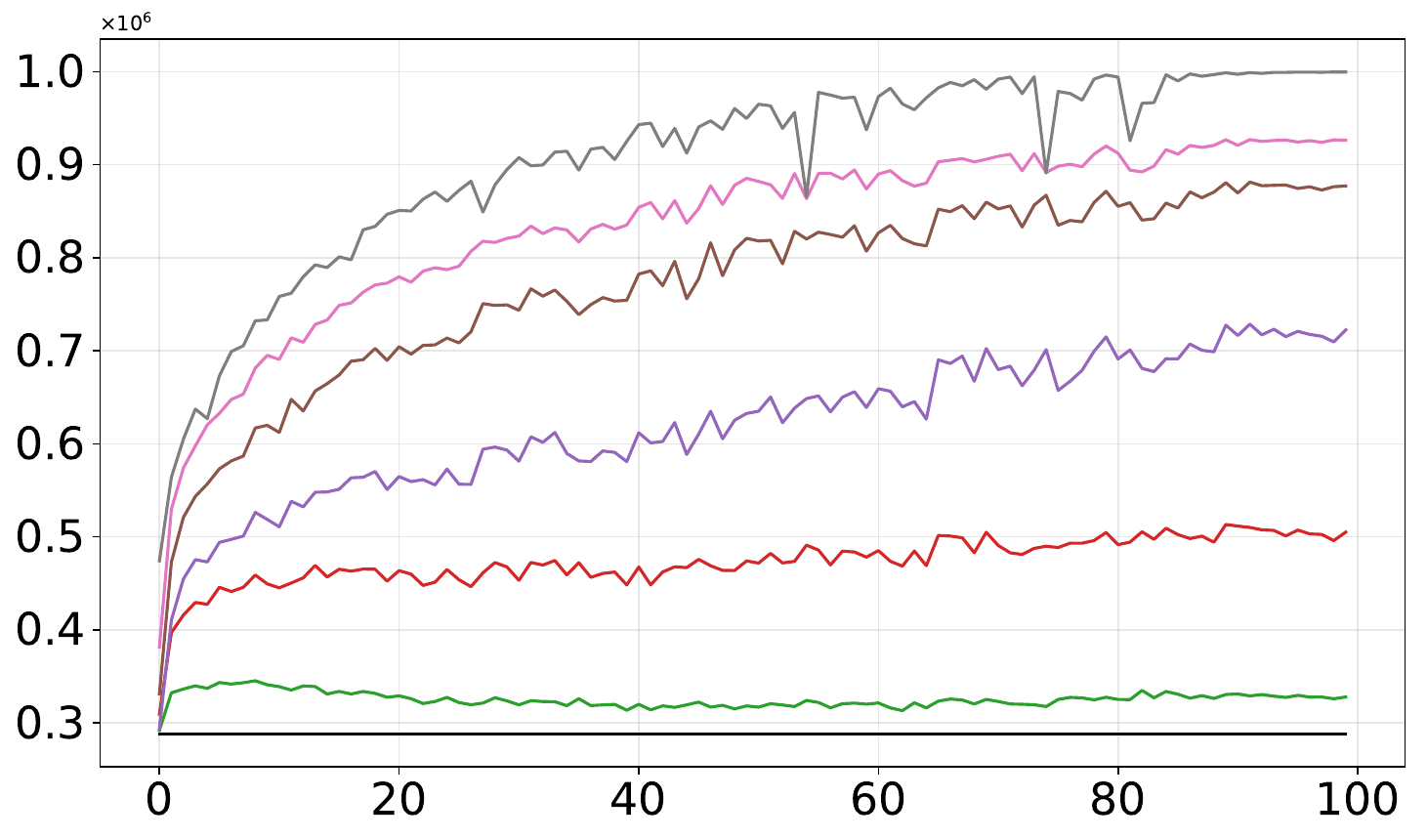}
	}
	\subfigure[Accuracy]{
		\includegraphics[width=0.4\textwidth]{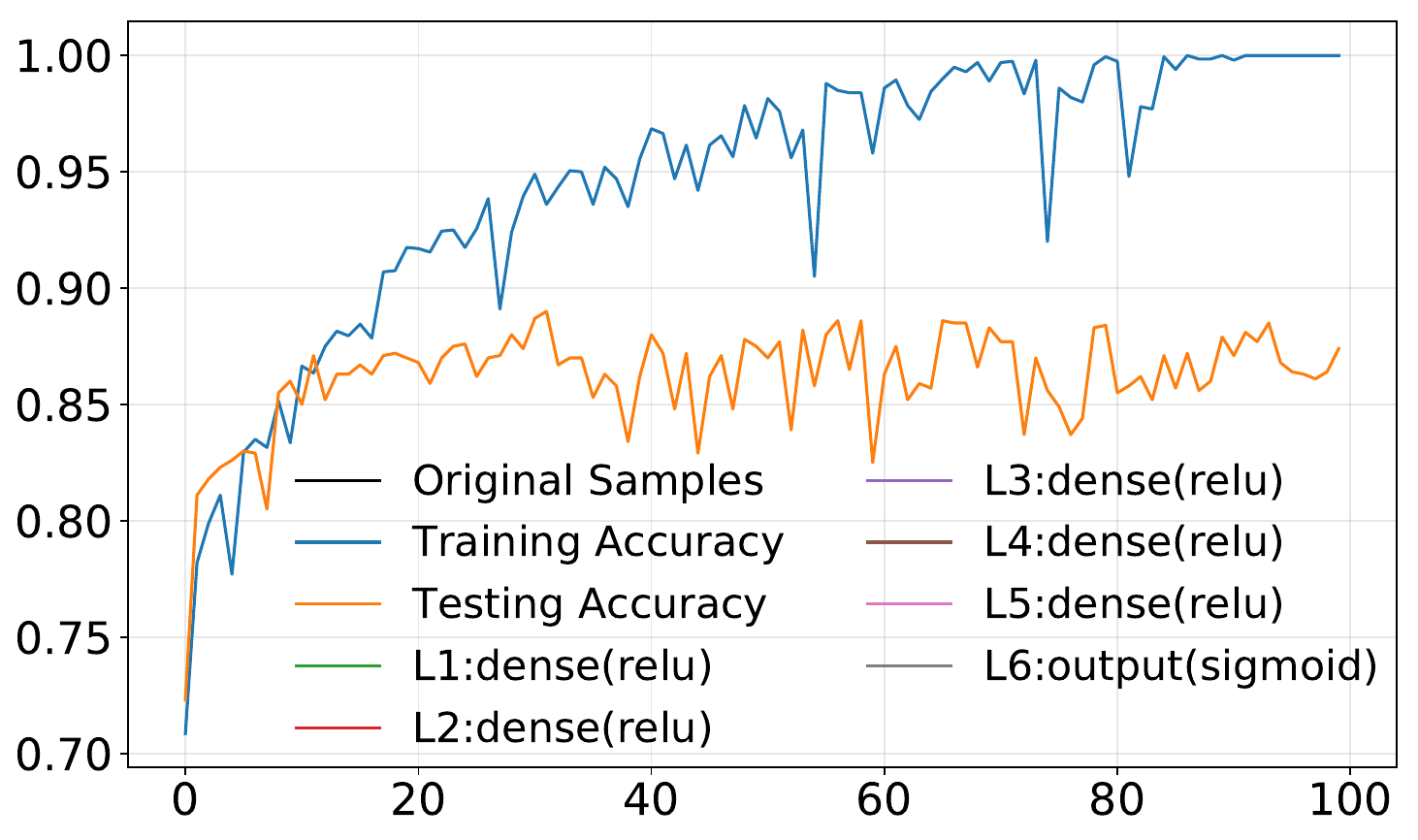}
	}
	 \vskip -0.1in
	\caption{MD-LSM and Accuracy Curves of Hidden Layers of MLP-5 (ReLU)}
	\label{fig:mlp-5-relu}
	\end{figure}

\begin{figure}[htbp]
	\centering	
	\subfigure[$ {\rm LS}_0$]{
		\includegraphics[width=0.4\textwidth]{figure/all-sigmoid_LS0.pdf}
	}
	\subfigure[$ {\rm LS}_1$]{
		\includegraphics[width=0.4\textwidth]{figure/all-sigmoid_LS1.pdf}
	}
	\subfigure[$ {\rm LS}_2$]{
		\includegraphics[width=0.4\textwidth]{figure/all-sigmoid_JW.pdf}
	}
	\subfigure[Accuracy]{
		\includegraphics[width=0.4\textwidth]{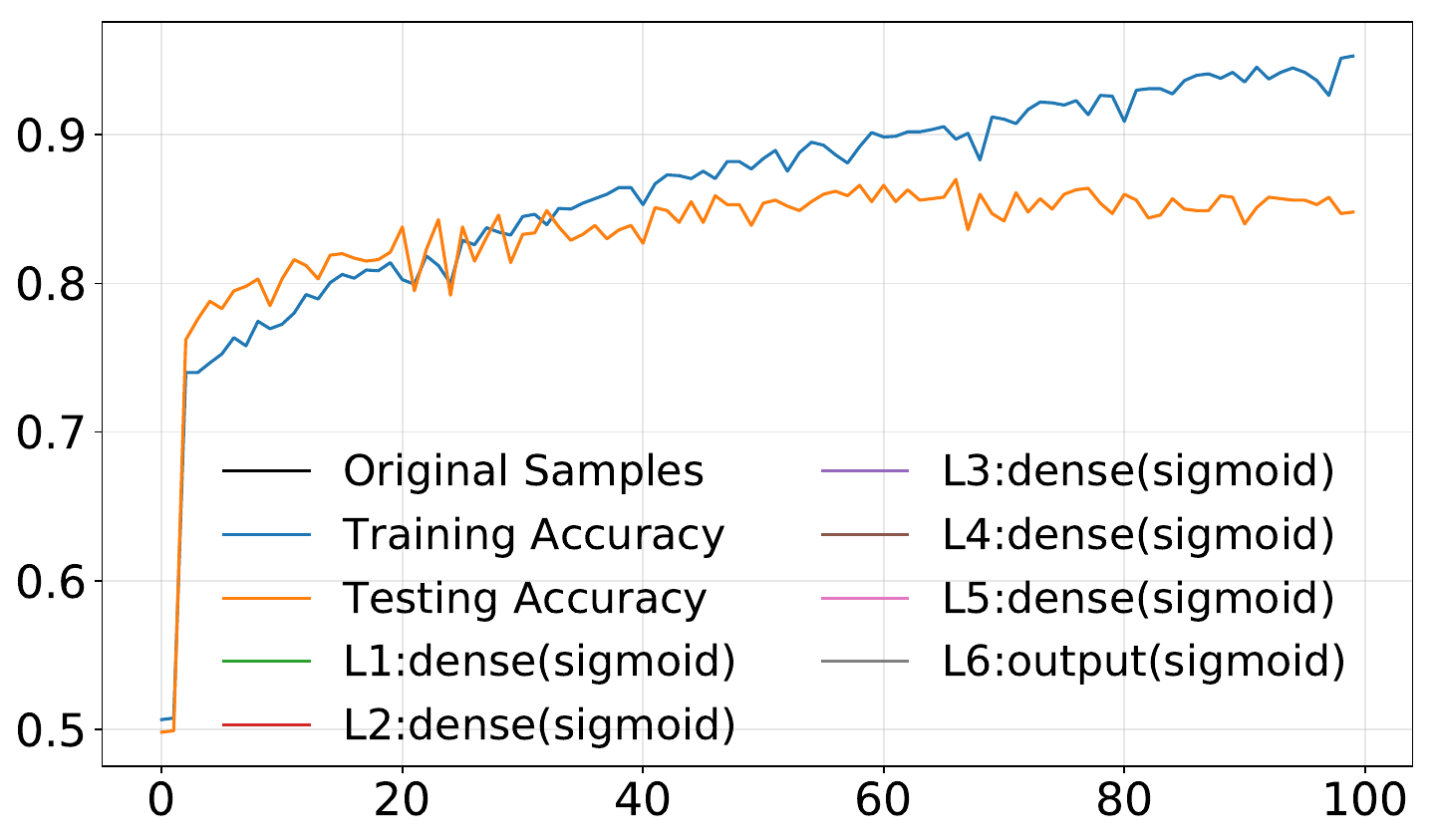}
	}
	 
	\caption{MD-LSM and Accuracy Curves of Hidden Layers of  MLP-5 (Sigmoid)}
	\label{fig:mlp-5-sigmoid}	
	\end{figure}

\begin{figure}[htbp]
	\centering	
	
	\subfigure[$ {\rm LS}_0$]{
		\includegraphics[width=0.4\textwidth]{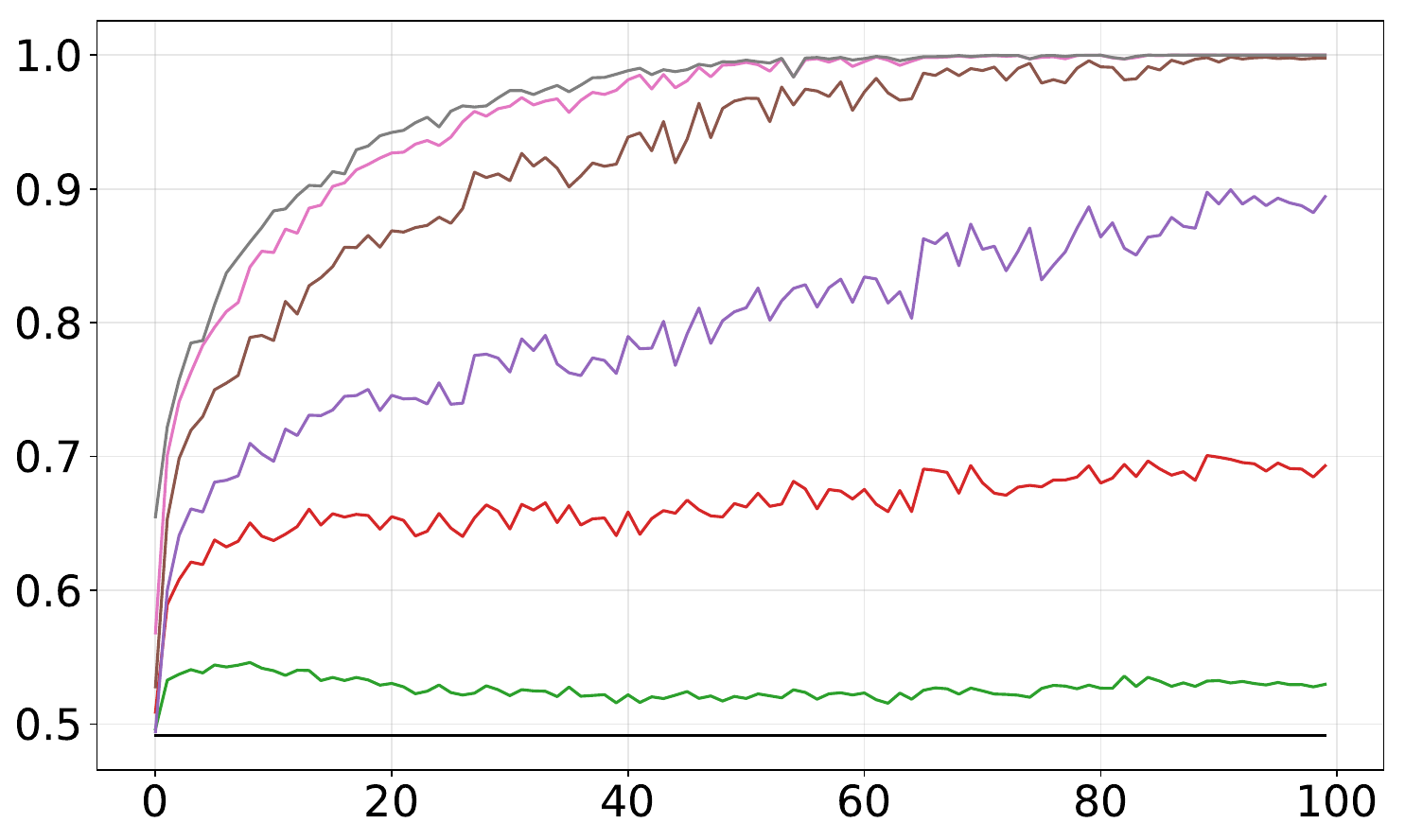}
	}
	\subfigure[$ {\rm LS}_1$]{
		\includegraphics[width=0.4\textwidth]{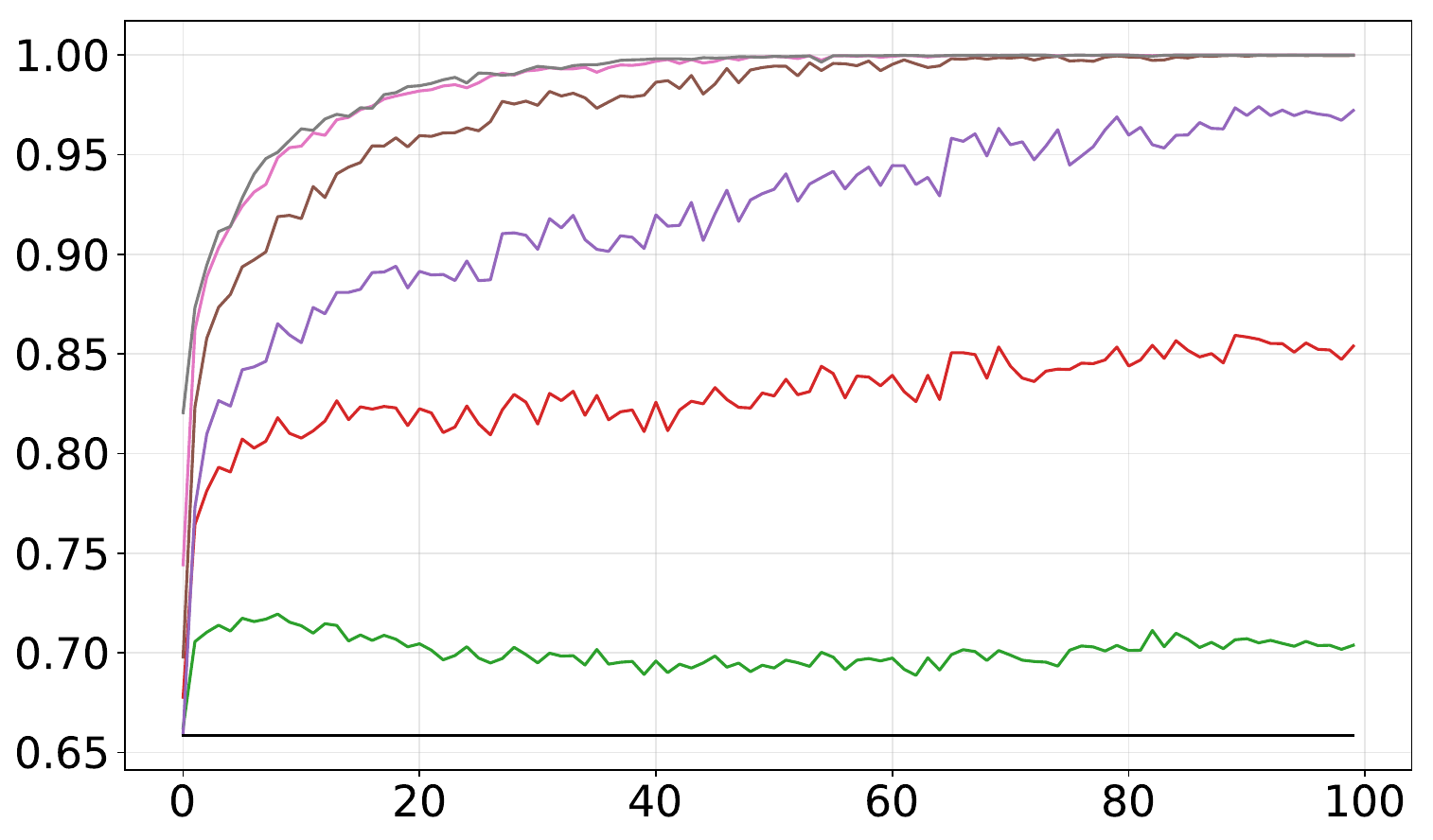}
	}
	\subfigure[$ {\rm LS}_2$]{
		\includegraphics[width=0.4\textwidth]{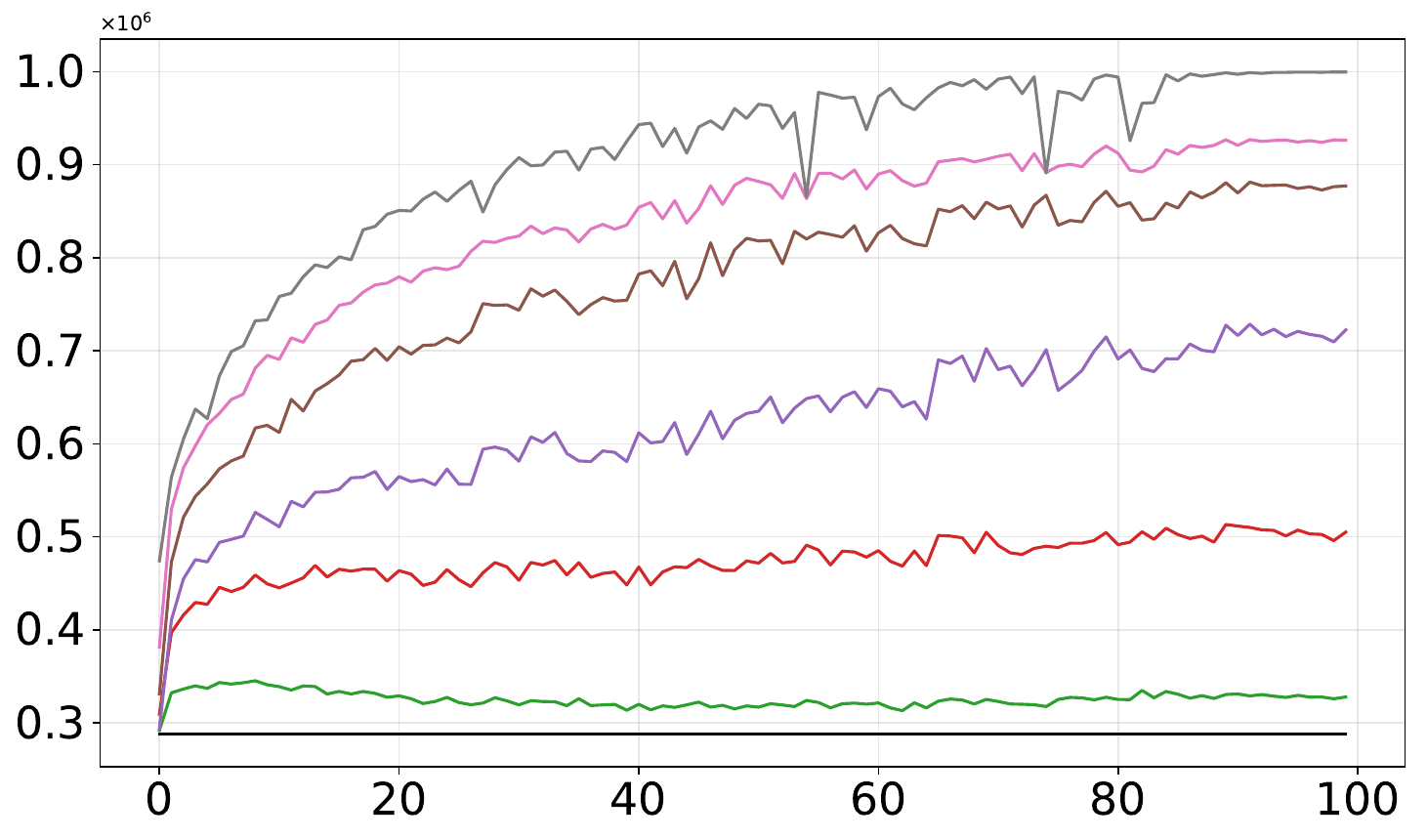}
	}
	\subfigure[Accuracy]{
		\includegraphics[width=0.4\textwidth]{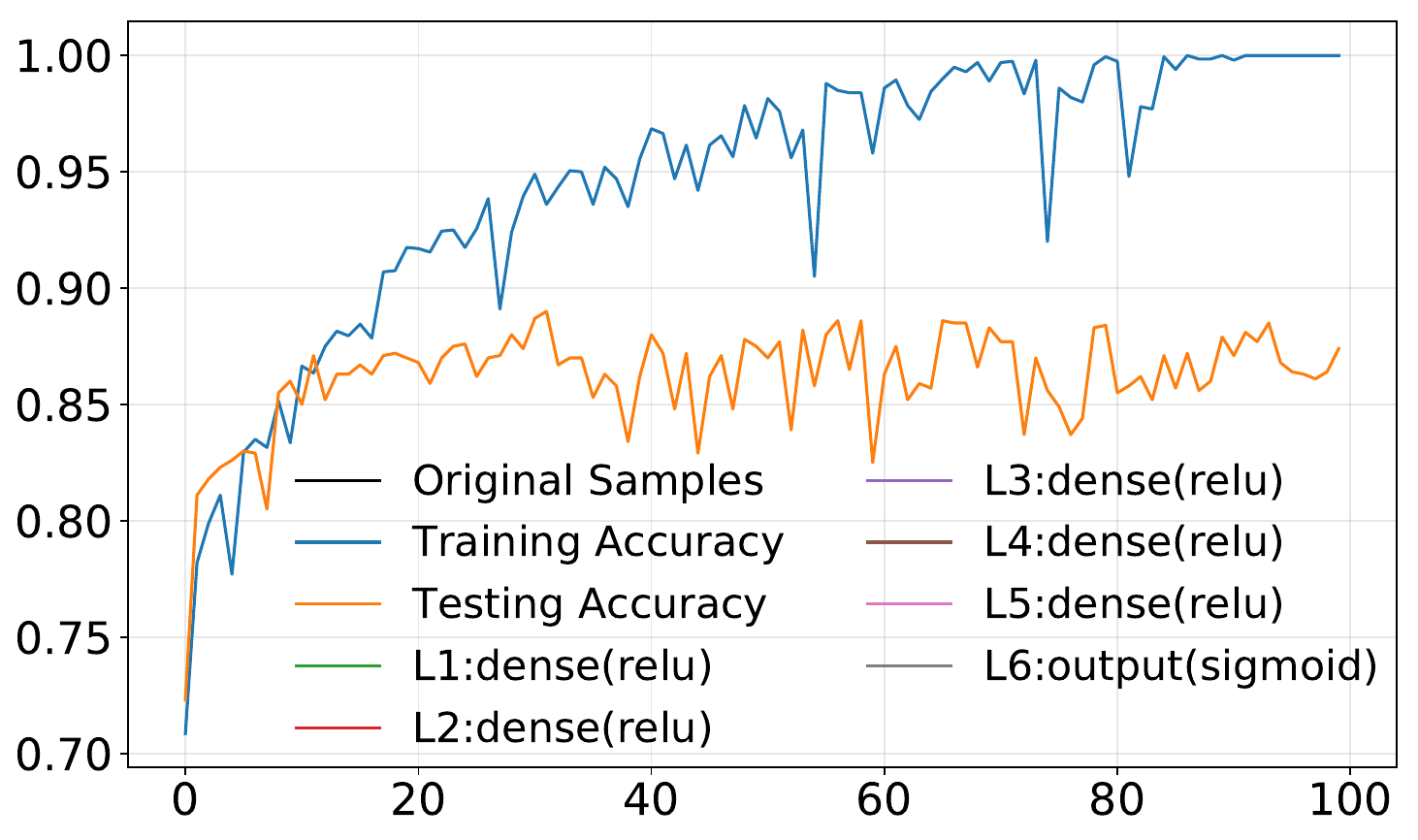}
	}
	 
	\caption{MD-LSM and Accuracy Curves of Hidden Layers of MLP-10 (ReLU)}
	\label{fig:mlp-10-relu}

		\end{figure}

\begin{figure}[htbp]
	\centering	

	\subfigure[$ {\rm LS}_0$]{
		\includegraphics[width=0.4\textwidth]{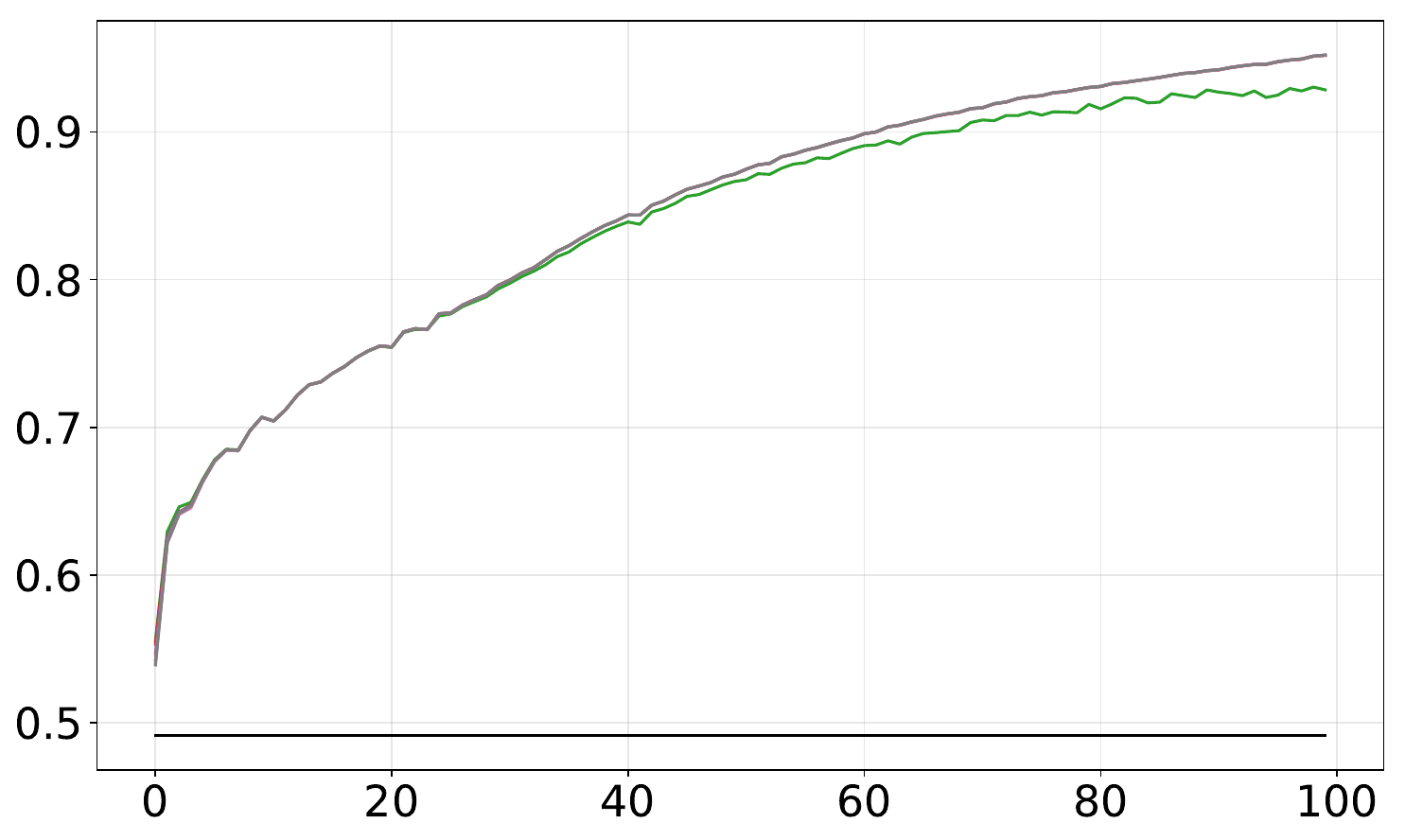}
	}
	\subfigure[$ {\rm LS}_1$]{
		\includegraphics[width=0.4\textwidth]{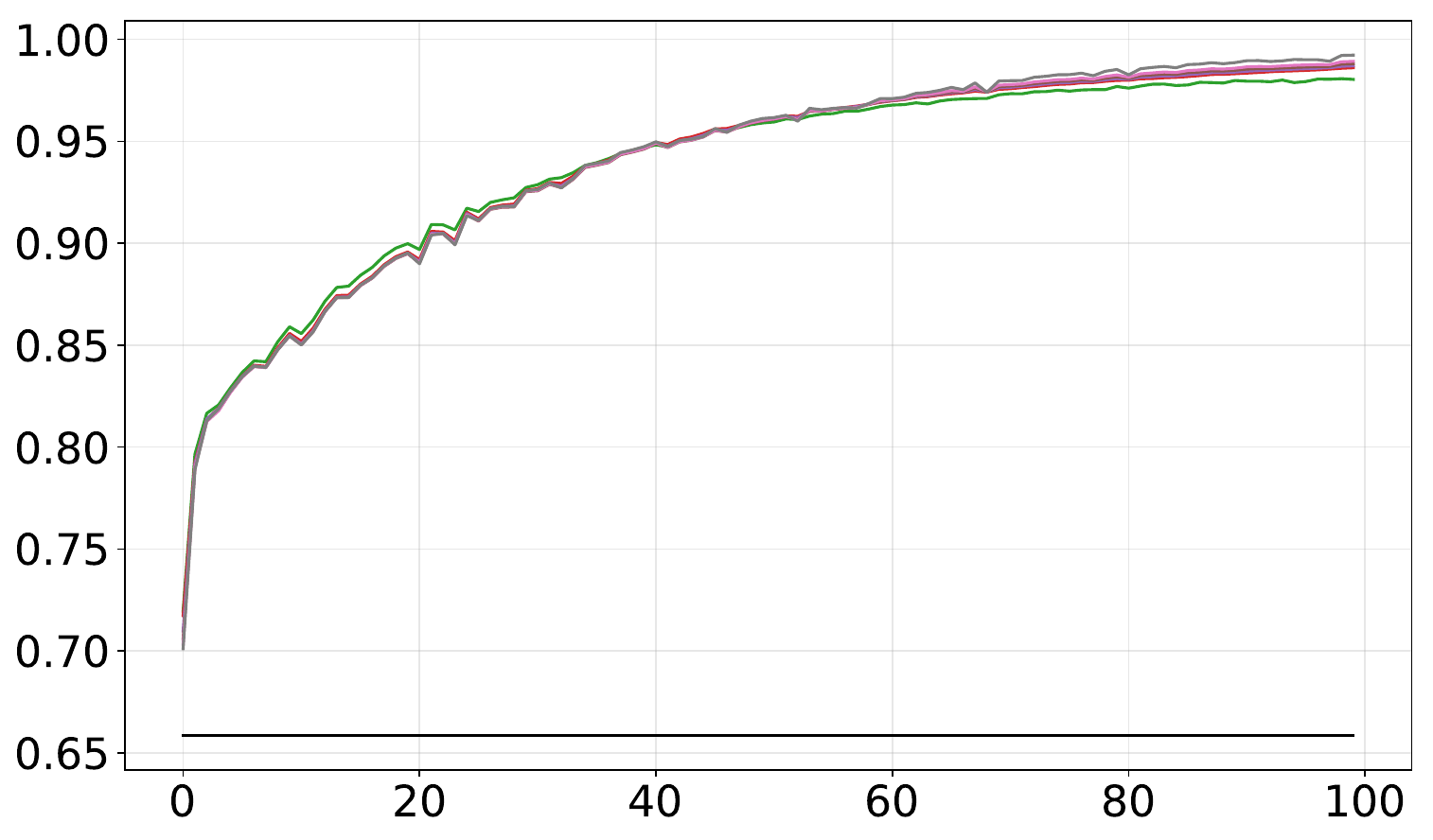}
	}
	\subfigure[$ {\rm LS}_2$]{
		\includegraphics[width=0.4\textwidth]{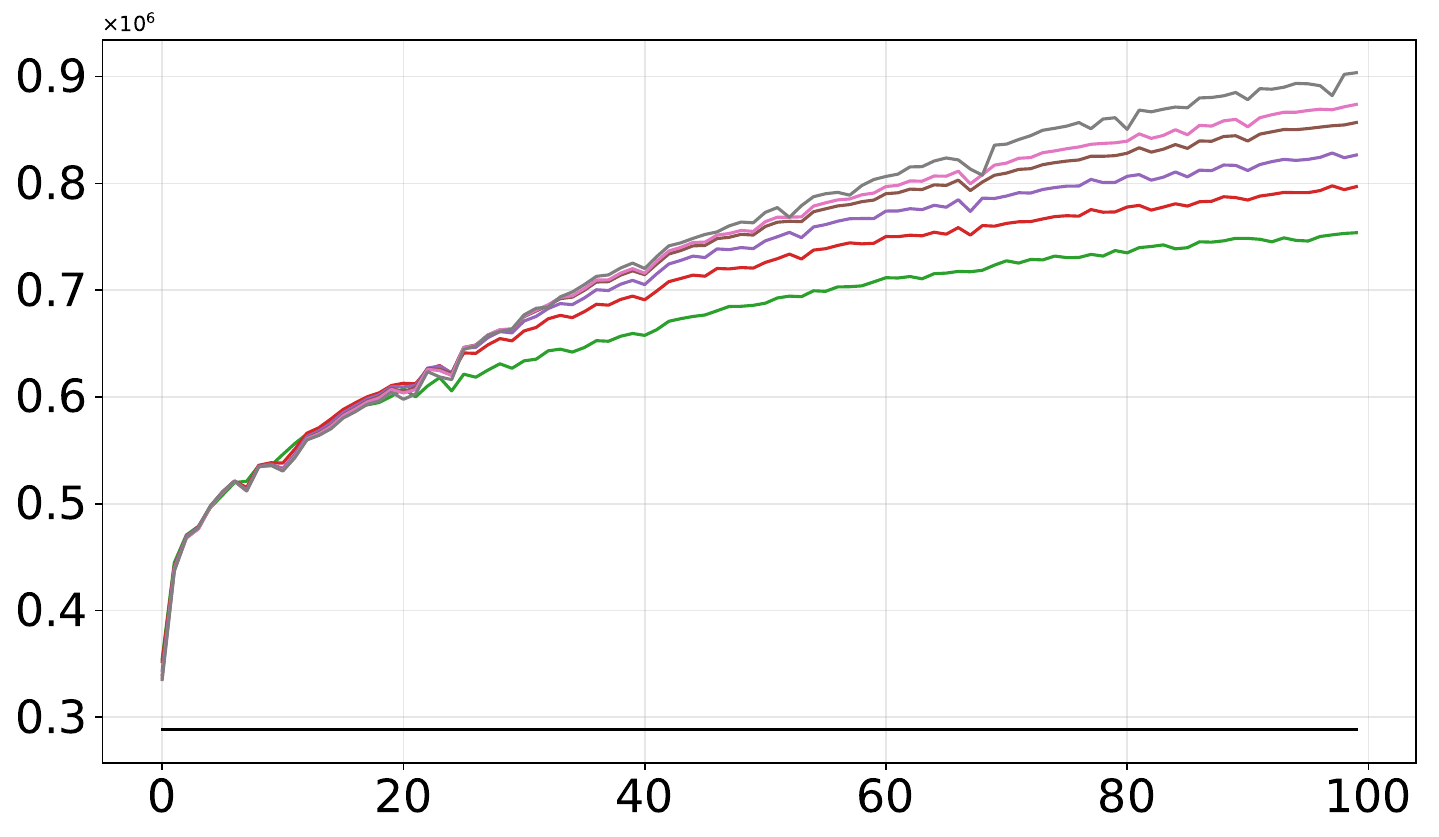}
	}
	\subfigure[Accuracy]{
		\includegraphics[width=0.4\textwidth]{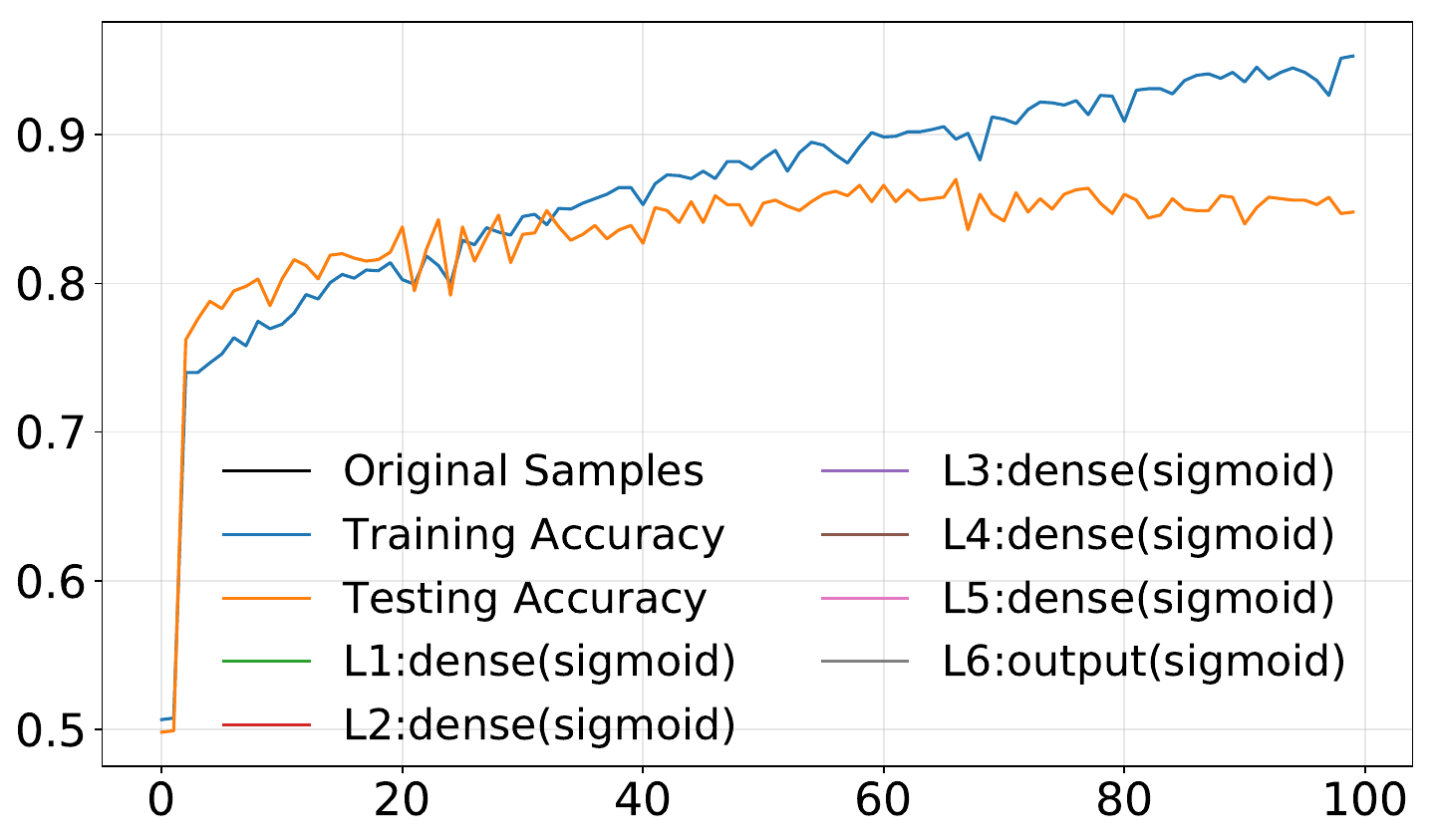}
	}
	 
	\caption{MD-LSM and Accuracy Curves of Hidden Layers of MLP-10 (Sigmoid)}
	\label{fig:mlp-10-sigmoid}		
		\end{figure}

\begin{figure}[htbp]
	\centering

	\subfigure[$ {\rm LS}_0$]{
		\includegraphics[width=0.4\textwidth]{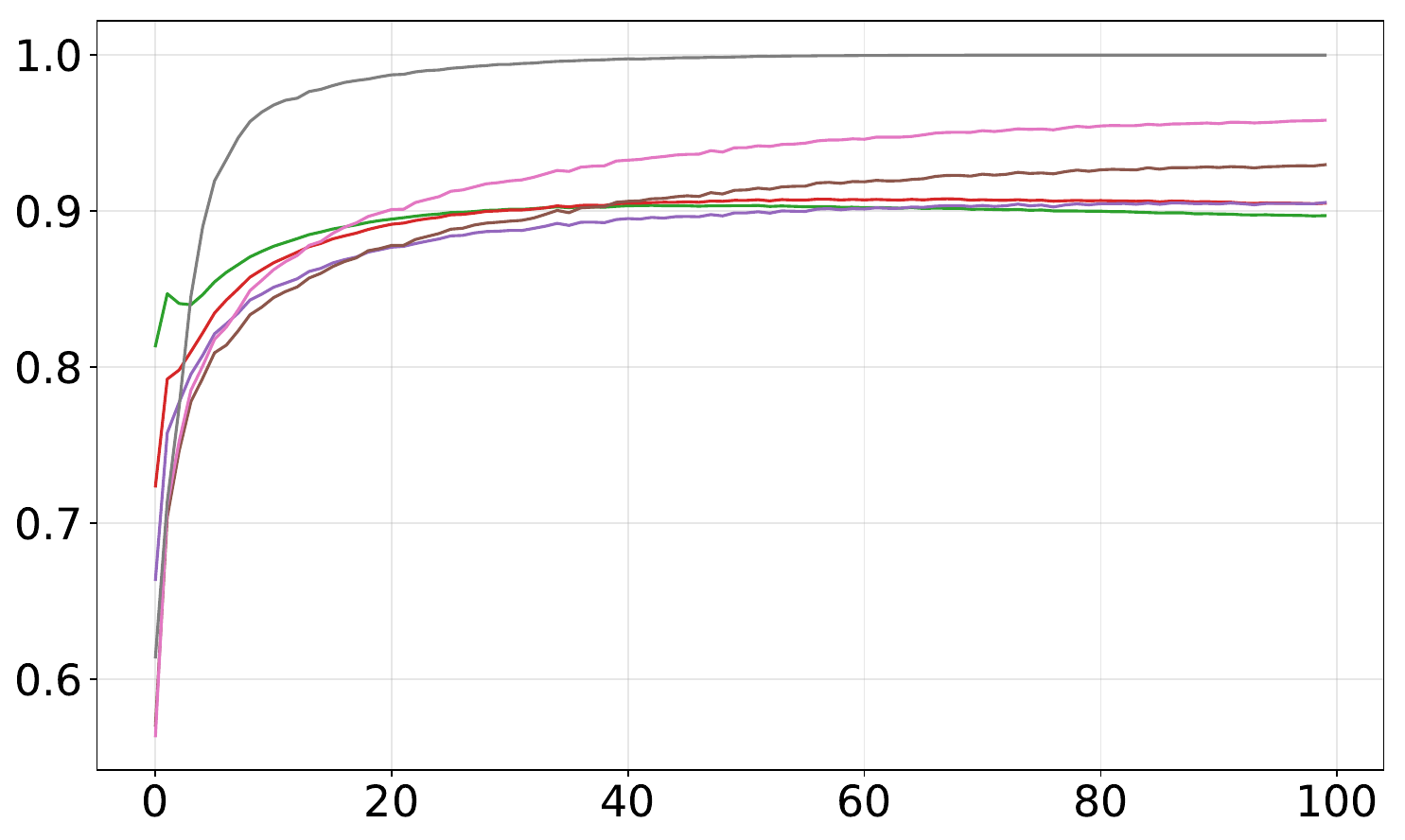}
	}
	\subfigure[$ {\rm LS}_1$]{
		\includegraphics[width=0.4\textwidth]{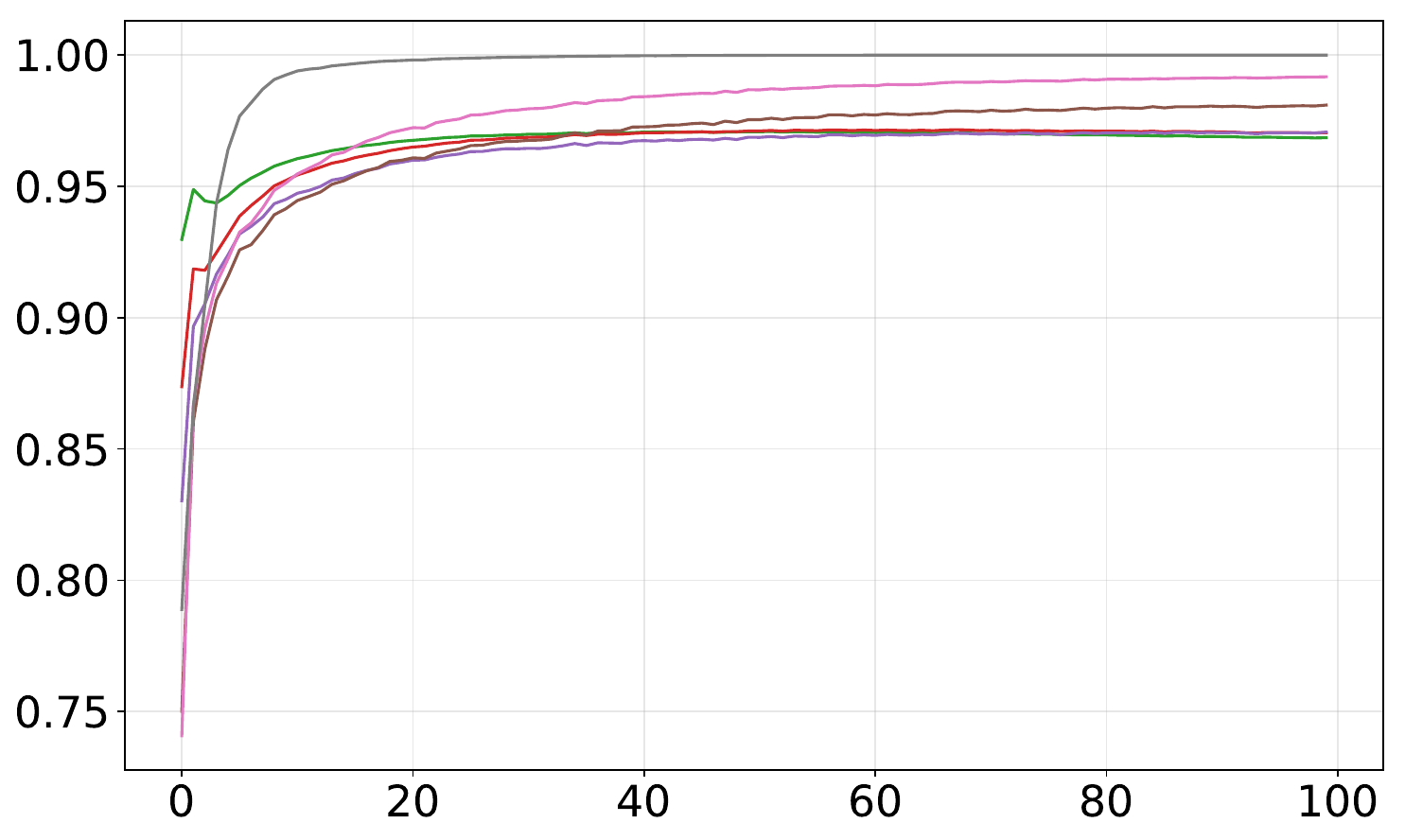}
	}
	\subfigure[$ {\rm LS}_2$]{
		\includegraphics[width=0.4\textwidth]{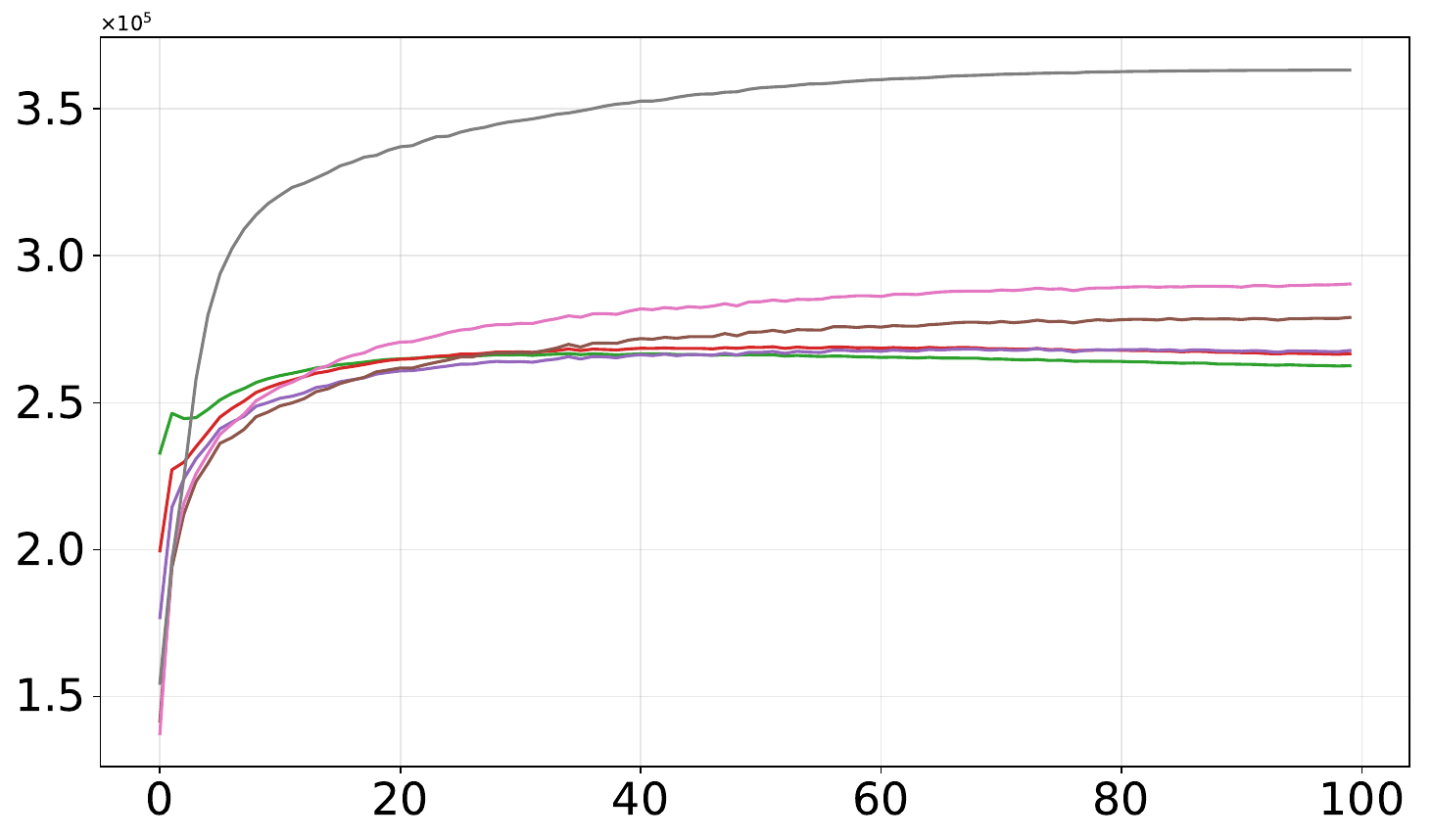}
	}
	\subfigure[Accuracy]{
		\includegraphics[width=0.4\textwidth]{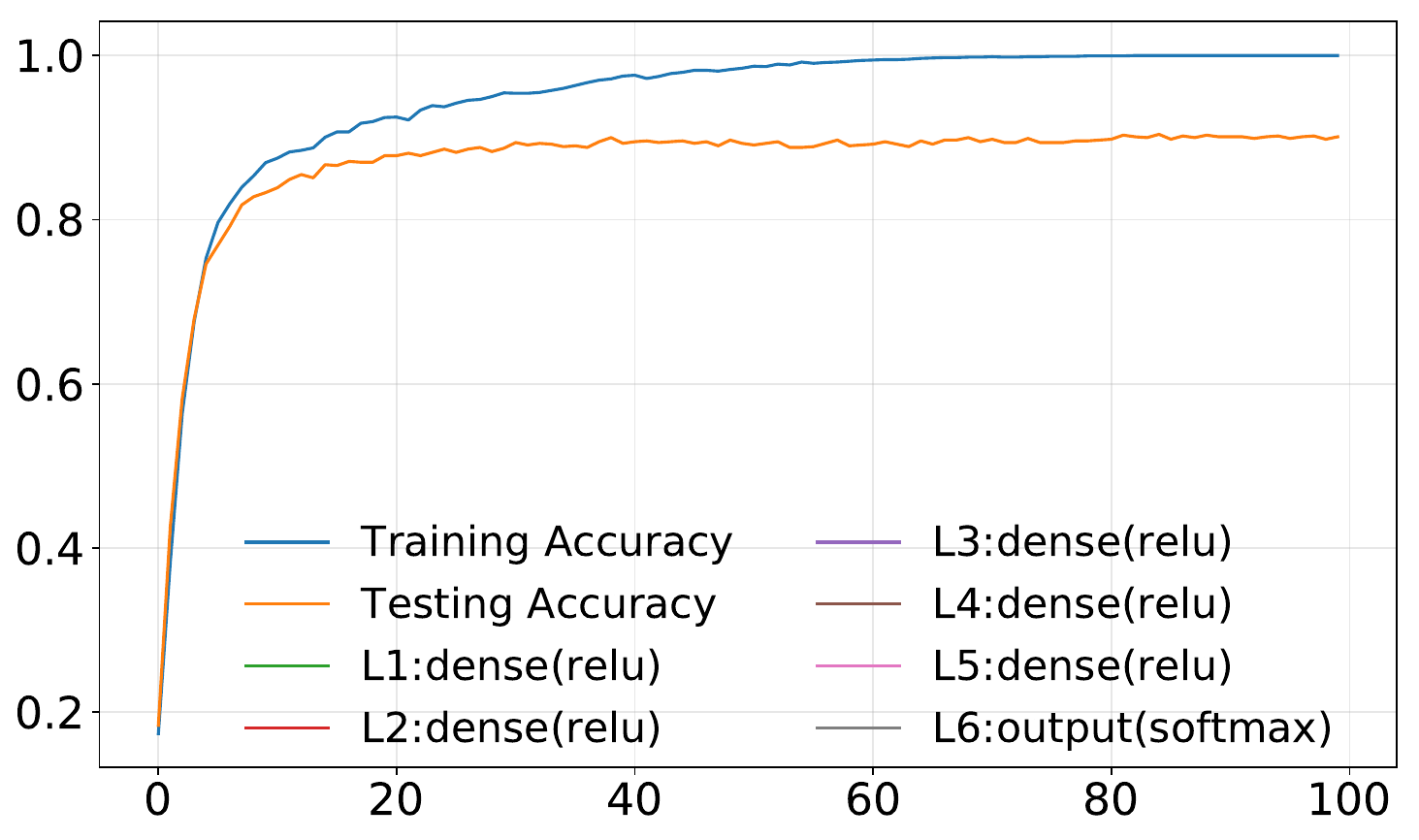}
	}
	 
	\caption{MD-LSM and Accuracy Curves of Hidden Layers of MLP (Ten-Class)}
	\label{fig:mlp-mnist}		
\end{figure}

First, we layer-wisely examine the linear separability of the MLPs with five hidden layers, denoted as MLP-5, and ten hidden layers, denoted as MLP-10, respectively. The hidden nodes of MLPs are activated by using Sigmoid functions (denoted as Sigmoid) and ReLU functions (denoted as ReLU), respectively. In Figs. \ref{fig:mlp-5-relu}--\ref{fig:mlp-10-sigmoid}, we illustrate the experimental results of MLPs in the binary classification tasks. In addition, we also consider the linear separability of MLPs in ten-class classification task, where the network has five hidden layers and its hidden nodes are activated by using ReLU ({\it cf.} Fig. \ref{fig:mlp-mnist}).

\subsection{CNN, AlexNet and DBN}

\begin{figure}[htbp]

	\centering	
	\subfigure[$ {\rm LS}_0$]{
		\includegraphics[width=0.4\textwidth]{figure/all-cnn_LS0.pdf}
	}
	\subfigure[$ {\rm LS}_1$]{
		\includegraphics[width=0.4\textwidth]{figure/all-cnn_LS1.pdf}
	}
	\subfigure[$ {\rm LS}_2$]{
		\includegraphics[width=0.4\textwidth]{figure/all-cnn_JW.pdf}
	}
	\subfigure[Accuarcy]{
		\includegraphics[width=0.4\textwidth]{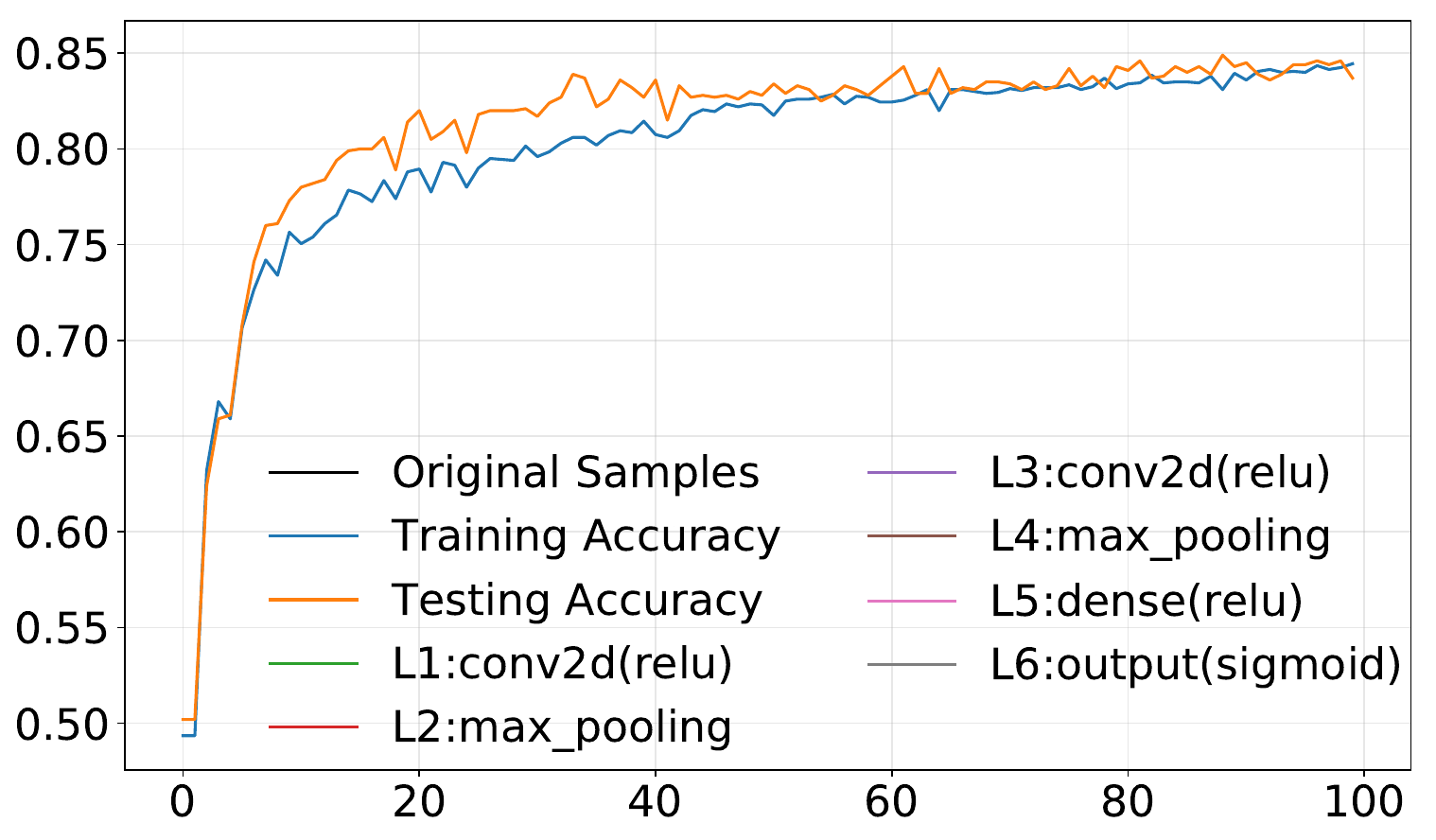}
	}
	 
	\caption{MD-LSM and Accuracy Curves of CNN's Hidden Layers}
	\label{fig:CNN}	
\end{figure}

\begin{figure}[htbp]

	\centering

	\subfigure[$ {\rm LS}_0$]{
		\includegraphics[width=0.4\textwidth]{figure/all-alexnet_LS0.pdf}
	}
	\subfigure[$ {\rm LS}_1$]{
		\includegraphics[width=0.4\textwidth]{figure/all-alexnet_LS1.pdf}
	}
	\subfigure[$ {\rm LS}_2$]{
		\includegraphics[width=0.4\textwidth]{figure/all-alexnet_JW.pdf}
	}
	\subfigure[Accuracy]{
		\includegraphics[width=0.4\textwidth]{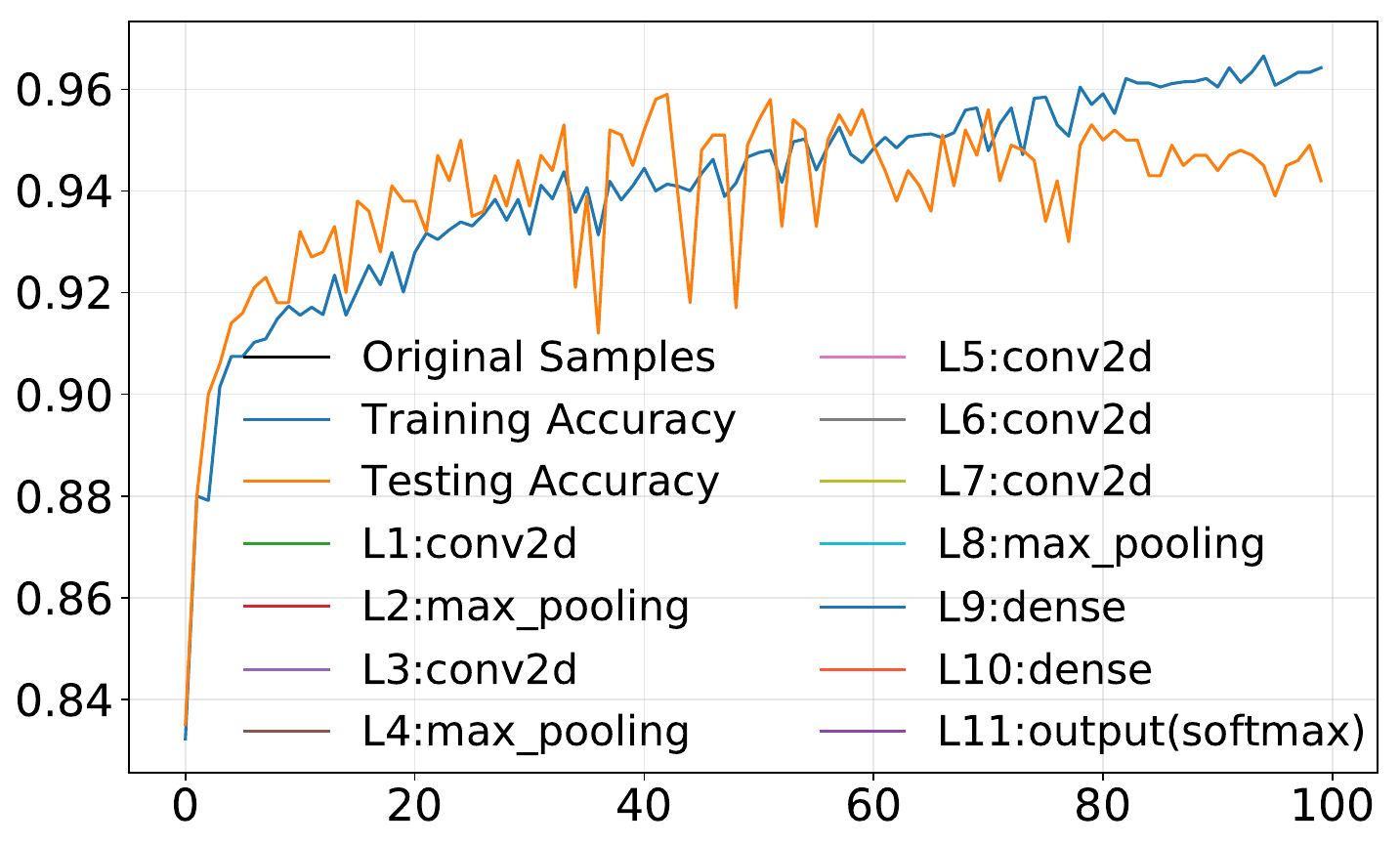}
	}
	 
	\caption{MD-LSM and Accuracy Curves of Hidden Layers of AlexNet (Softmax)}
	\label{fig:AlexNet-softmax}				
	
	\end{figure}

Moreover, we examine the linear separability of CNNs with two convolution layers and two pooling layers in binary classification task. All hidden nodes of CNNs are activated by using ReLU ({\it cf.} Fig. \ref{fig:CNN}). Moreover, the linear separability of AlexNet and DBN is also considered in the same task ({\it cf.} Figs. \ref{fig:AlexNet-softmax}--\ref{fig:dbn}). It is noteworthy that we consider two kinds of AlexNets that have different output activation functions: one is Softmax, denoted as AlexNet (Softmax), and the other is Sigmoid, denoted as AlexNet (Sigmoid). We also simplify the process of training AlexNet (Sigmoid), where the tricks of learning rate decay and data augmentation are not used. Since the learning task (binary classification) is much simpler than the task (ImageNet classification) for which AlexNet was originally designed, the simplified training process is enough to provide a good performance. Thus, the curves of AlexNet (Sigmoid) are smoother than those of AlexNet (Softmax), especially for the ${\rm LS}_2$.

\subsection{VGGNet, GoogLeNet, ResNet and ViT}\label{sec:deepnet}

Here, we consider the linear separability of the deep networks with complicated hidden-layer structures, including VGGNet, GoogLeNet-V1, ResNet-20 and ViT. Since the structures of these networks can be split into some individual blocks, we first examine the linear separability of the outputs of their main blocks, and then illustrate the LSMs of hidden layers of these networks.

In Tab. \ref{tab:arrange}, we show the arrangement of the structure diagrams and the experimental results.
\vskip -0.1in
\begin{table}[htbp]
	\caption{Numerical Experiment Results}\label{tab:arrange}
	\centering
	\resizebox{0.8\textwidth}{!}{\begin{tabular}{c |c|c| c }
		\hline
		Deep Networks & Structure Diagram &  Main Blocks    & Hidden Layers   \\
		\hline 
		MLP-5 (ReLU)  & & &  Fig. \ref{fig:mlp-5-relu} \\
		\hline
		MLP-5 (Sigmoid) &&&
		Fig. \ref{fig:mlp-5-sigmoid}\\
		\hline
		MLP-10 (ReLU)&&& Fig. \ref{fig:mlp-10-relu}\\
		\hline
		MLP-10 (Sigmoid)
		&&& Fig. \ref{fig:mlp-10-sigmoid}\\
		\hline
		MLP (Ten-Class)&&& Fig. \ref{fig:mlp-mnist} \\
		\hline
		CNN&&& Fig. \ref{fig:CNN} \\
		\hline
		AlexNet (Softmax)
		&&& Fig. \ref{fig:AlexNet-softmax}\\
		\hline
		AlexNet (Sigmoid)
		&&& Fig. \ref{fig:AlexNet-sigmoid}\\
		\hline
		DBN&&& Fig. \ref{fig:dbn}\\
		\hline
		VGGNet & Fig. \ref{fig:vgg-structure}   &  Fig. \ref{fig:vgg-block}    & Fig. \ref{fig:vgg-layer}  \\
		\hline 
		GoogLeNet-V1 & Fig. \ref{fig:google-structure}   &  Fig. \ref{fig:google-block}    & Fig. \ref{fig:google-layer}  \\
		\hline 
		ResNet-20 & Fig. \ref{fig:res-structure}   &  Fig. \ref{fig:res-block}    & Fig. \ref{fig:res-layer}  \\
		\hline 
		ViT & Fig. \ref{fig:vit-structure}   &  Fig. \ref{fig:vit-block}    & Fig. \ref{fig:vit-layer}  \\
		\hline 
	\end{tabular}}
\end{table}%



\begin{figure}[htbp]

	\centering

	\subfigure[$ {\rm LS}_0$]{
		\includegraphics[width=0.4\textwidth]{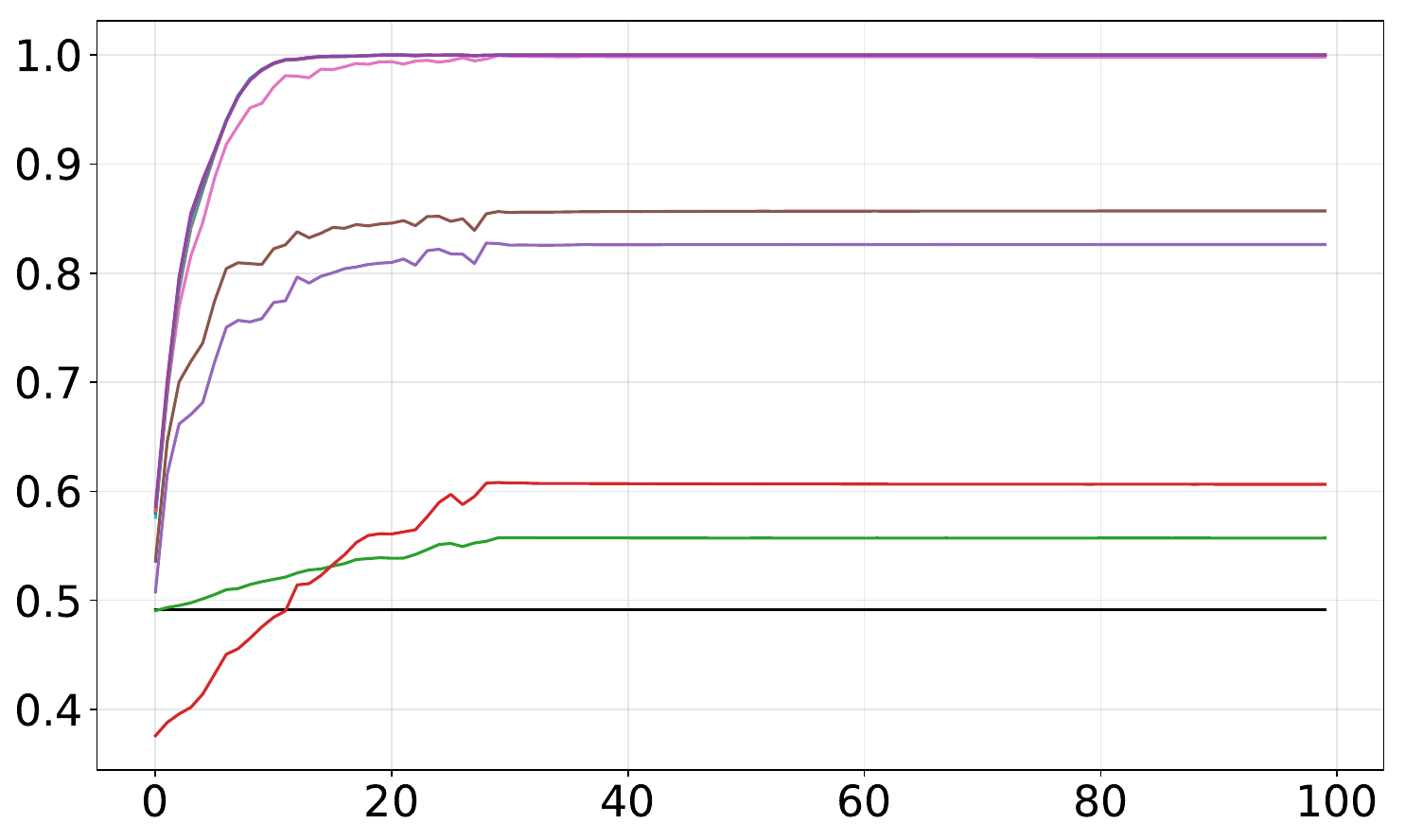}
	}
	\subfigure[$ {\rm LS}_1$]{
		\includegraphics[width=0.4\textwidth]{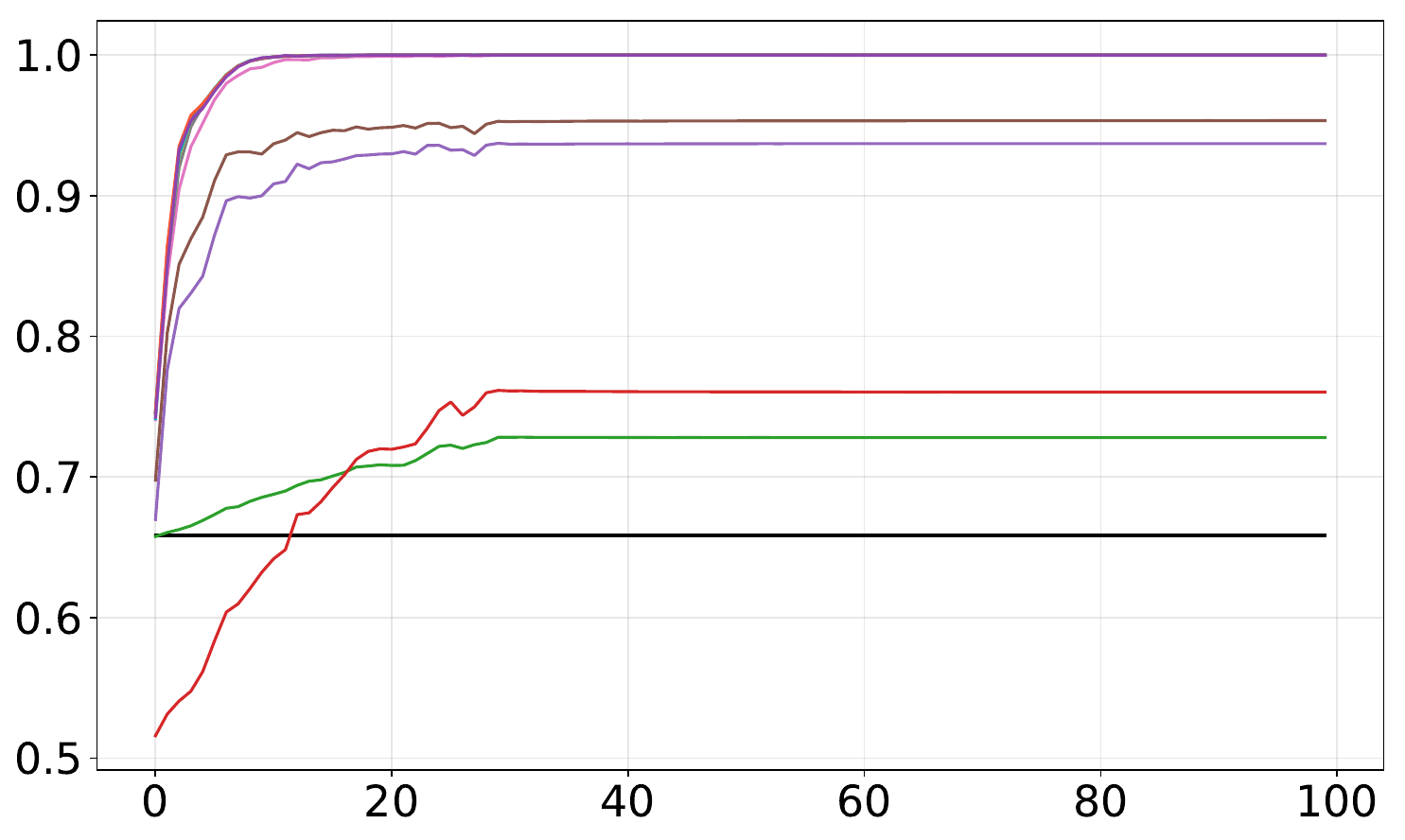}
	}
	\subfigure[$ {\rm LS}_2$]{
		\includegraphics[width=0.4\textwidth]{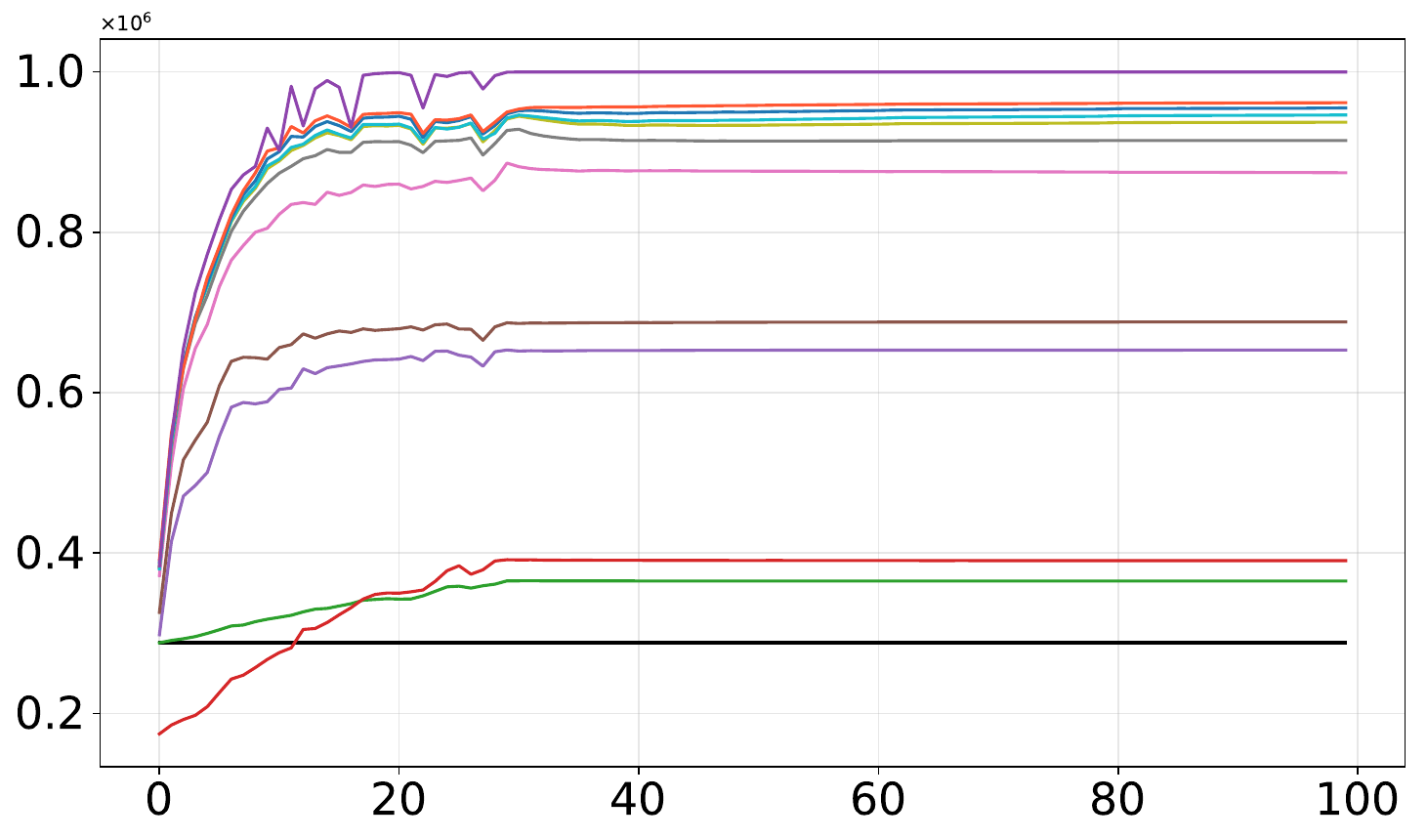}
	}
	\subfigure[Accuracy]{
		\includegraphics[width=0.4\textwidth]{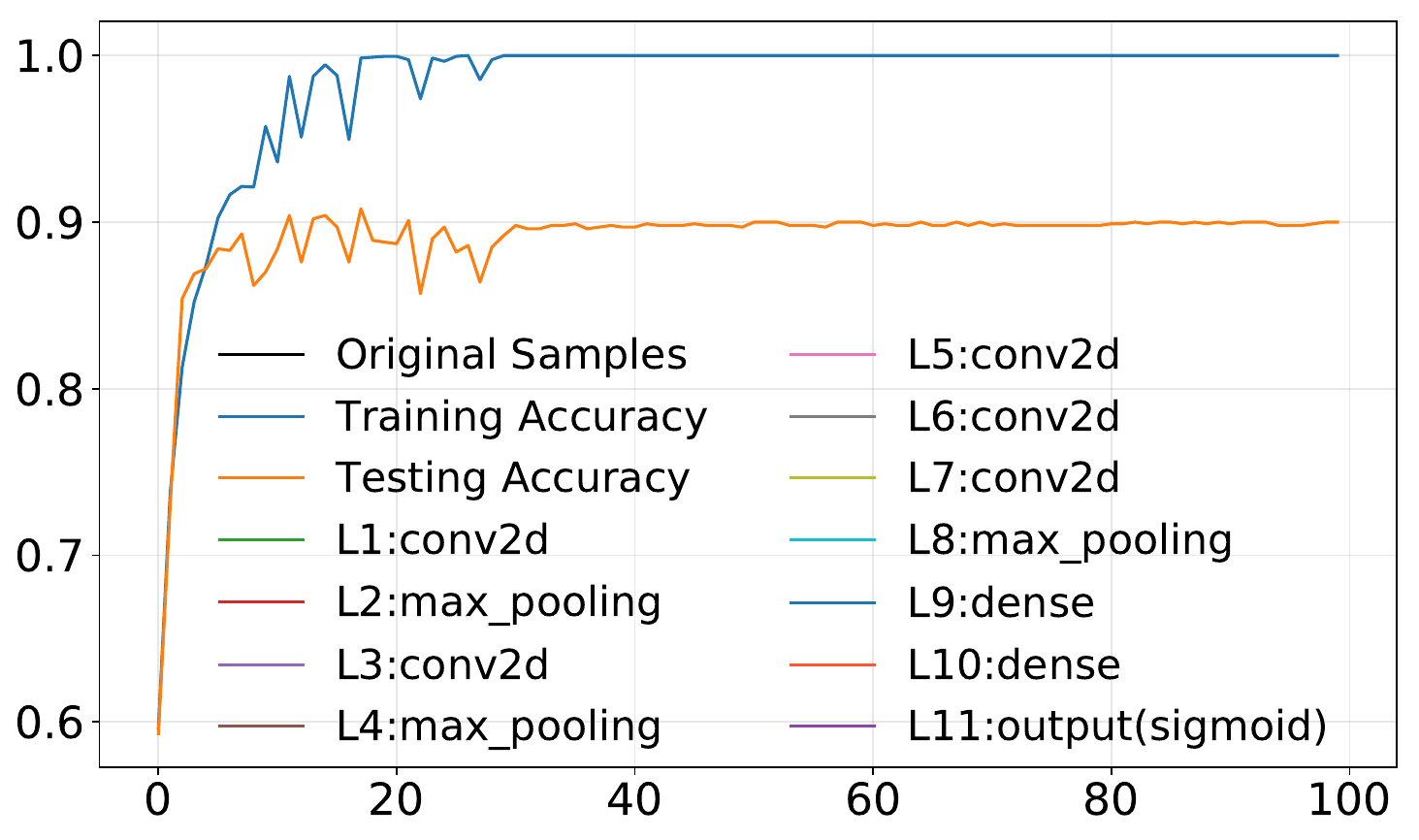}
	}
	 
	\caption{MD-LSM and Accuracy Curves of Hidden Layers of AlexNet (Sigmoid)}
	\label{fig:AlexNet-sigmoid}				
	
	
	\end{figure}	


\begin{figure}[htbp]

	\centering	

	\subfigure[$ {\rm LS}_0$]{
		\includegraphics[width=0.4\textwidth]{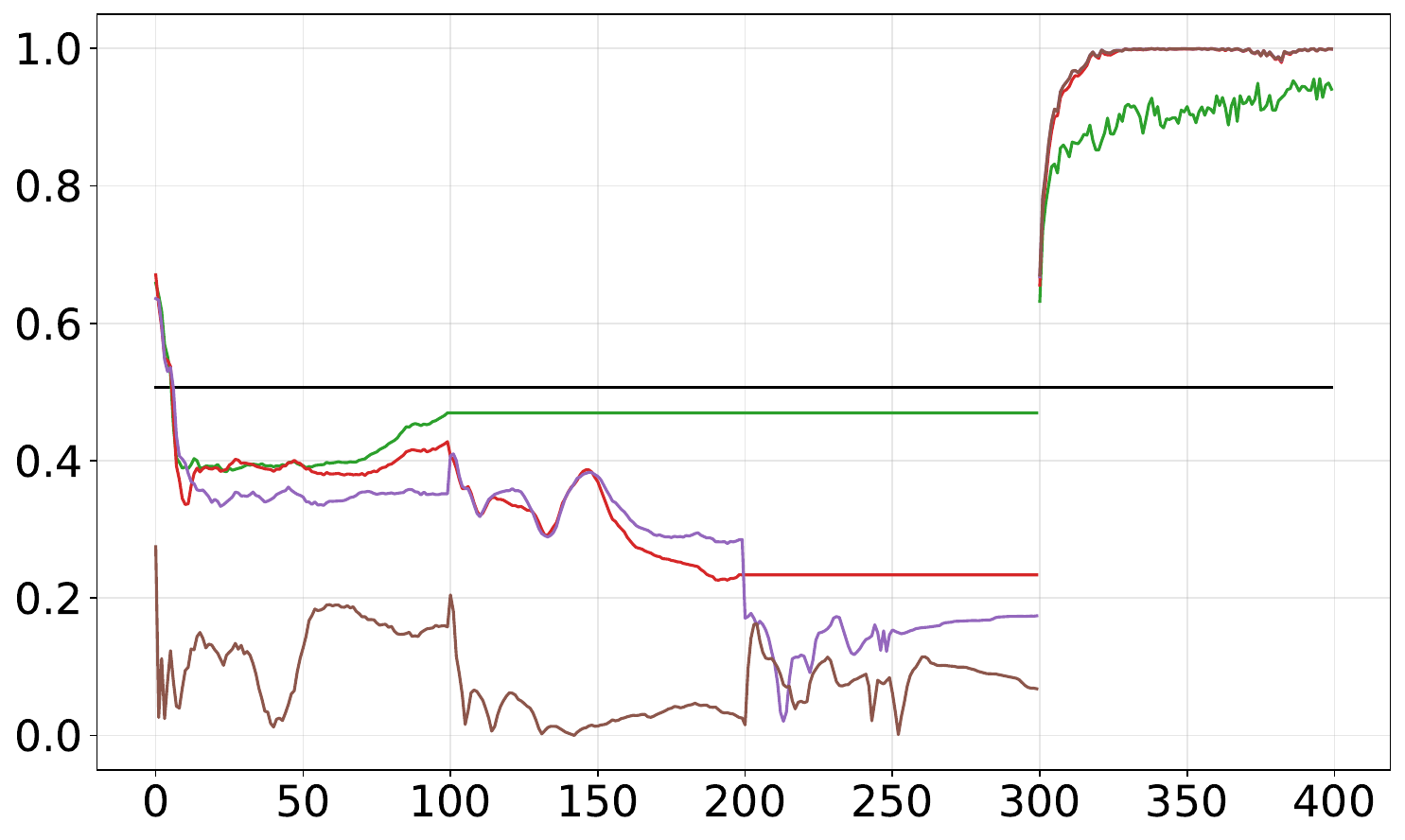}
	}
	\subfigure[$ {\rm LS}_1$]{
		\includegraphics[width=0.4\textwidth]{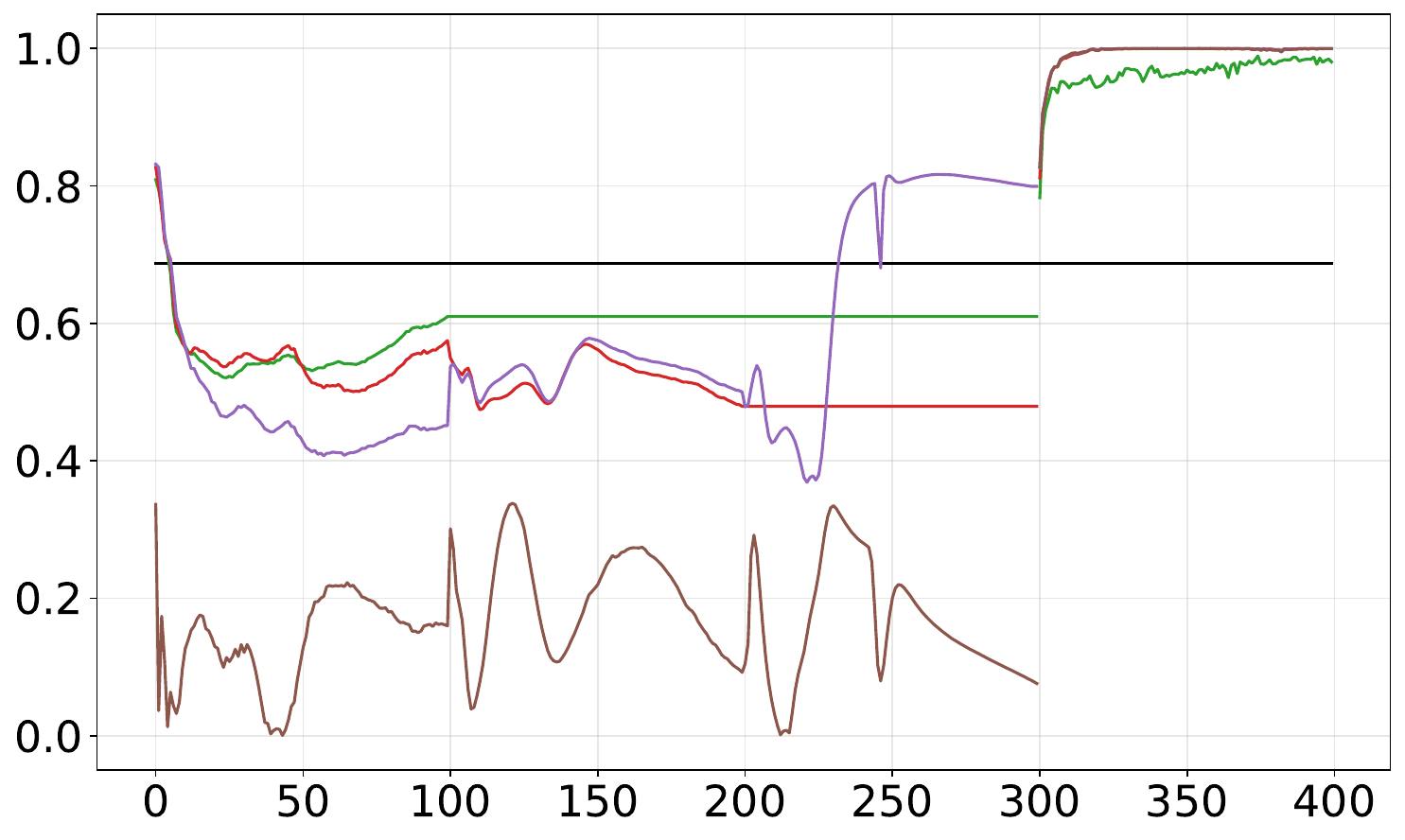}
	}
	\subfigure[$ {\rm LS}_2$]{
		\includegraphics[width=0.4\textwidth]{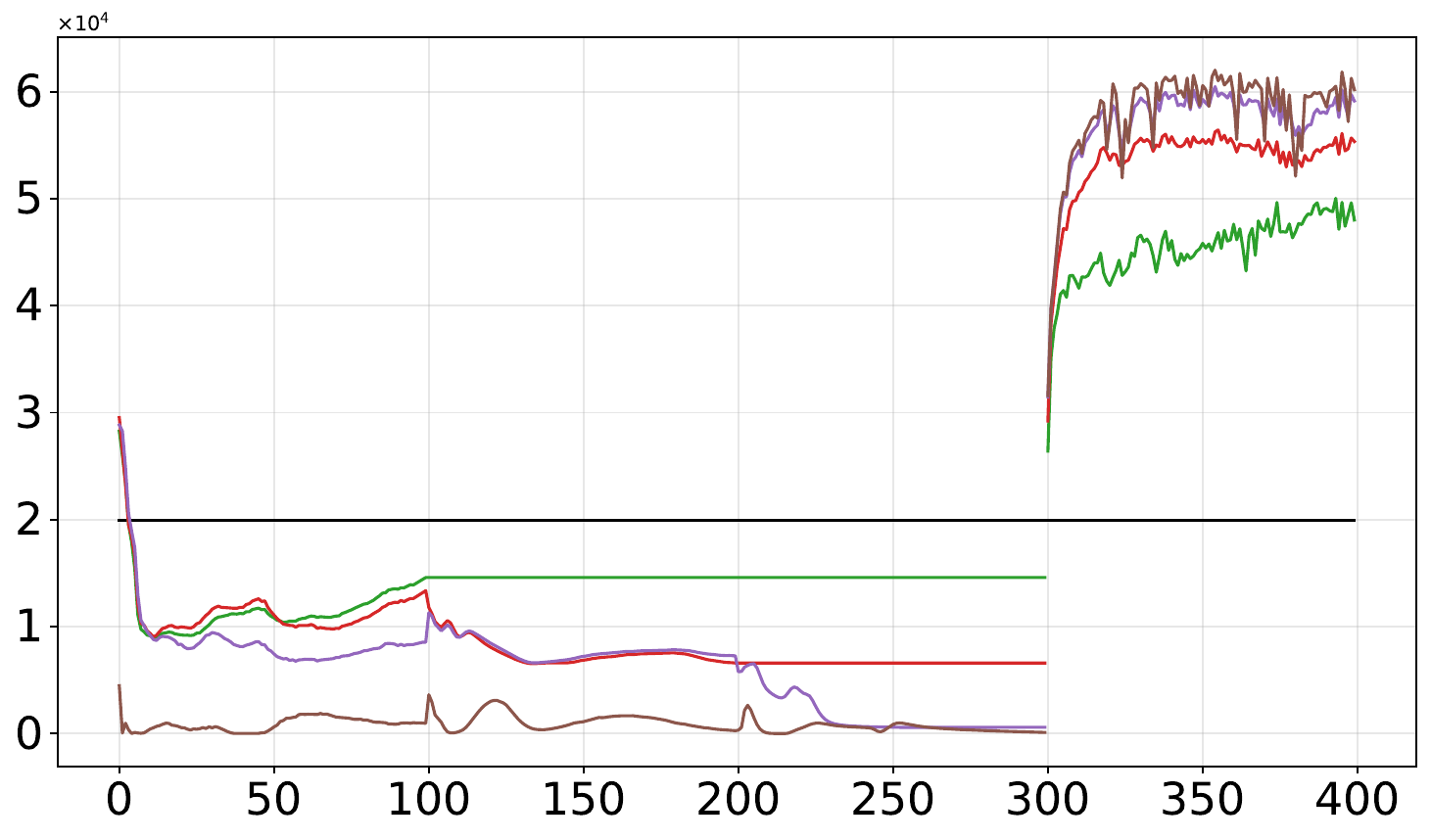}
	}
	\subfigure[Accuracy]{
		\includegraphics[width=0.4\textwidth]{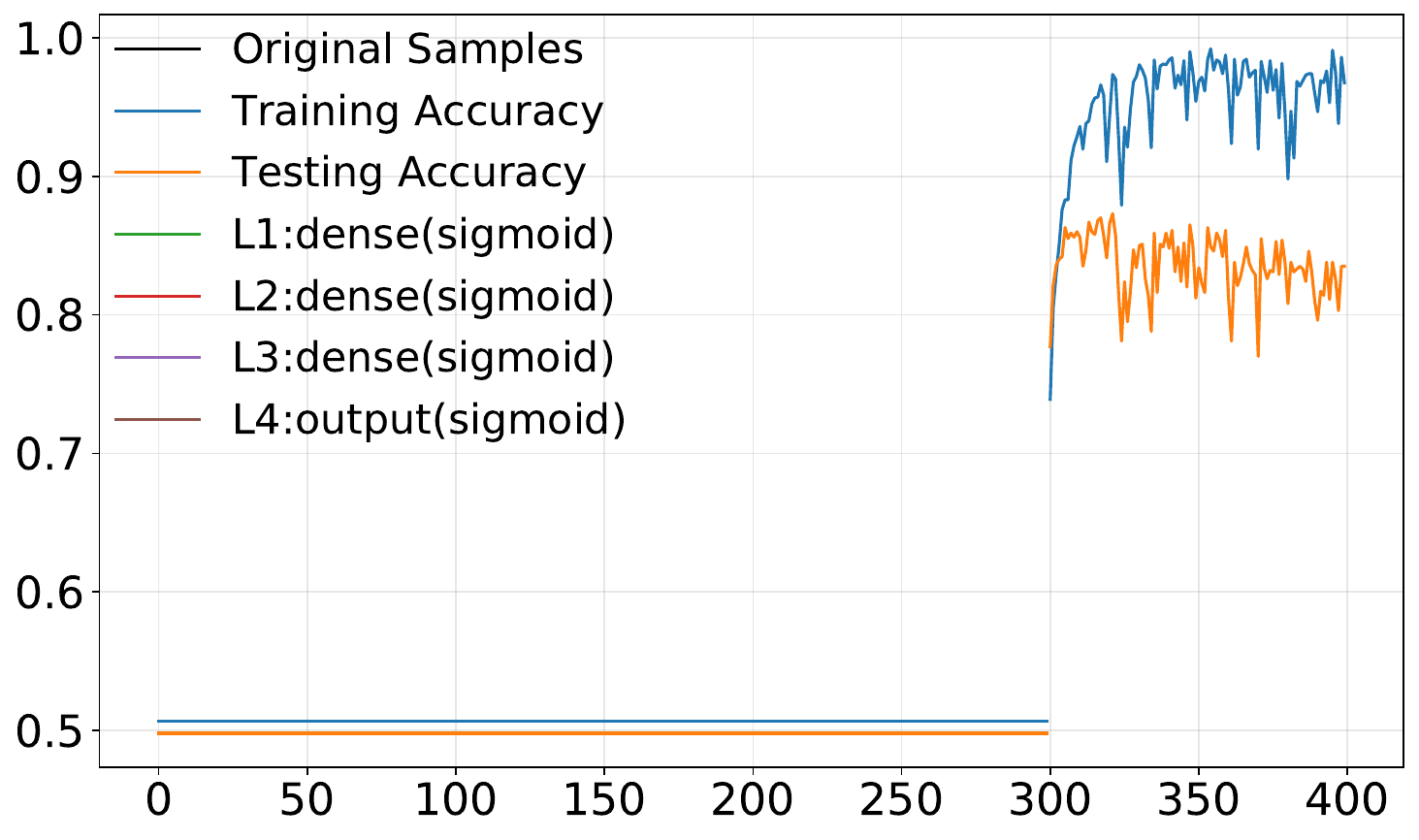}
	}
	 
	\caption{MD-LSM and Accuracy Curves of DBN's Hidden Layers}
	\label{fig:dbn}	
	
\end{figure}	

\begin{figure}[htbp]
	\centering	
	\subfigure[$ {\rm LS}_0$]{
		\includegraphics[width=0.4\textwidth]{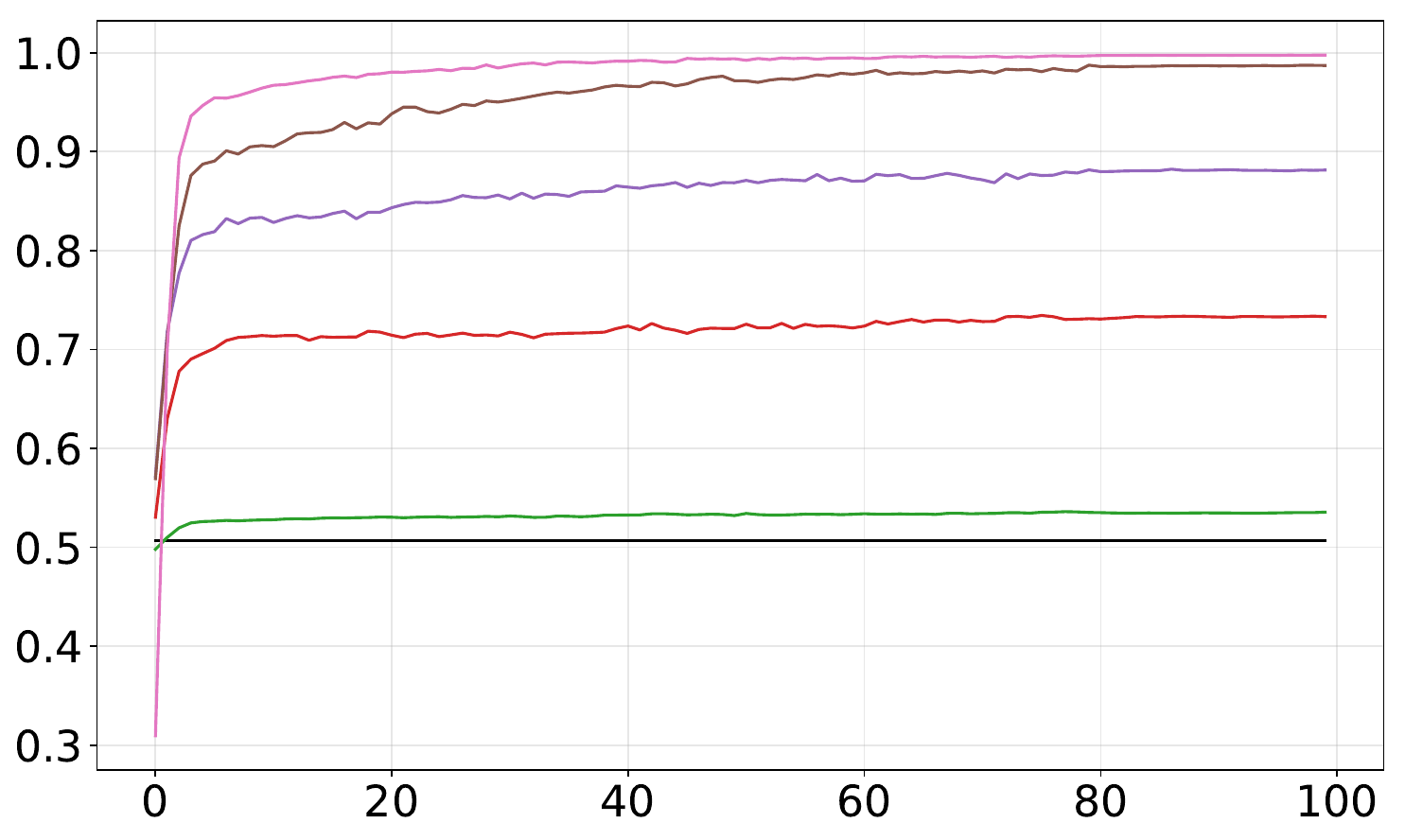}
	}
	\subfigure[$ {\rm LS}_1$]{
		\includegraphics[width=0.4\textwidth]{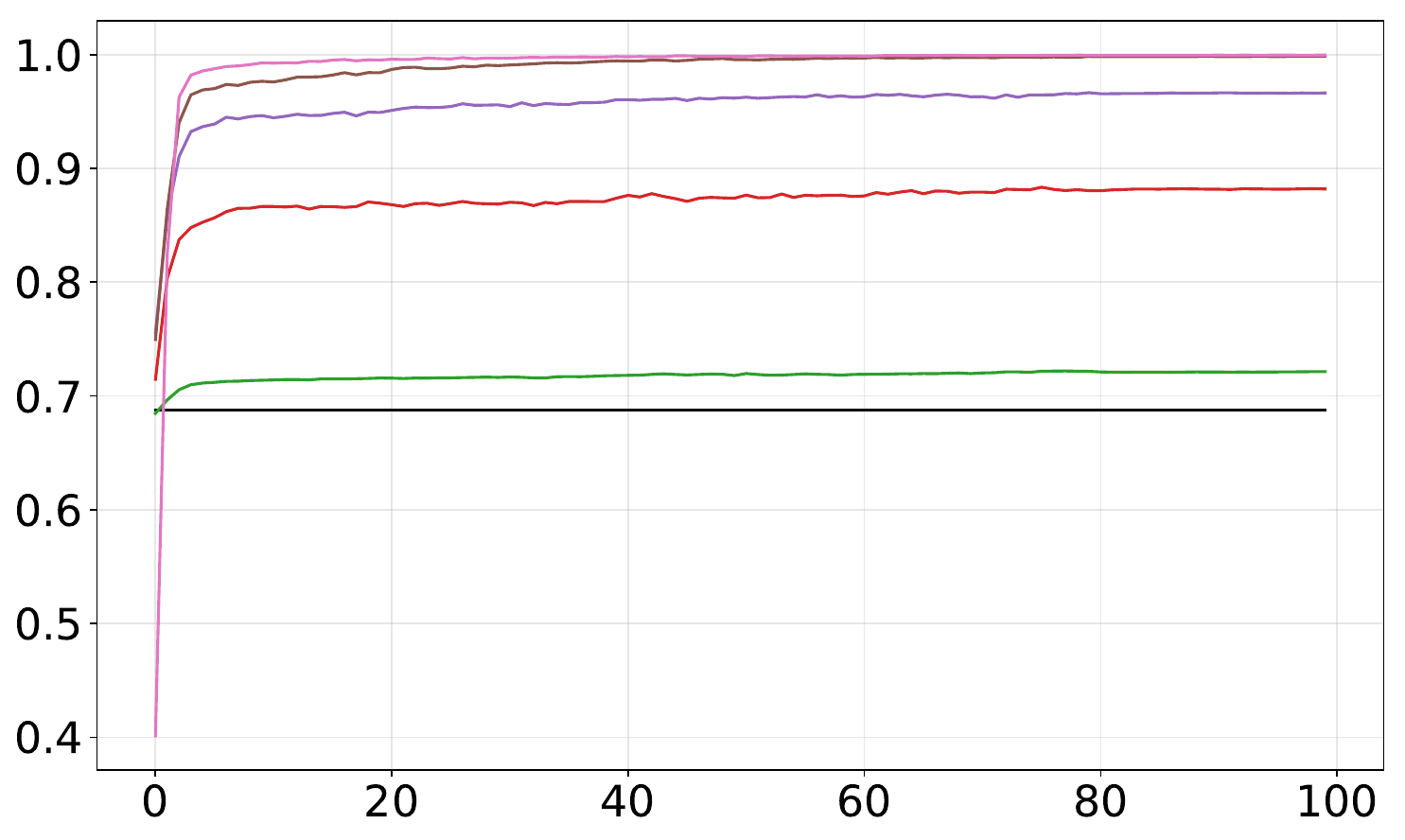}
	}
	\subfigure[$ {\rm LS}_2$]{
		\includegraphics[width=0.4\textwidth]{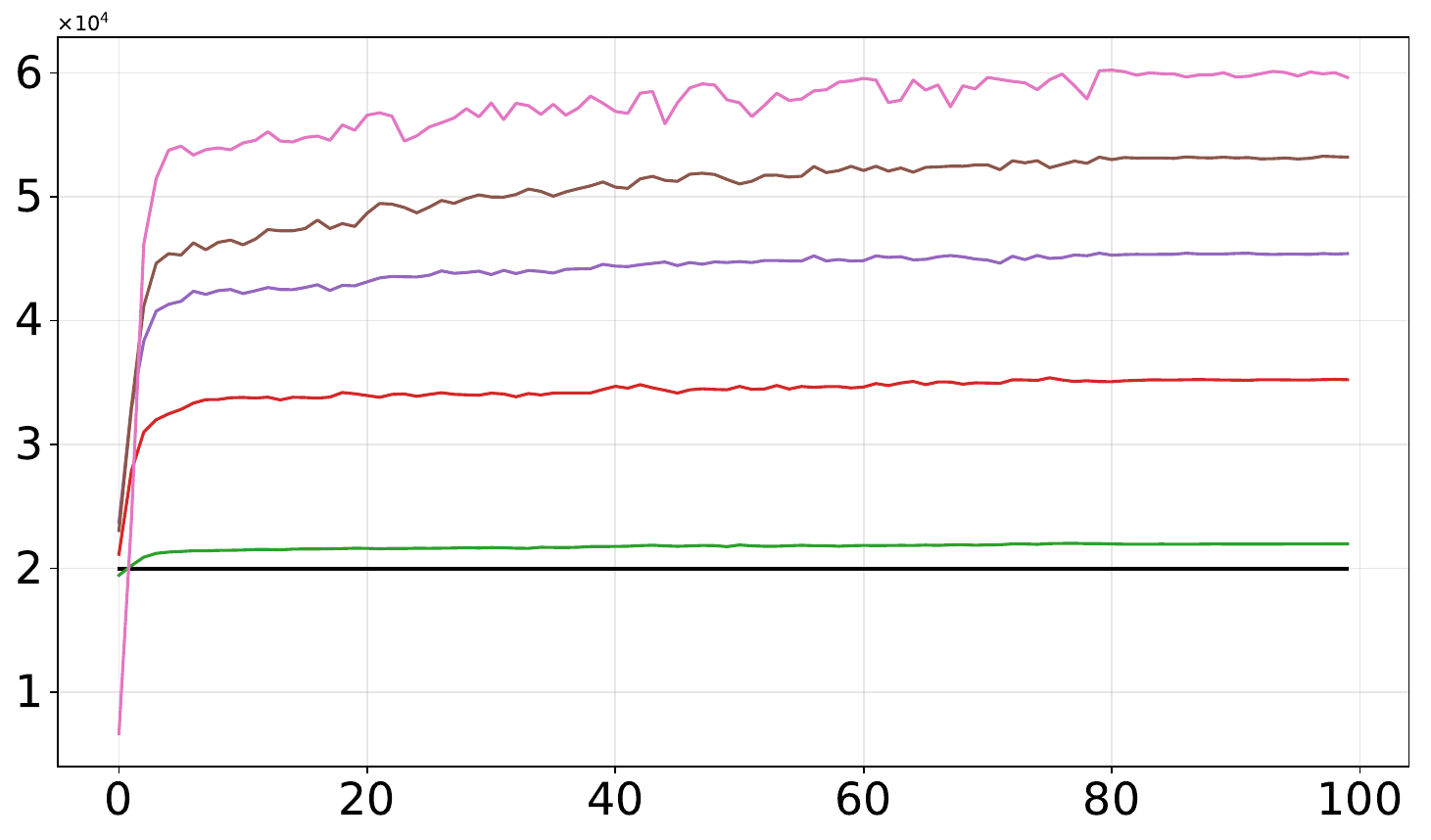}
	}
	\subfigure[Accuracy]{
		\includegraphics[width=0.4\textwidth]{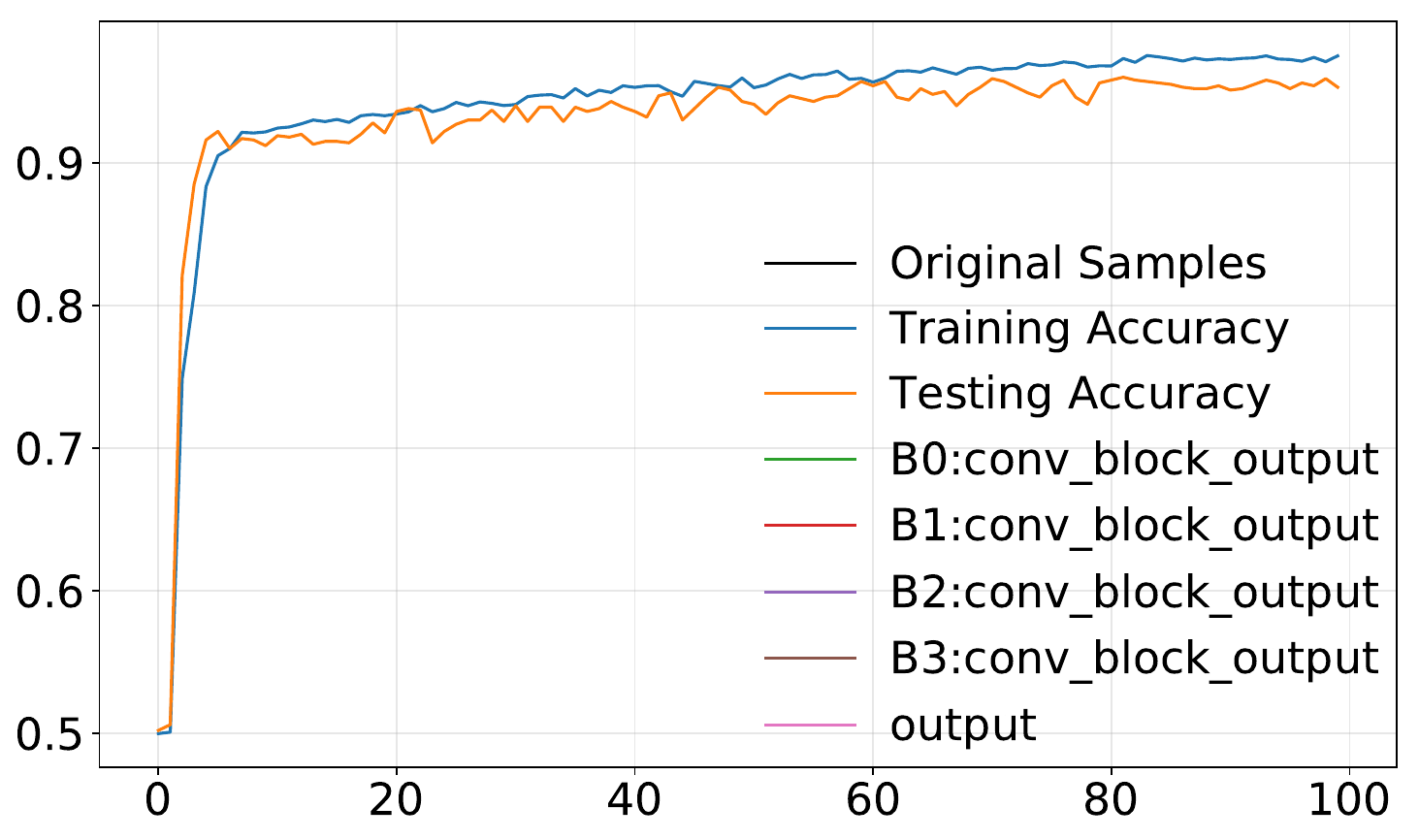}
	}
	\caption{MD-LSM and Accuracy Curves of VGGNet's Main Blocks}
	\label{fig:vgg-block}	
\end{figure}


\begin{figure}[htbp]
	\centering
	\includegraphics[width=0.6\textwidth]{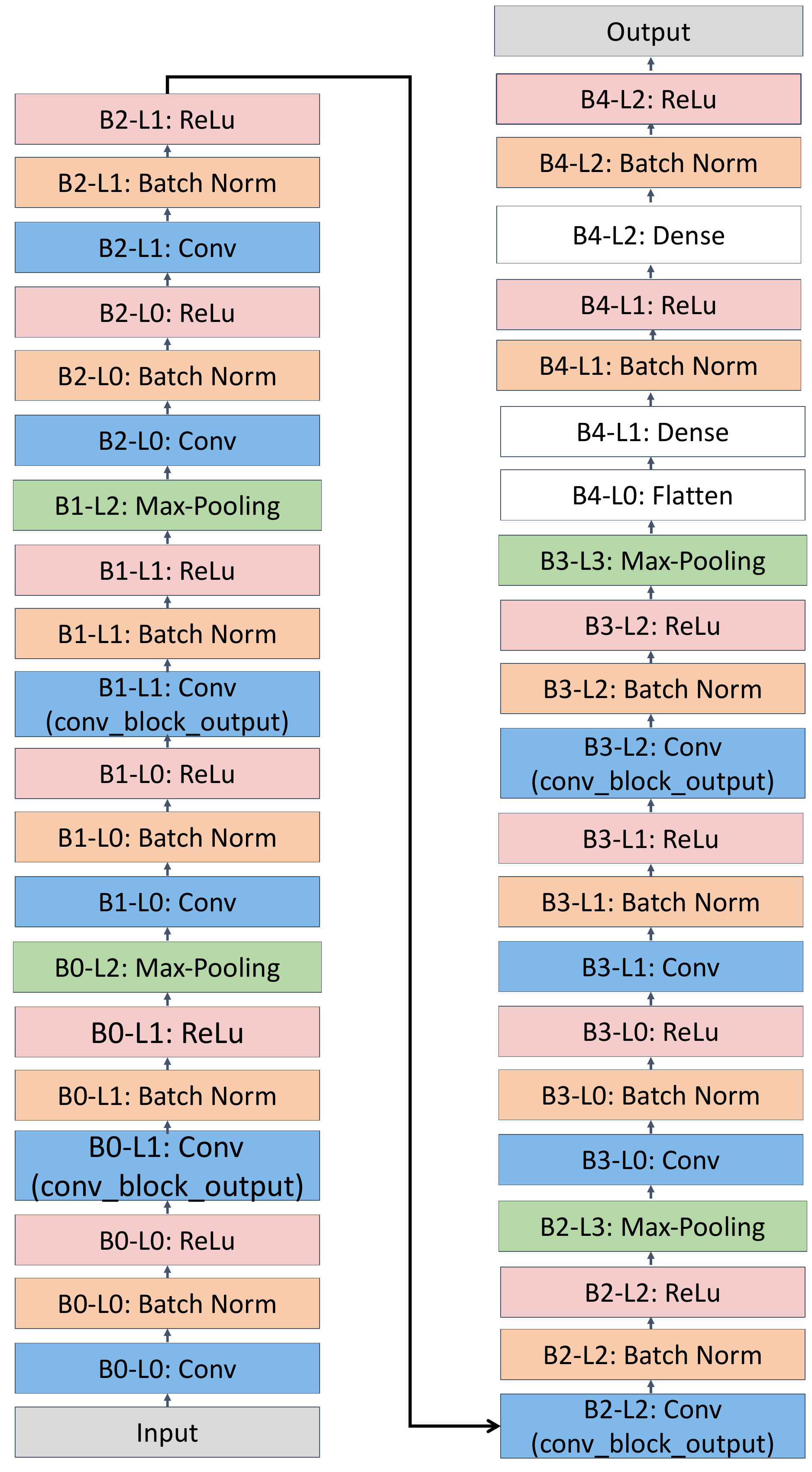}
	\caption{Structure of VGGNet}
	\label{fig:vgg-structure}
\end{figure}

\begin{figure}[htbp]
	
	\centering	
	\subfigure[$ {\rm LS}_0$]{
		\includegraphics[width=0.98\textwidth]{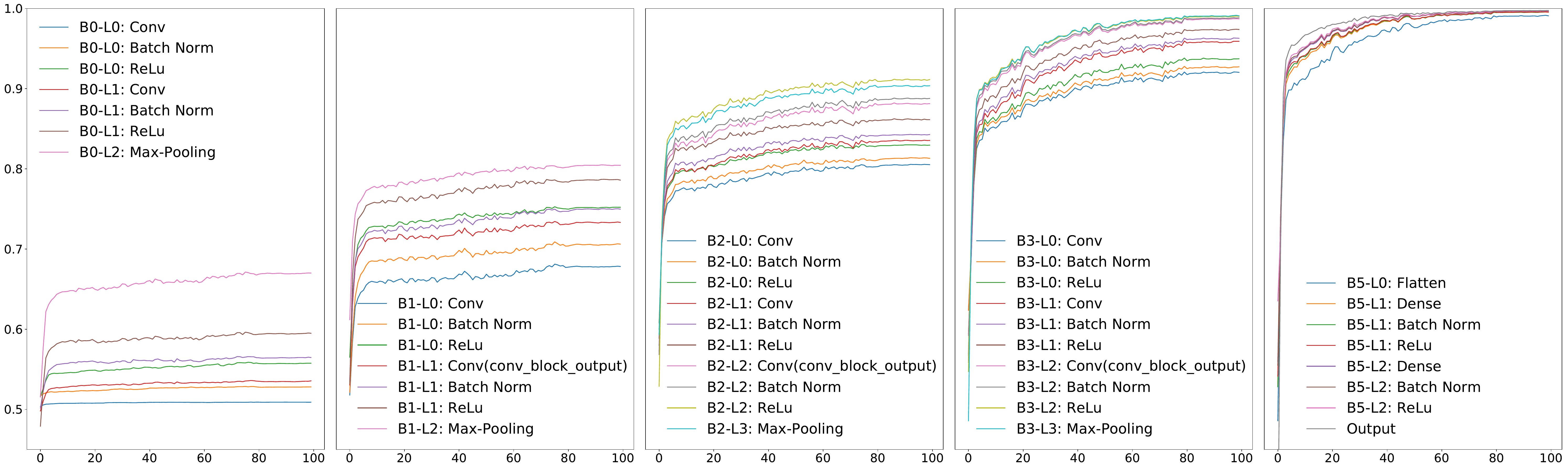}
	}
	\subfigure[$ {\rm LS}_1$]{
		\includegraphics[width=0.98\textwidth]{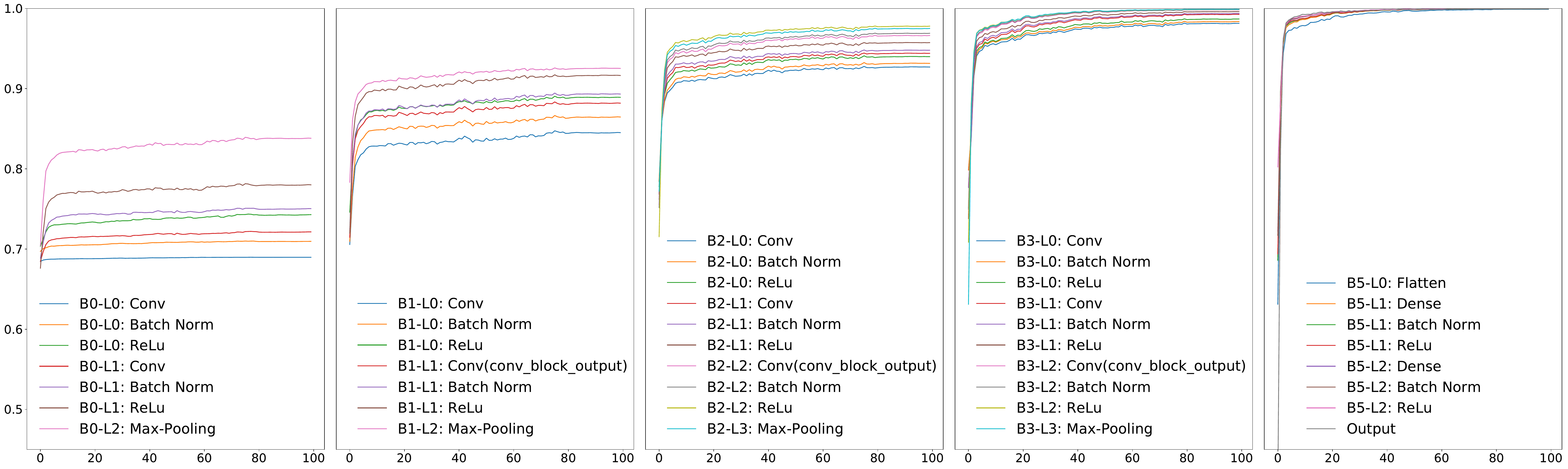}
	}
	\subfigure[$ {\rm LS}_2$]{
		\includegraphics[width=0.98\textwidth]{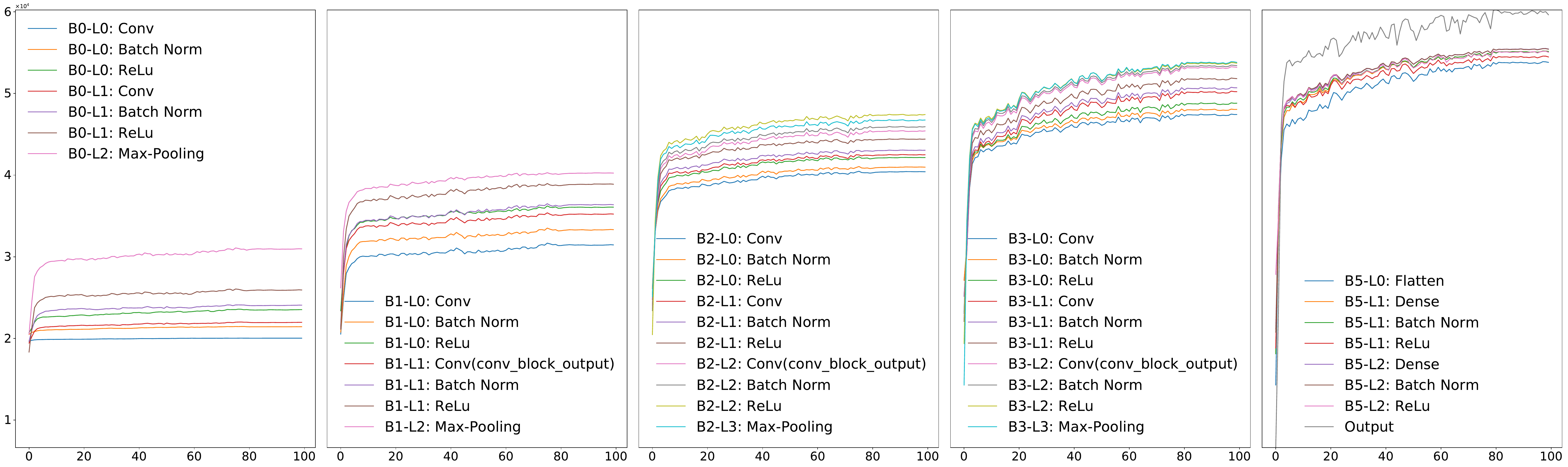}
	}
	\caption{MD-LSM and Accuracy Curves of VGGNet's Hidden Layers}
	\label{fig:vgg-layer}
\end{figure}	


\begin{figure}[htbp]
	\centering
	\includegraphics[width=0.45\textwidth]{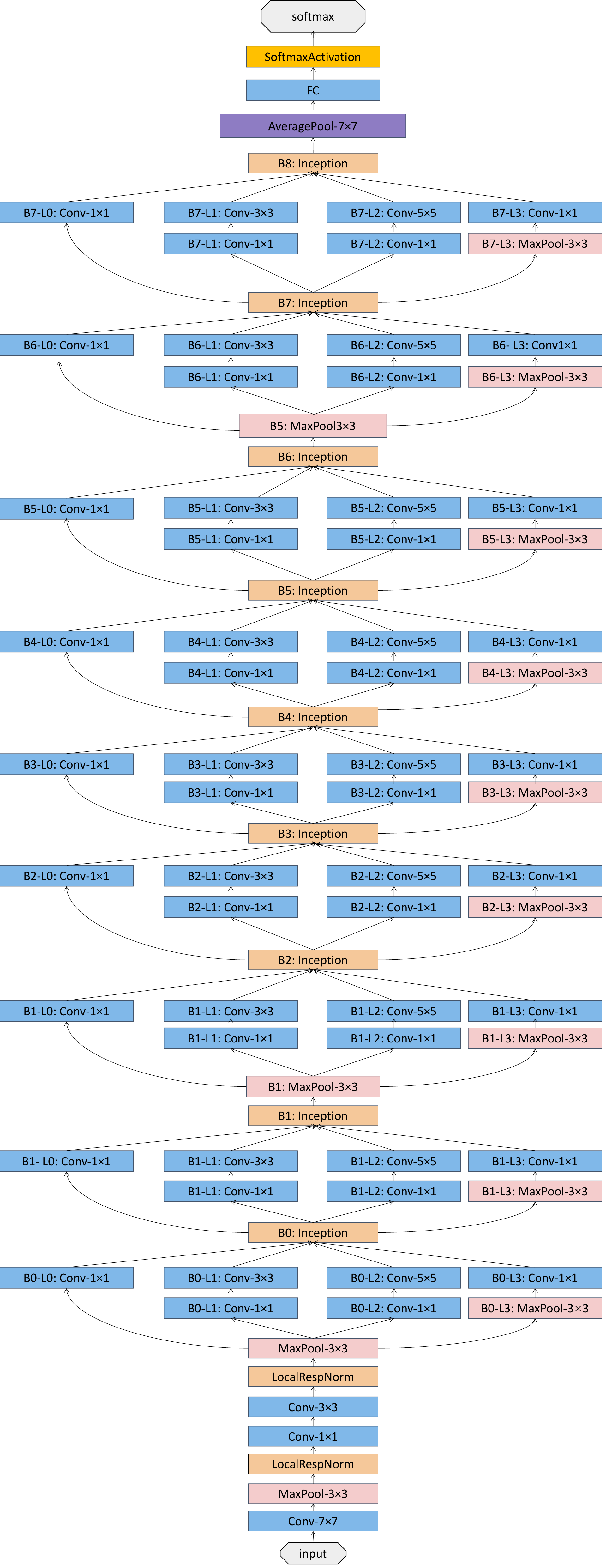}
	\caption{Structure of GoogleNet-V1}
	\label{fig:google-structure}
\end{figure}

\begin{figure}[htbp]
	\centering
	\subfigure[$ {\rm LS}_0$]{
		\includegraphics[width=0.4\textwidth]{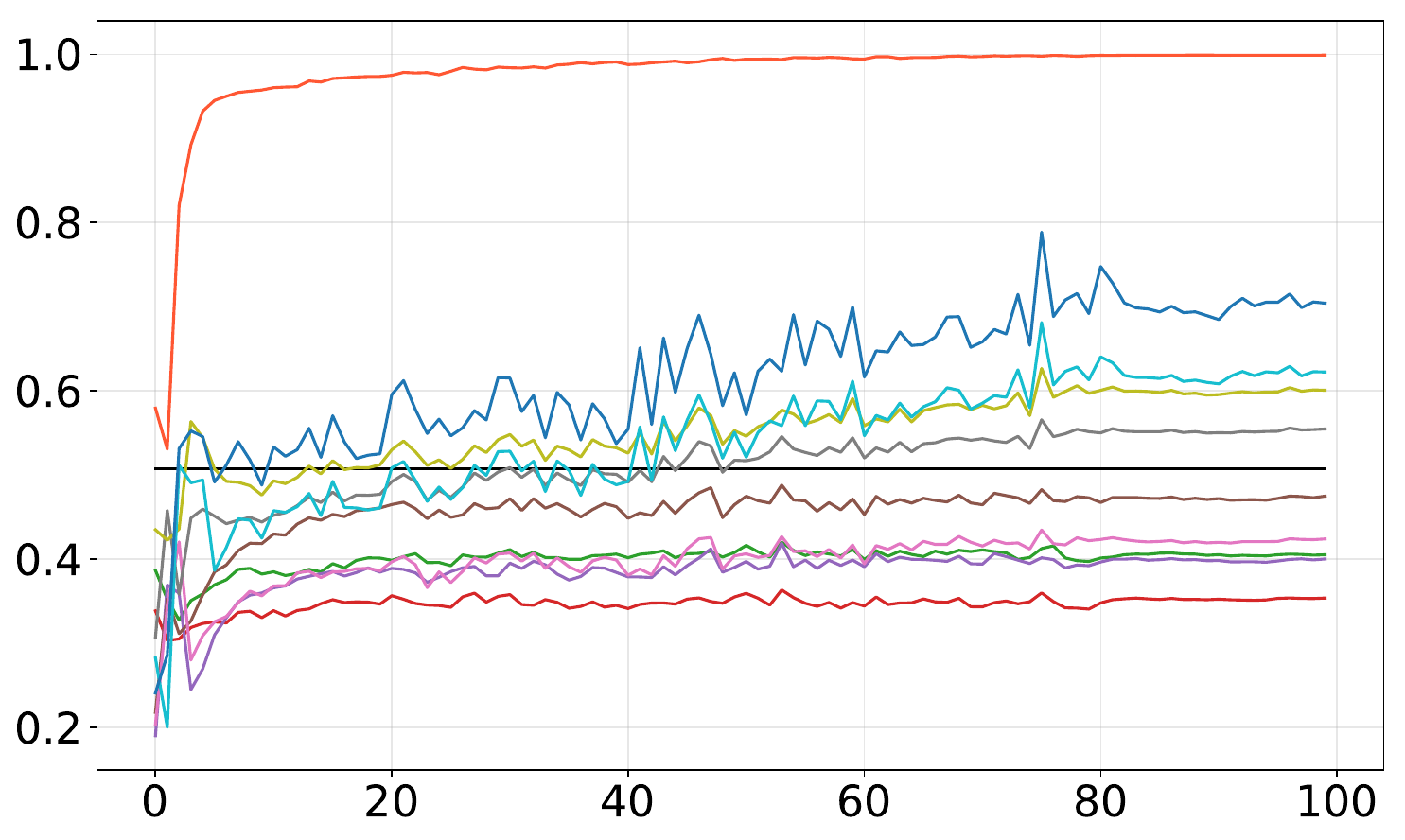}
	}
	\subfigure[$ {\rm LS}_1$]{
		\includegraphics[width=0.4\textwidth]{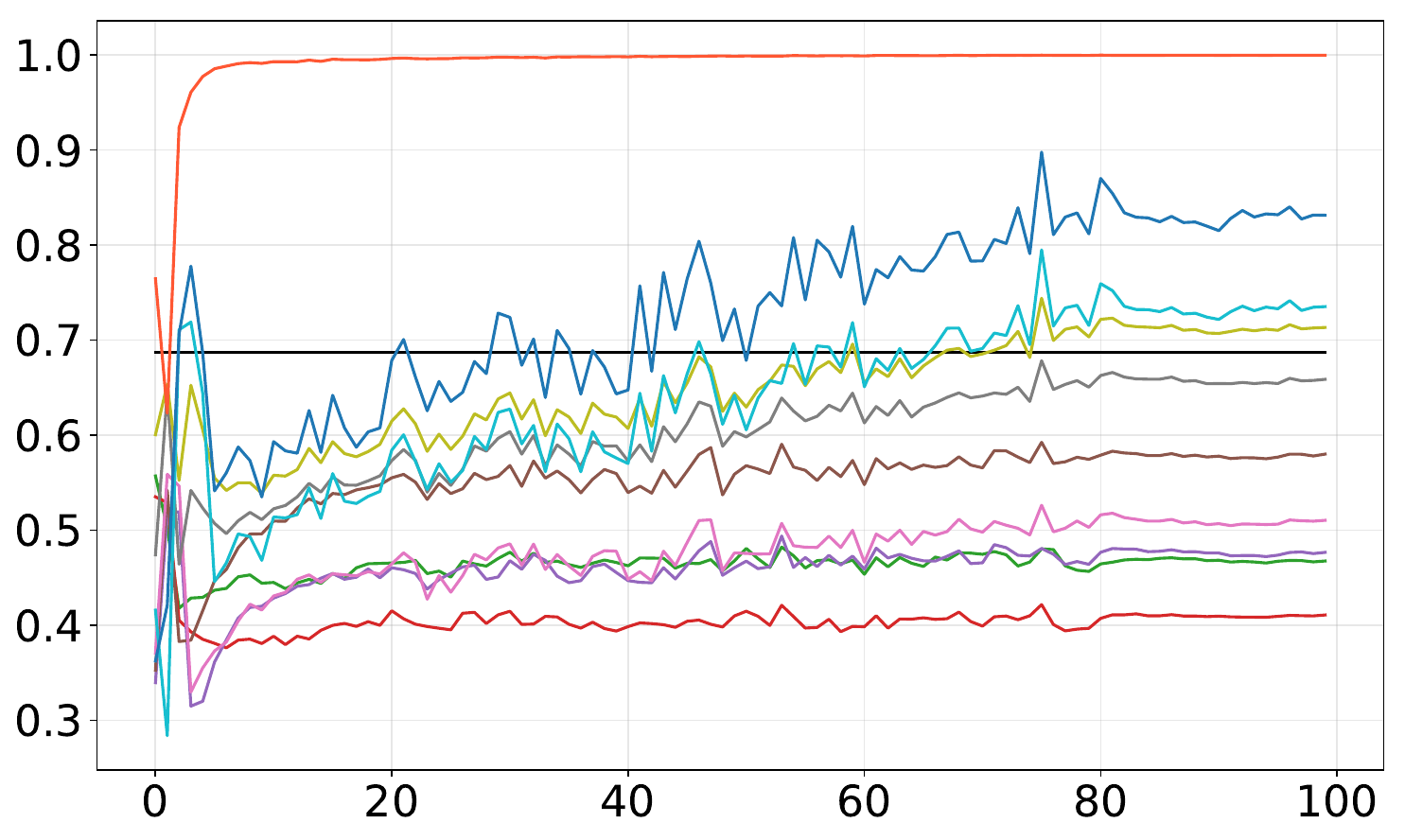}
	}
	\subfigure[$ {\rm LS}_2$]{
		\includegraphics[width=0.4\textwidth]{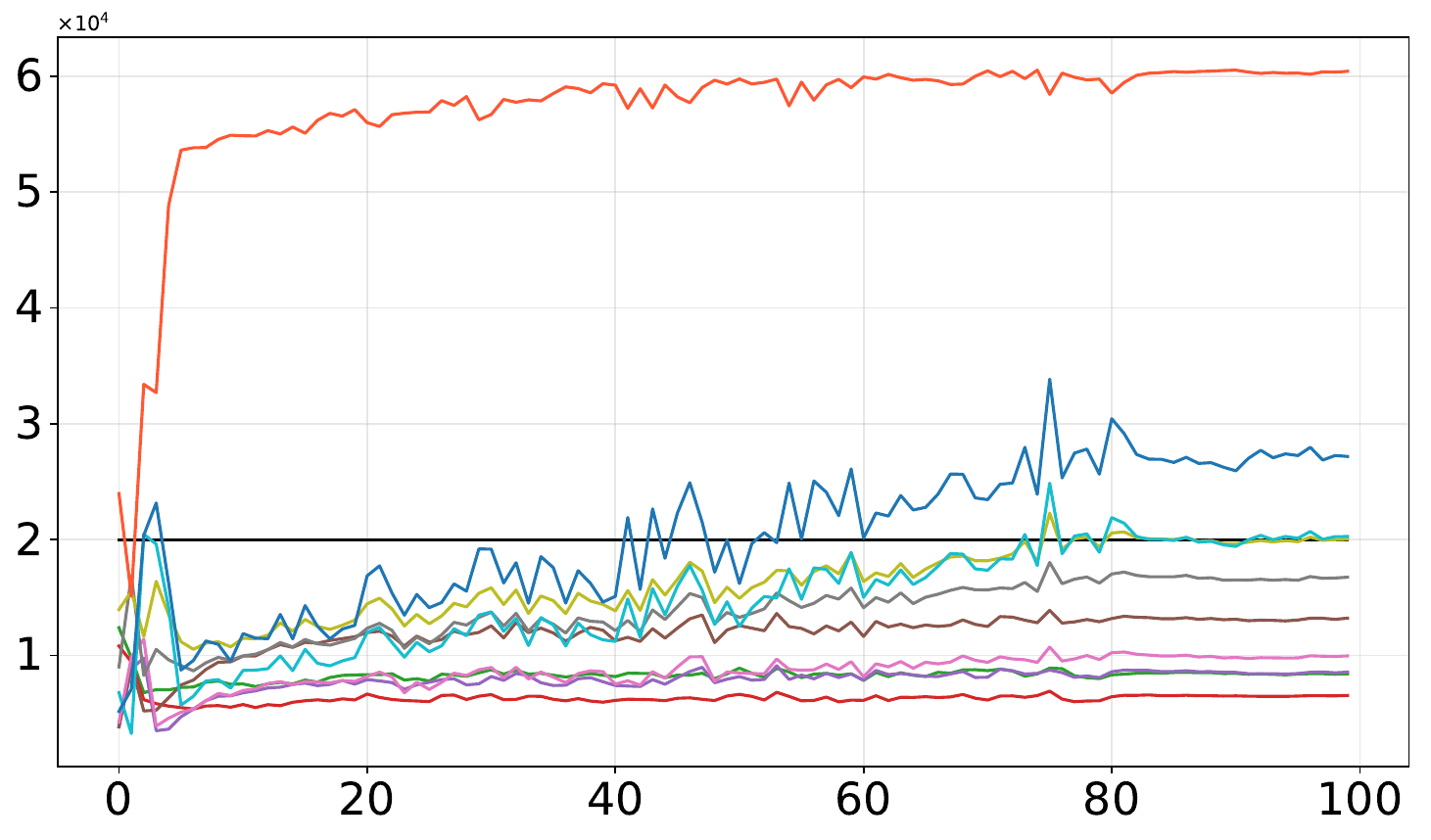}
	}
	\subfigure[Accuracy]{
		\includegraphics[width=0.4\textwidth]{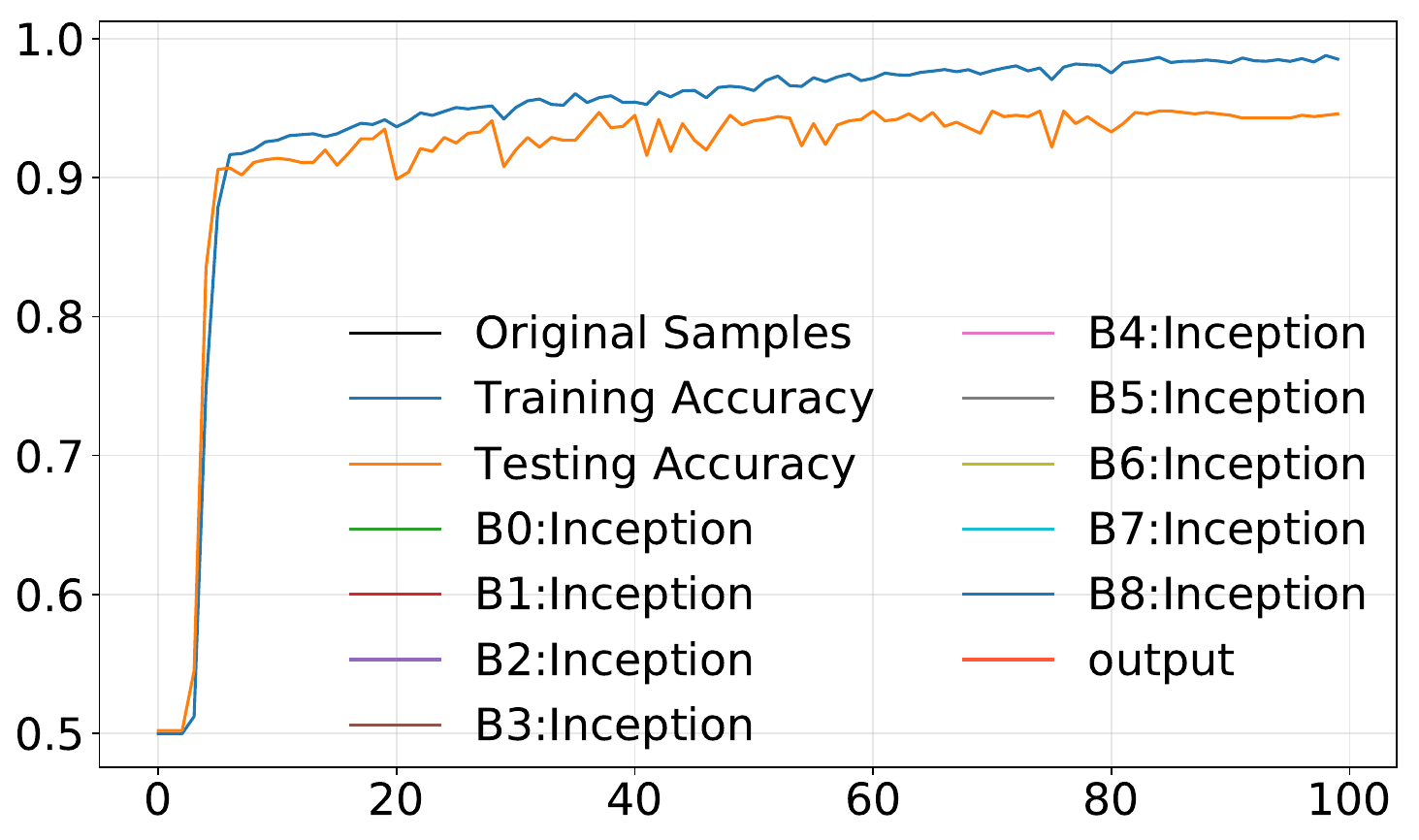}
	}
	 
	\caption{MD-LSM and Accuracy Curves of GoogleNet's Main Blocks }
	\label{fig:google-block}	
	
\end{figure}

\begin{figure}[htbp]
	\centering

	\subfigure[$ {\rm LS}_0$]{
		\includegraphics[width=0.98\textwidth]{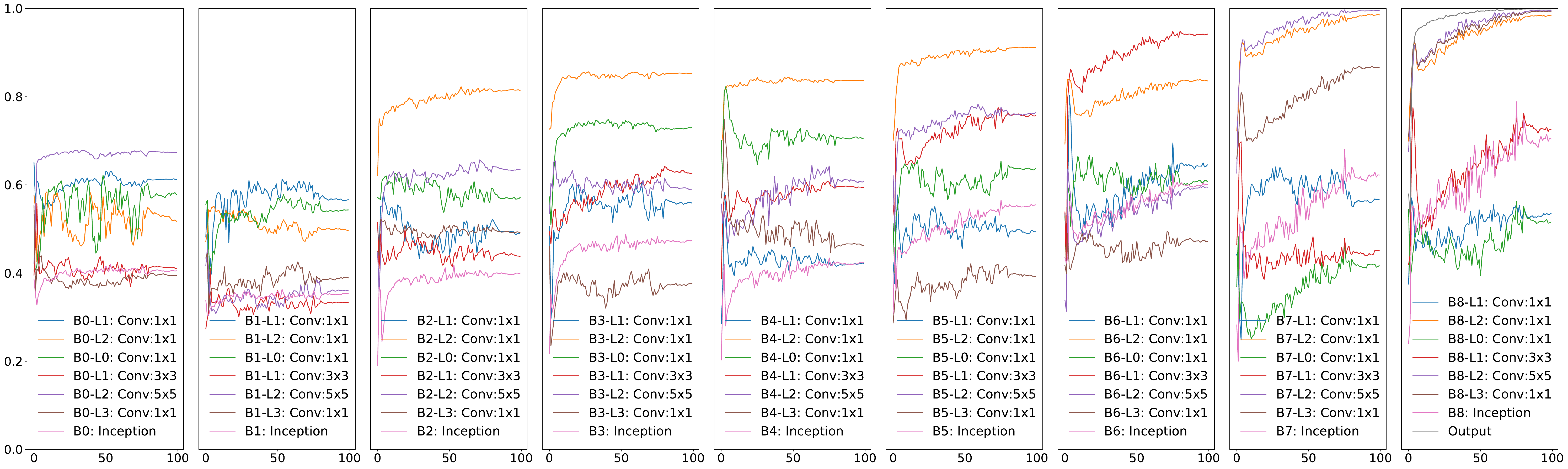}
	}
	\subfigure[$ {\rm LS}_1$]{
		\includegraphics[width=0.98\textwidth]{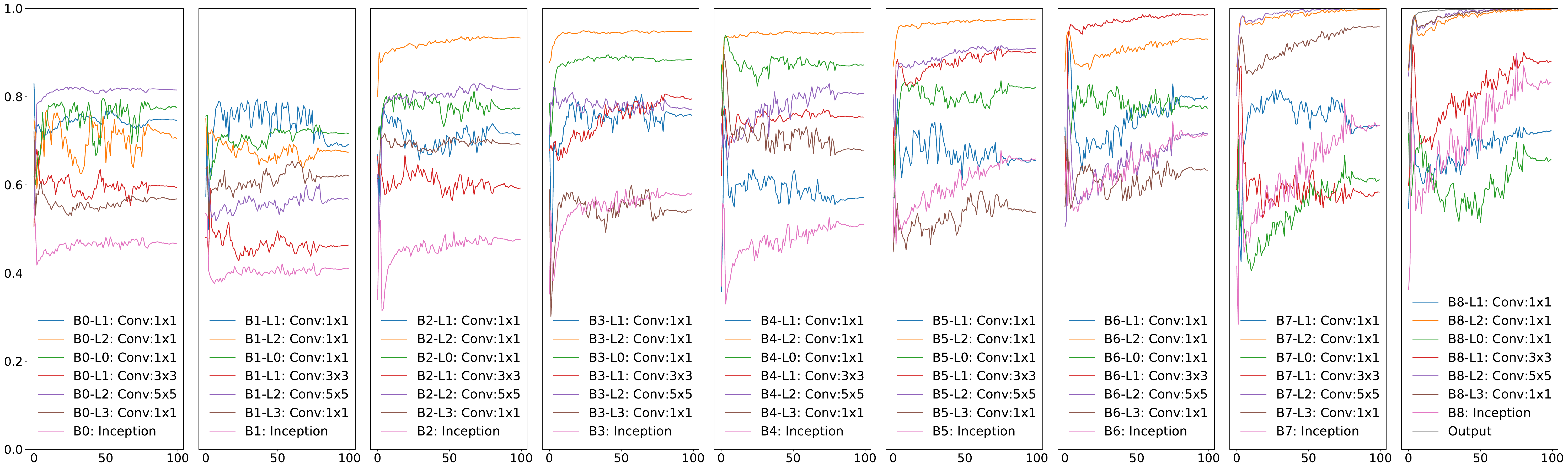}
	}
	\subfigure[$ {\rm LS}_2$]{
		\includegraphics[width=0.98\textwidth]{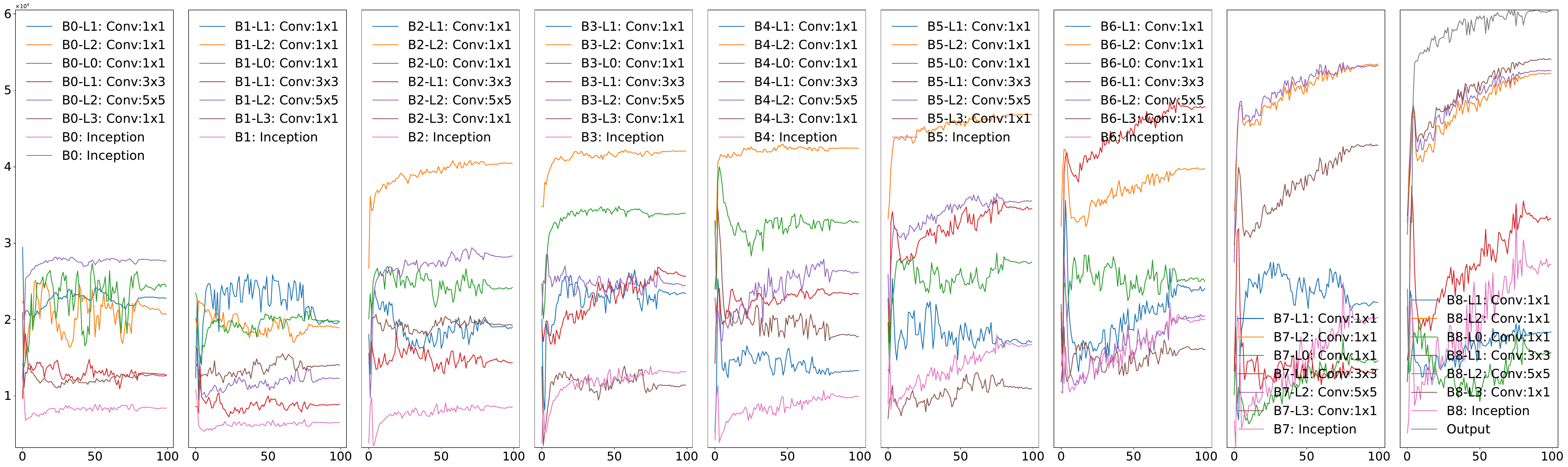}
	}
	\caption{MD-LSM and Accuracy Curves of GoogLeNet's Hidden Layers}\label{fig:google-layer}
\end{figure}

\begin{figure}[htbp]
	\centering
	\includegraphics[width=0.55\textwidth]{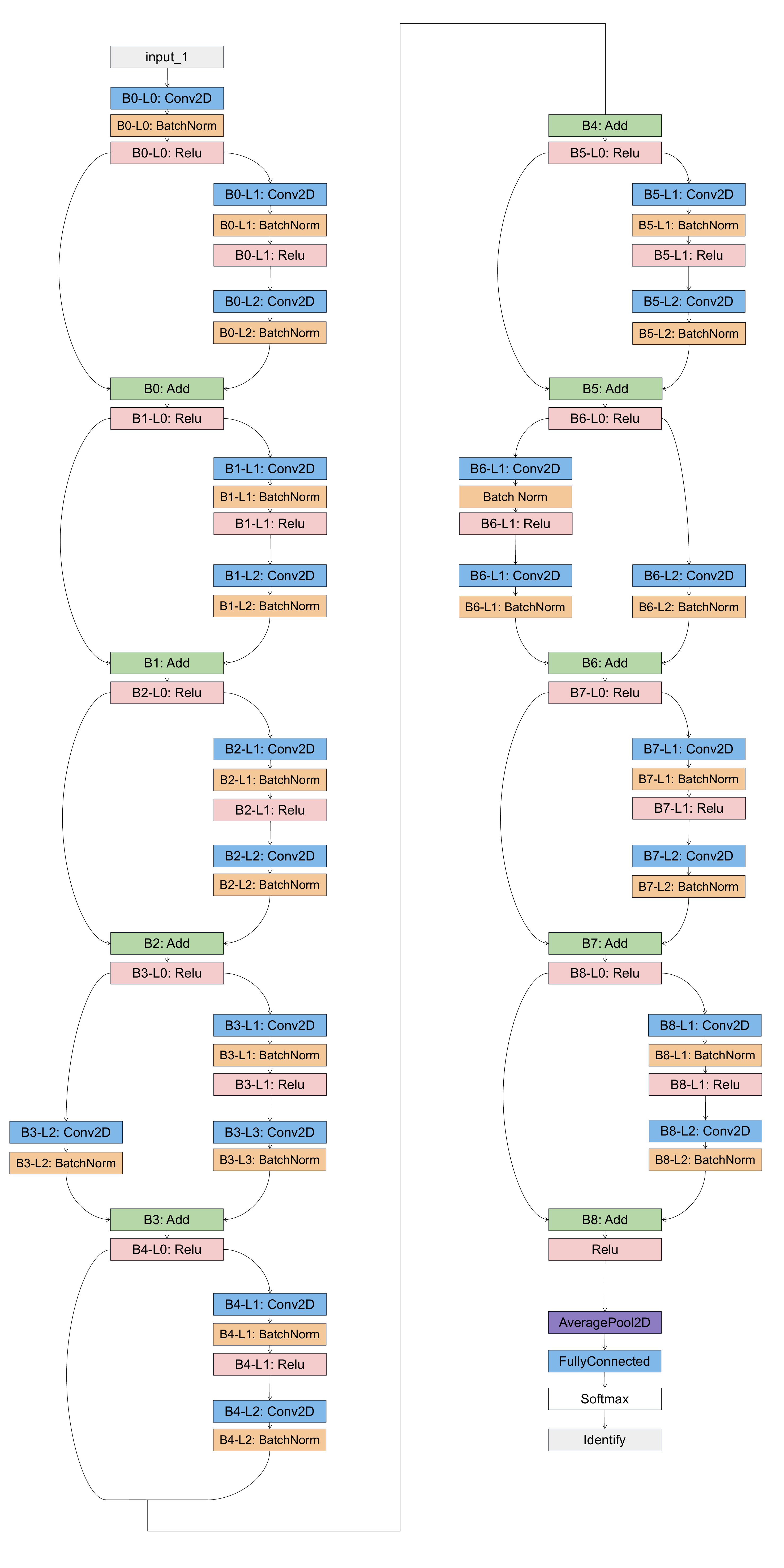}
	\caption{Structure of ResNet-20}
	\label{fig:res-structure}
\end{figure}

\begin{figure}[htbp]
	\centering	
	\subfigure[$ {\rm LS}_0$]{
		\includegraphics[width=0.4\textwidth]{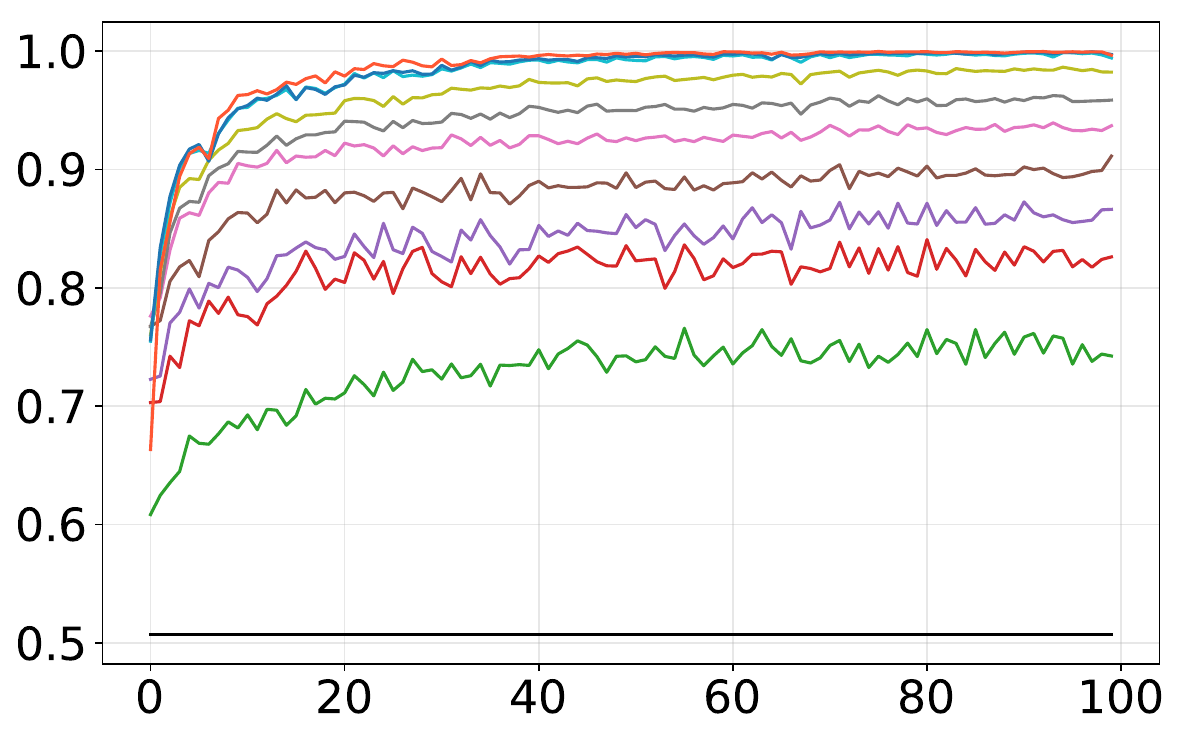}
	}
	\subfigure[$ {\rm LS}_1$]{
		\includegraphics[width=0.4\textwidth]{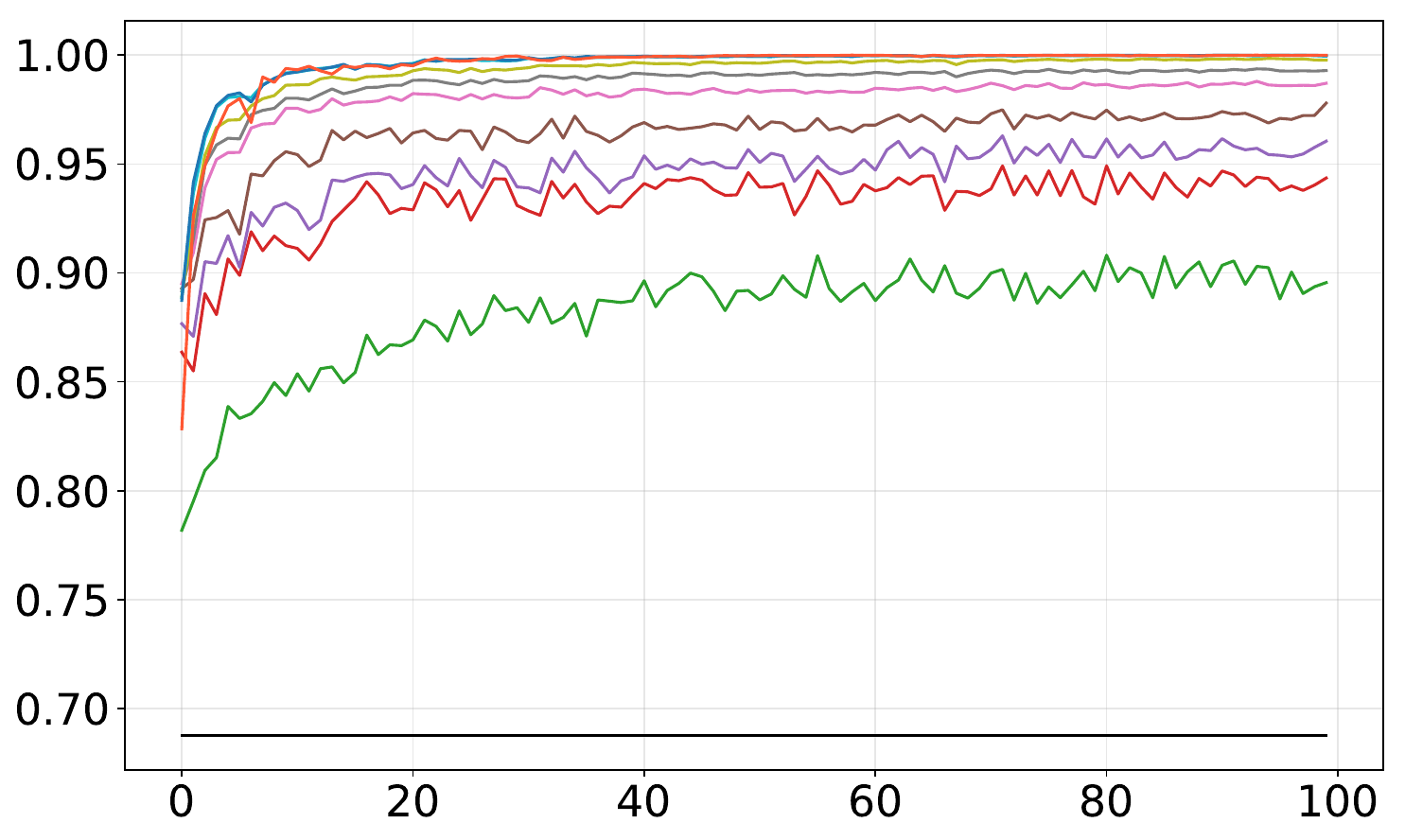}
	}
	\subfigure[$ {\rm LS}_2$]{
		\includegraphics[width=0.4\textwidth]{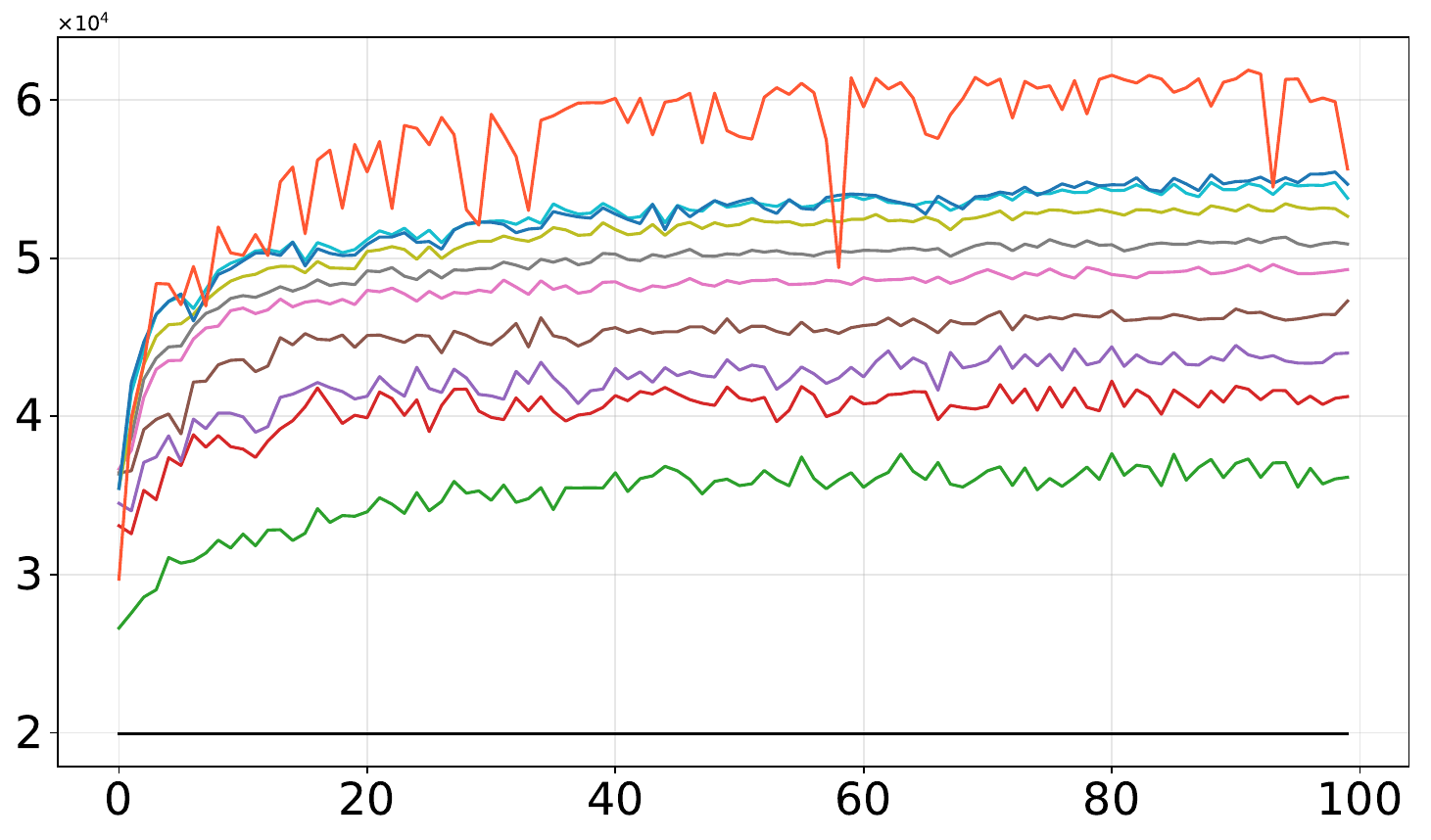}
	}
	\subfigure[Accuracy]{
		\includegraphics[width=0.4\textwidth]{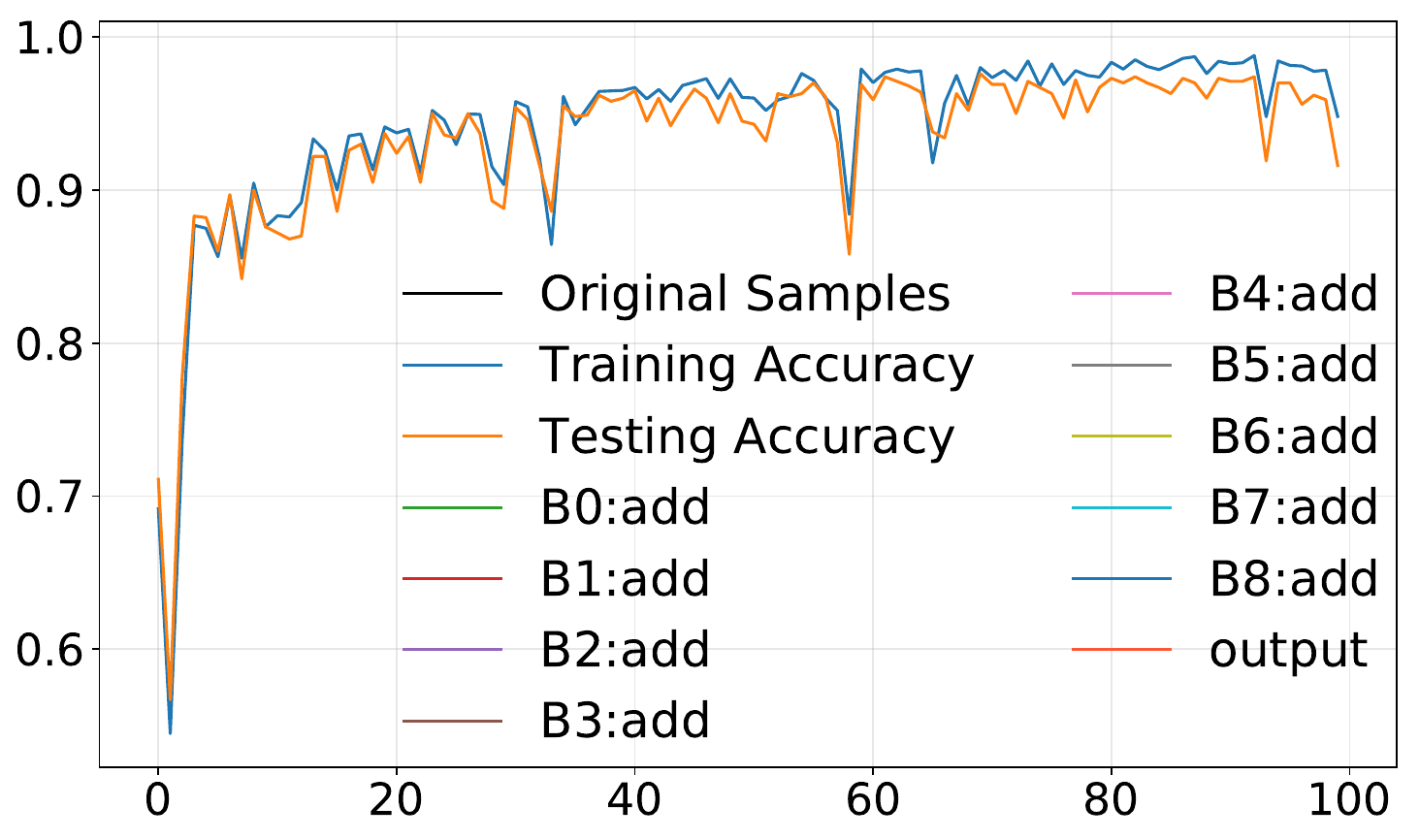}
	}
	
	\caption{MD-LSM and Accuracy Curves of ResNet's Main Blocks}
	\label{fig:res-block}

\end{figure}

\begin{figure}[htbp]

	\centering	
	\subfigure[$ {\rm LS}_0$]{
		\includegraphics[width=0.98\textwidth]{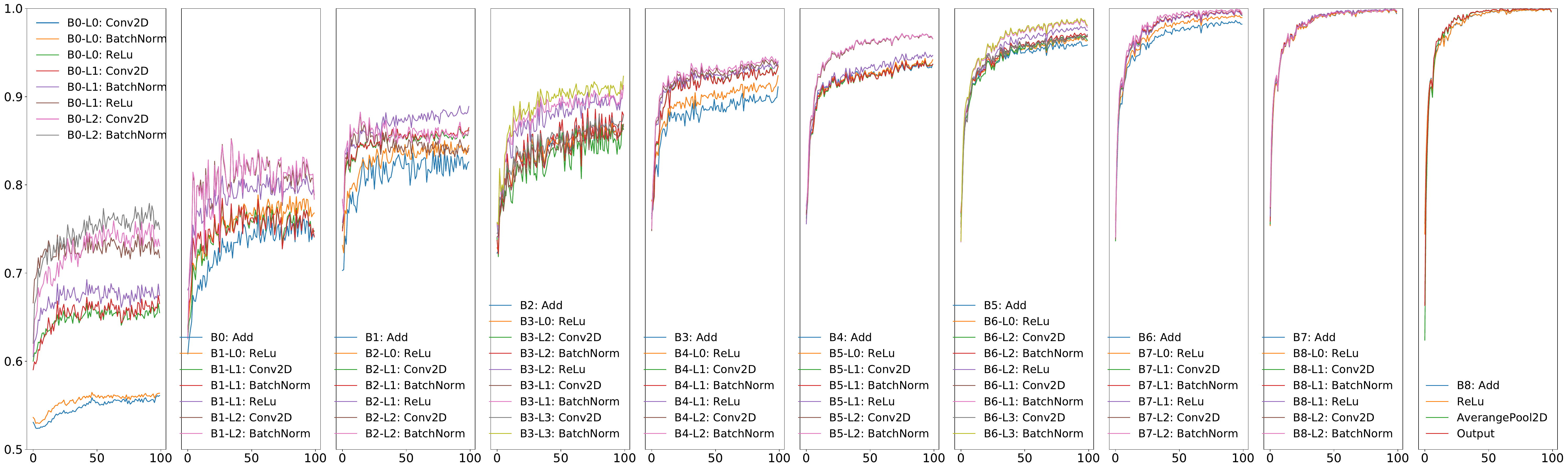}
	}
	\subfigure[$ {\rm LS}_1$]{
		\includegraphics[width=0.98\textwidth]{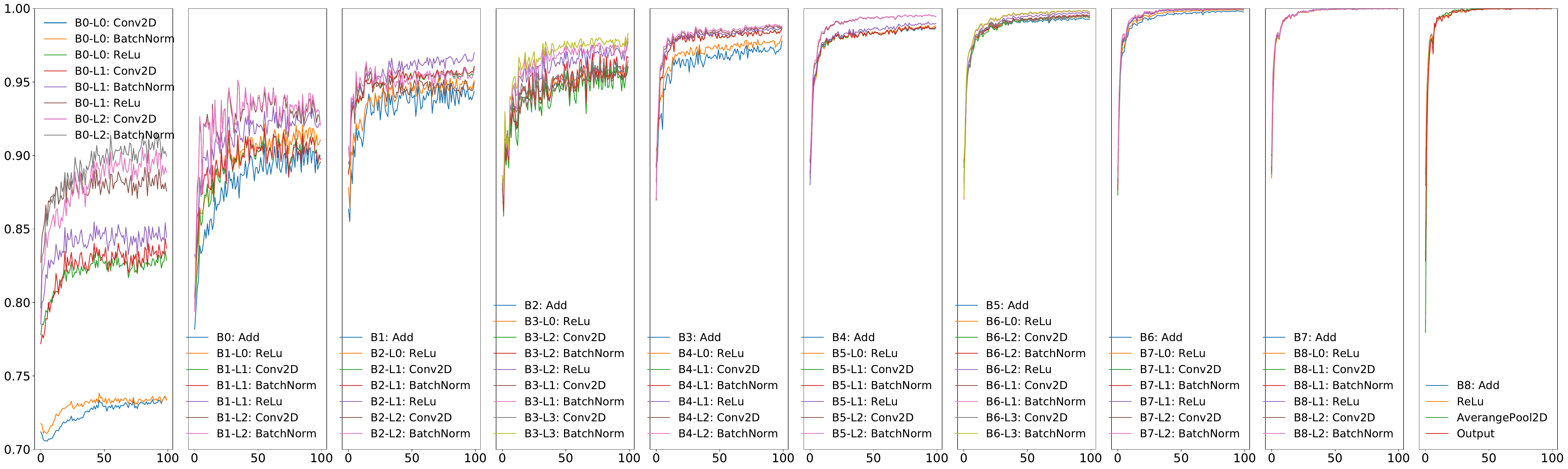}
	}
	\subfigure[$ {\rm LS}_2$]{
		\includegraphics[width=0.98\textwidth]{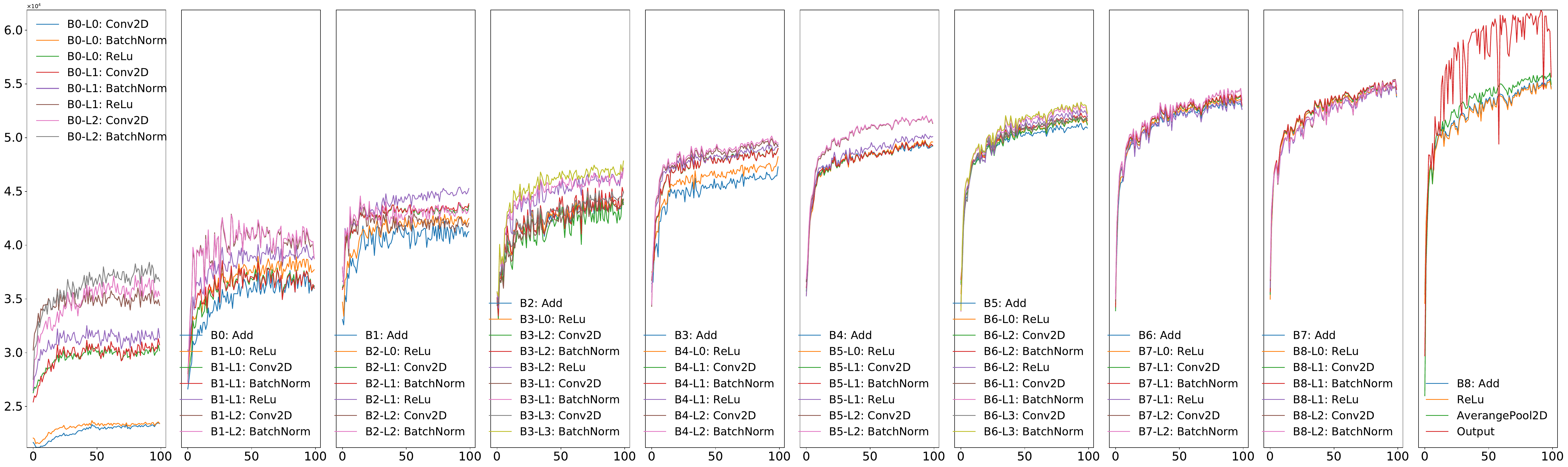}
	}
	
	\caption{MD-LSM and Accuracy Curves of ResNet's Hidden Layers}\label{fig:res-layer}
	
\end{figure}


\begin{figure}[htbp]
	\centering
	\includegraphics[width=0.9\textwidth]{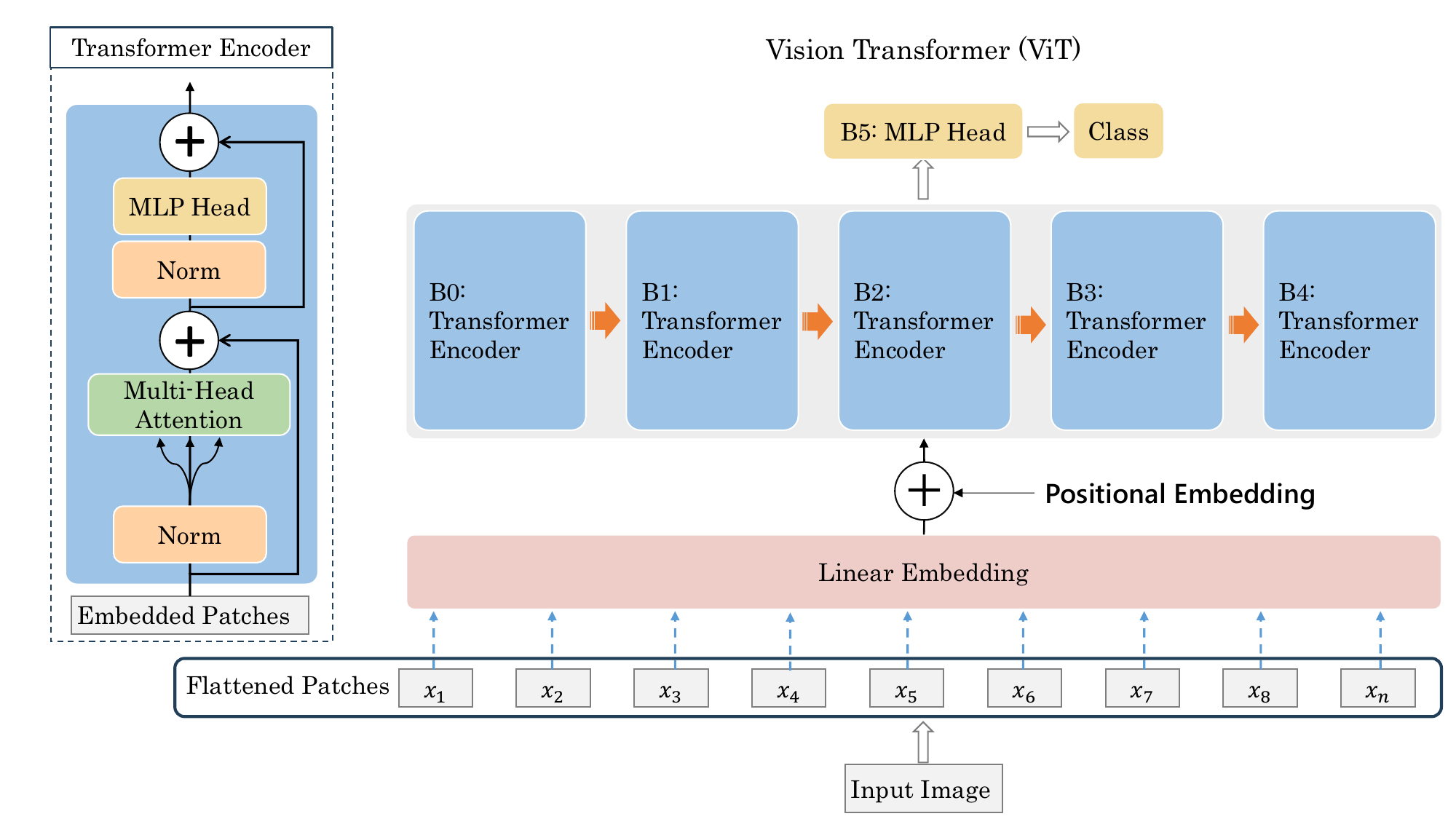}
	\caption{Structure of ViT}
	\label{fig:vit-structure}
\end{figure}

\begin{figure}[htbp]
	
		\centering	
	\subfigure[$ {\rm LS}_0$]{
		\includegraphics[width=0.4\textwidth]{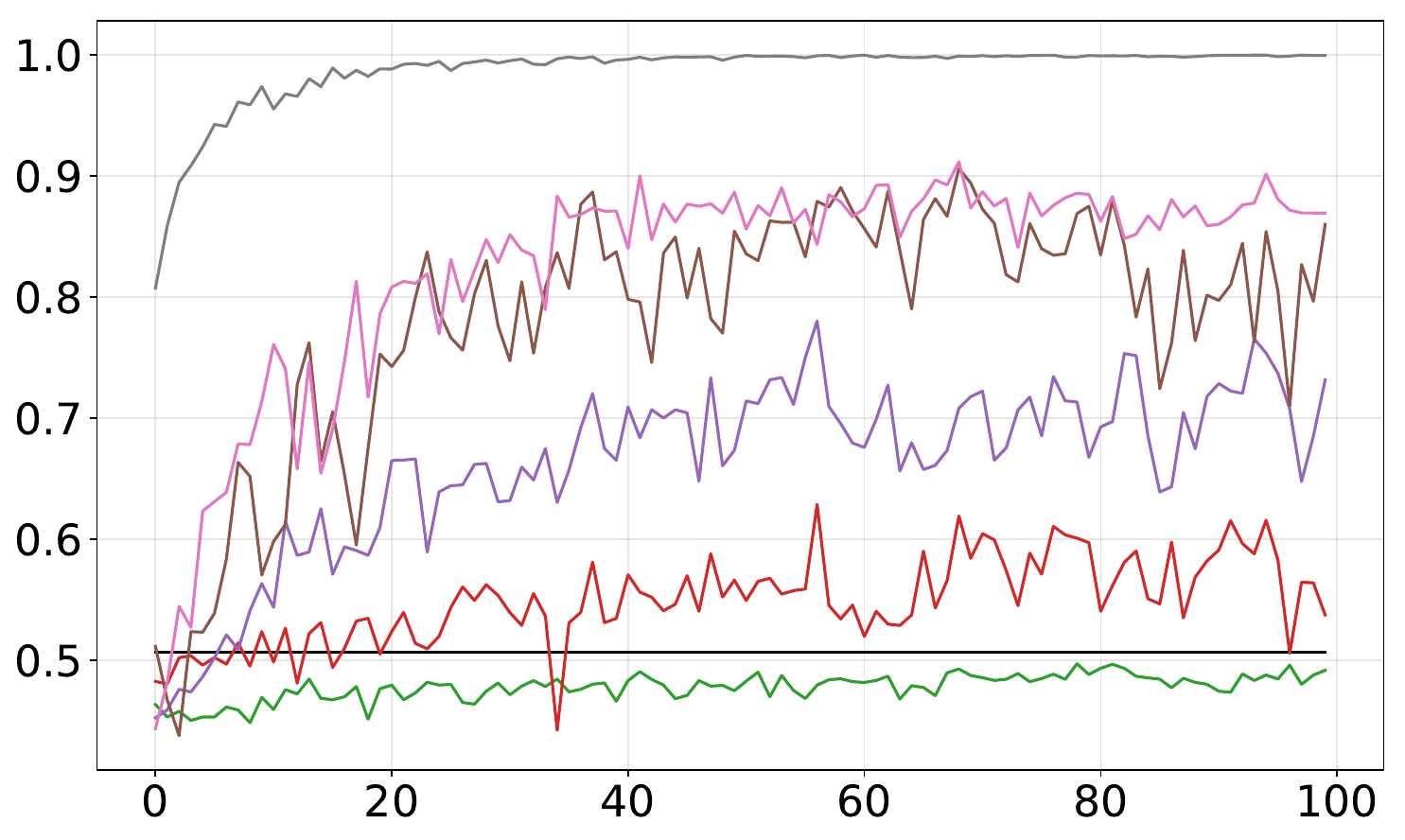}
	}
	\subfigure[$ {\rm LS}_1$]{
		\includegraphics[width=0.4\textwidth]{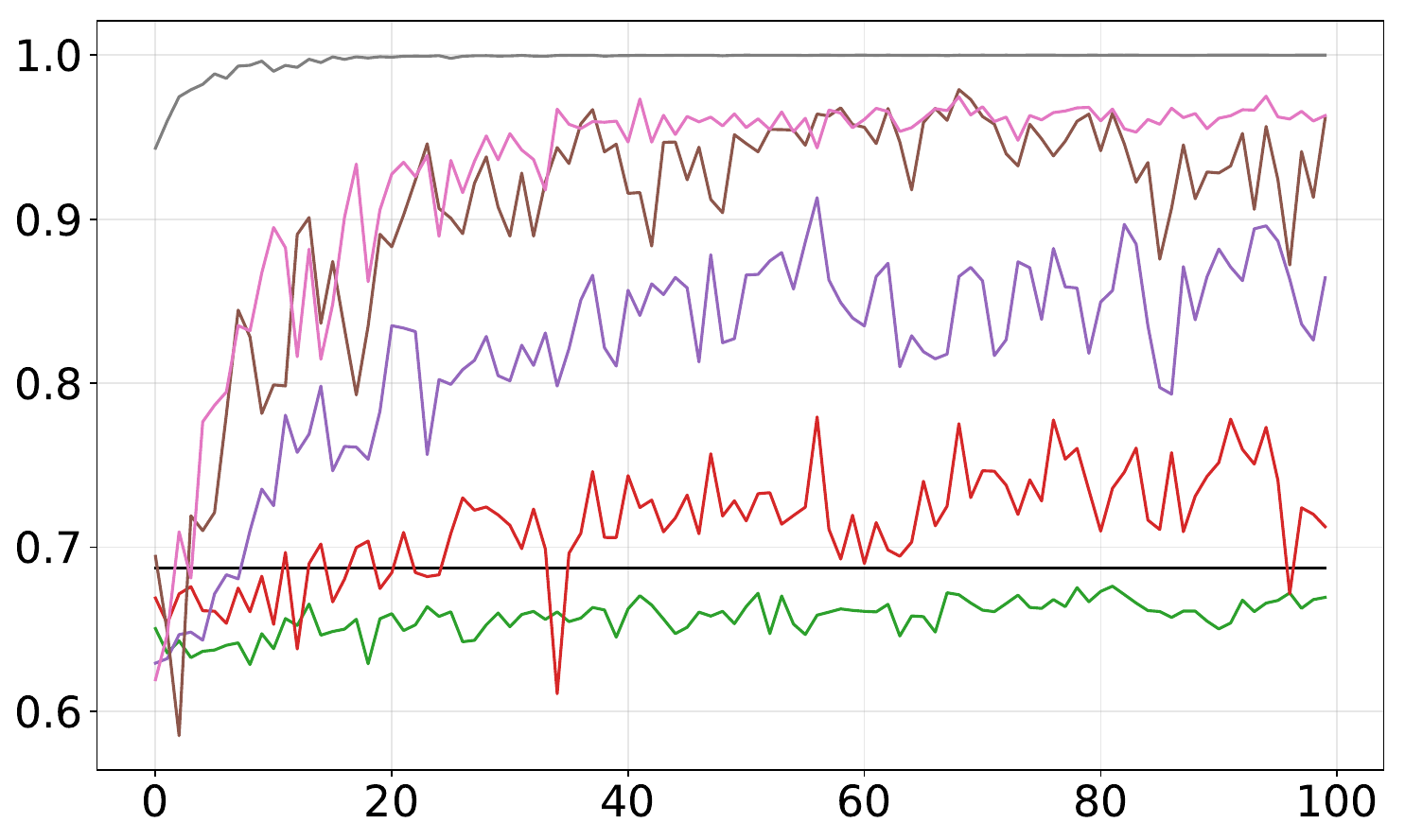}
	}
	\subfigure[$ {\rm LS}_2$]{
		\includegraphics[width=0.4\textwidth]{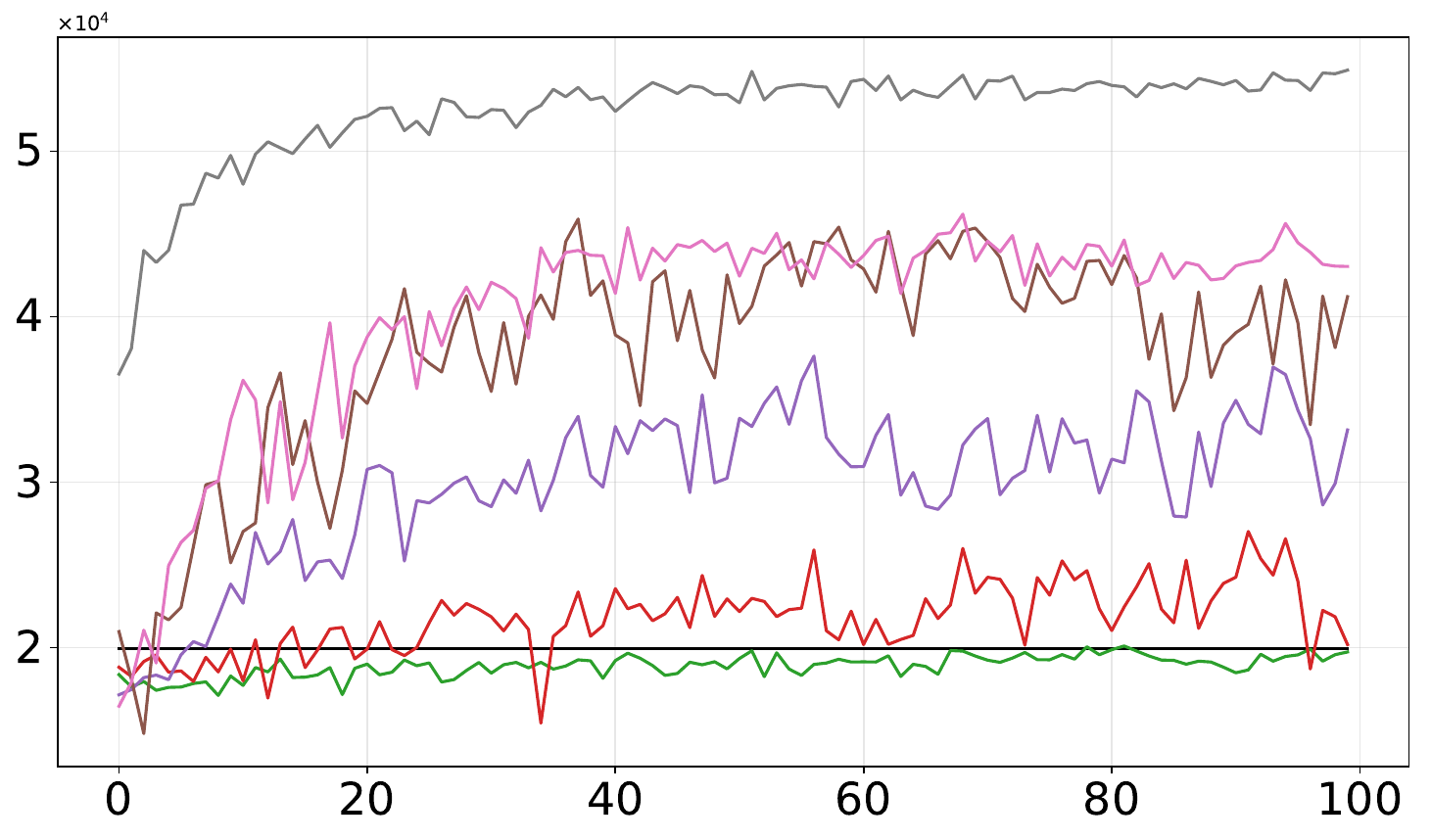}
	}
	\subfigure[Accuracy]{
		\includegraphics[width=0.4\textwidth]{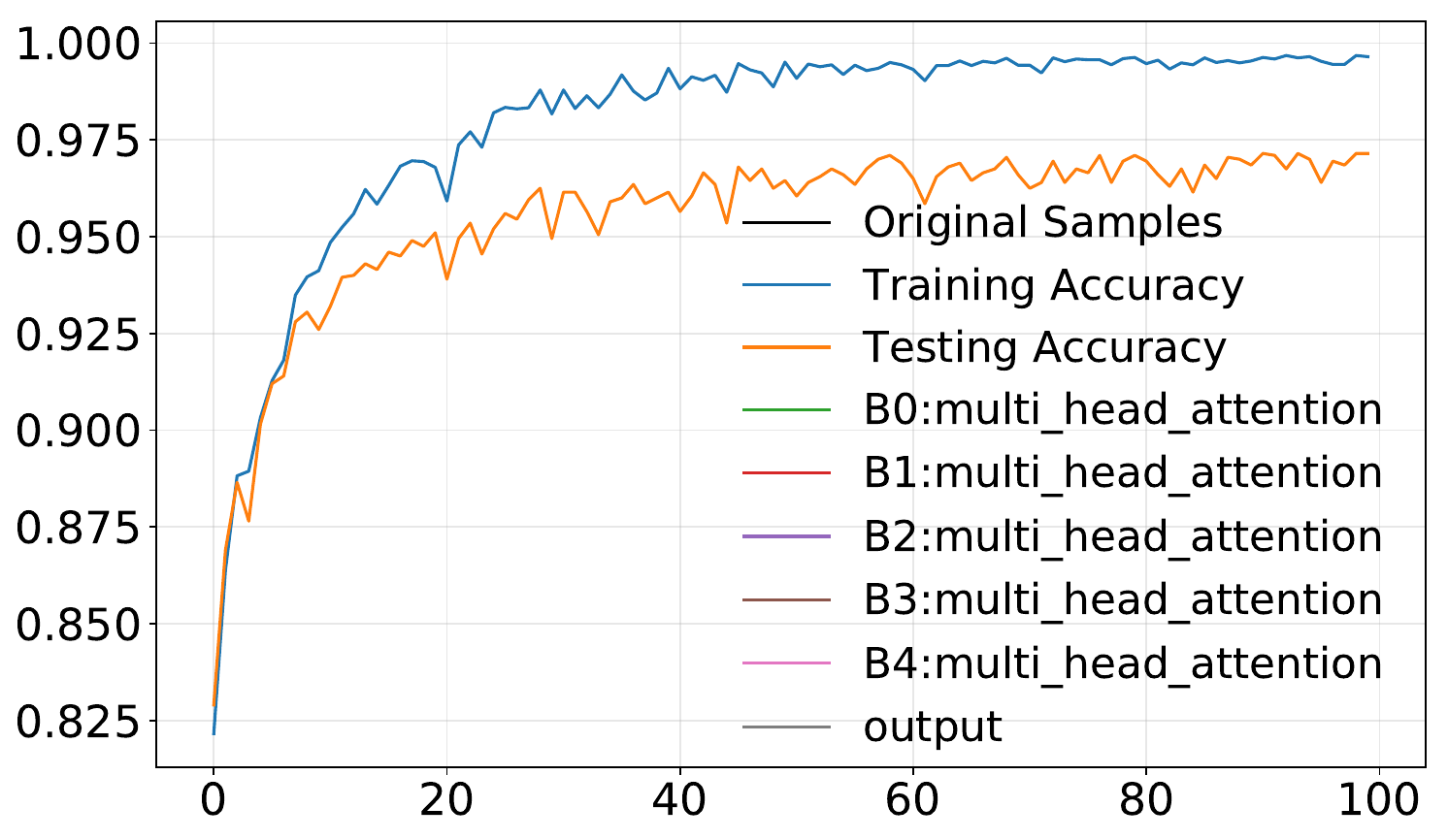}
	}
	
	\caption{MD-LSM and Accuracy Curves of ViT's Main Blocks}
	\label{fig:vit-block}	
\end{figure}

\begin{figure}[htbp]
	
	\centering	
	\subfigure[$ {\rm LS}_0$]{
		\includegraphics[width=0.98\textwidth]{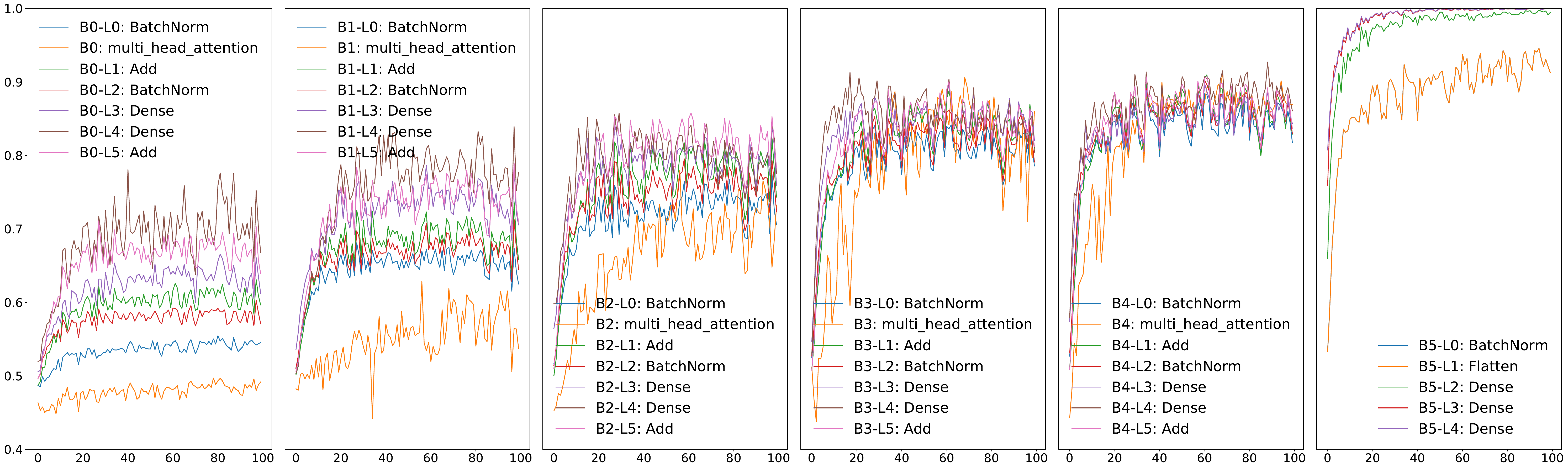}
	}
	\subfigure[$ {\rm LS}_1$]{
		\includegraphics[width=0.98\textwidth]{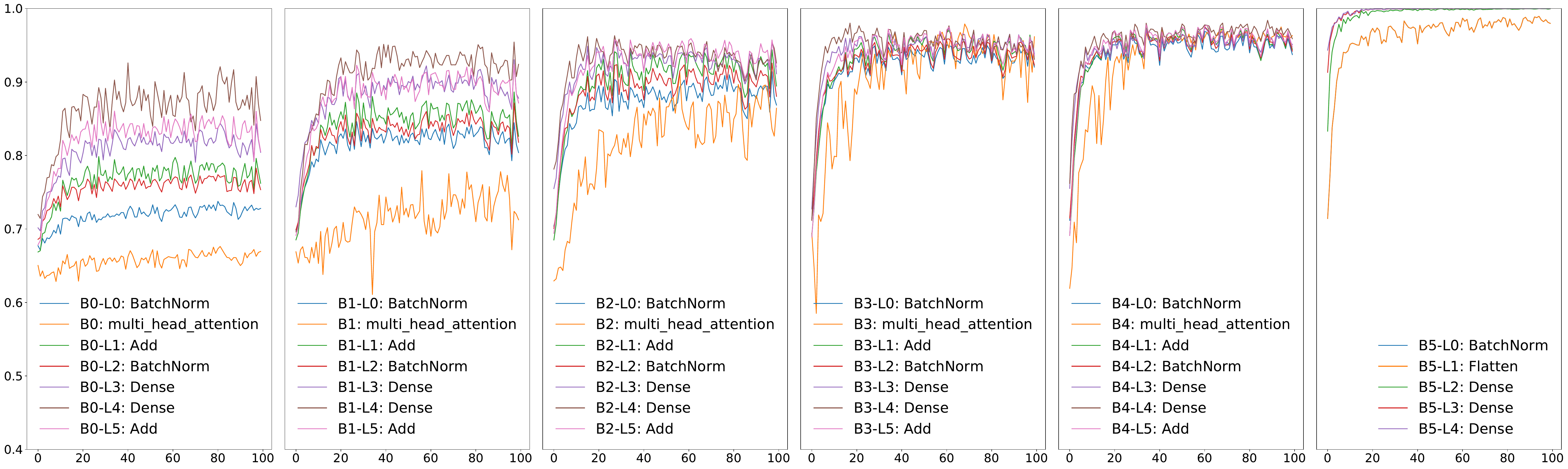}
	}
	\subfigure[$ {\rm LS}_2$]{
		\includegraphics[width=0.98\textwidth]{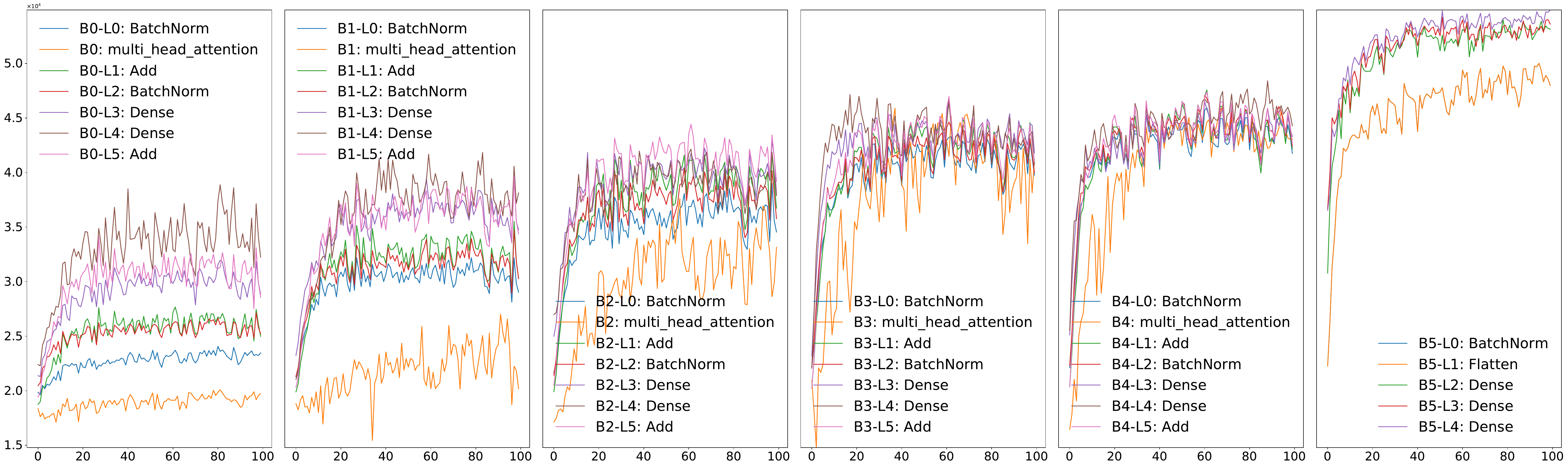}
	}
	
	\caption{MD-LSM and Accuracy Curves of ViT's Hidden Layers}\label{fig:vit-layer}	
	
\end{figure}


\clearpage


\end{document}